\documentclass[10pt]{article} 

\usepackage[accepted]{tmlr}

\usepackage{amsmath,amsfonts,bm}

\def\eqref#1{equation~\ref{#1}}

\def\1{\bm{1}}

\def\vd{{\bm{d}}}
\def\ve{{\bm{e}}}

\def\vx{{\bm{x}}}
\def\vy{{\bm{y}}}
\def\vz{{\bm{z}}}

\DeclareMathAlphabet{\mathsfit}{\encodingdefault}{\sfdefault}{m}{sl}
\SetMathAlphabet{\mathsfit}{bold}{\encodingdefault}{\sfdefault}{bx}{n}

\usepackage{hyperref}
\usepackage{url}

\usepackage{algorithm}
\usepackage{algorithmic}

\title{
Unsupervised Discovery of Steerable Factors\\When Graph Deep Generative Models Are Entangled
}

\author{
\name Shengchao Liu \email shengchao.liu@umontreal.ca\\
\addr Quebec AI Institute (Mila)\\
University de Montréal\AND
\name Chengpeng Wang \email cw83@illinois.edu\\
\addr University of Illinois Urbana-Champaign
\AND
\name Jiarui Lu \email jiarui.lu@umontreal.ca\\
\addr Quebec AI Institute (Mila)\\
University de Montréal
\AND
\name Weili Nie \email wnie@nvidia.com\\
\addr Nvidia Research
\AND
\name Hanchen Wang \email hw501@cam.ac.uk\\
\addr University of Cambridge
\AND
\name Zhuoxinran Li \email zhuoxinran.li@mail.utoronto.ca\\
\addr University of Toronto
\AND
\name Bolei Zhou \email bolei@cs.ucla.edu\\
\addr University of California, Los Angeles
\AND
\name Jian Tang \email jian.tang@hec.ca\\
\addr Quebec AI Institute (Mila)\\
HEC Montréal
}

\def\month{01}  
\def\year{2024} 
\def\openreview{\url{https://openreview.net/forum?id=wyU3Q4gahM}} 

\usepackage{booktabs}

\usepackage[capitalize,noabbrev]{cleveref}
\usepackage{multicol, multirow}
\usepackage{makecell}
\usepackage{subfigure}
\usepackage{adjustbox}
\usepackage{enumitem}

\usepackage{wrapfig}
\usepackage{bbm}
\usepackage{longtable}

\newcommand{\ie}{\em{i.e.}}
\newcommand{\eg}{\em{e.g.}}

\usepackage{xcolor}
\newcommand{\model}{GraphCG}
\newcommand{\mz}{\bar\vz}

\begin{document}

\maketitle

\begin{abstract}
Deep generative models (DGMs) have been widely developed for graph data. However, much less investigation has been carried out on understanding the latent space of such pretrained graph DGMs. These understandings possess the potential to provide constructive guidelines for crucial tasks, such as graph controllable generation. Thus in this work, we are interested in studying this problem and propose GraphCG, a method for the unsupervised discovery of steerable factors in the latent space of pretrained graph DGMs. We first examine the representation space of three pretrained graph DGMs with six disentanglement metrics, and we observe that the pretrained representation space is entangled. Motivated by this observation, GraphCG learns the steerable factors via maximizing the mutual information between semantic-rich directions, where the controlled graph moving along the same direction will share the same steerable factors. We quantitatively verify that GraphCG outperforms four competitive baselines on two graph DGMs pretrained on two molecule datasets. Additionally, we qualitatively illustrate seven steerable factors learned by GraphCG on five pretrained DGMs over five graph datasets, including two for molecules and three for point clouds.
\end{abstract}

\section{Introduction} \label{sec:intro}

The graph is a general format for many real-world data. For instance, molecules can be treated as graphs~\citep{duvenaud2015convolutional,gilmer2017neural} where the chemical atoms and bonds correspond to the topological nodes and edges respectively. Processing point clouds as graphs is also a popular strategy~\citep{shi2020point,wang2019dynamic}, where points are viewed as nodes and edges are built among the nearest neighbors. Many existing works on deep generative models (DGMs) focus on modeling the graph data and improving the synthesis quality. However, understanding the pretrained graph DGMs and their learned representations has been much less explored. This may hinder the development of important applications like \textit{graph controllable generation} (also referred to as \textit{graph editing}) and the discovery of interpretable graph structure.

Concretely, the graph controllable generation task refers to modifying the steerable factors of the graph so as to obtain graphs with desired properties easily~\citep{drews2000drug,pritch2009shift}. This is an important task in many applications, but traditional methods ({\eg}, manual editing) possess inherent limitations under particular circumstances. A typical example is molecule editing: it aims at modifying the substructures of molecules~\citep{mihalic1992graph} and is related to certain key tactics in drug discovery like functional group change~\citep{ertl2020most} and scaffold hopping~\citep{bohm2004scaffold,hu2017recent}. This is a routine task in pharmaceutical companies, yet, relying on domain experts for manual editing can be subjective or biased~\citep{drews2000drug,gomez2018decision}. Different from previous works, this paper starts to explore unsupervised graph editing on pretrained DGMs. It can act as a complementary module to conventional methods and bring many crucial benefits: (1) It enables efficient graph editing in a large-scale setting. (2) It alleviates the requirements for extensive domain knowledge for factor change labeling. (3) It provides a constructive perspective for editing preference, which can reduce biases from the domain experts.

\textbf{Disentanglement for editing.} One core property relevant to the general unsupervised data editing using DGMs is disentanglement. While there does not exist a widely-accepted definition of disentanglement, the key intuition~\citep{locatello2019challenging} is that a disentangled representation should separate the distinct, informative, and steerable factors of variations in the data. Thus, the controllable generation task would become trivial with the disentangled DGMs as the backbone. Such a disentanglement assumption has been widely used in generative modeling on the image data, {\eg}, $\beta$-VAE~\citep{higgins2016beta} learns disentangled representation by forcing the representation to be close to an isotropic unit Gaussian. However, it may introduce extra constraints on the formulations and expressiveness of DGMs~\citep{higgins2016beta,ridgeway2018learning,eastwood2018framework,wu2021stylespace}.\looseness=-1

\textbf{Entanglement on pretrained graph DGMs.} Thus for graph data, one crucial question arises: \textit{Is the latent representation space from pretrained graph DGMs disentangled or not?} In image generation, a series of work~\citep{collins2020editing,shen2020interpreting,harkonen2020ganspace,tewari2020stylerig,wu2021stylespace} has shown the disentanglement properties on pretrained DGMs. However, such property of pretrained graph DGMs is much less explored. In~\Cref{sec:entangled_representation}, we first study the latent space of three pretrained graph DGMs and empirically illustrate that the learned space is not perfectly disentangled or entangled. In what follows, we adopt the term ``entangled'' for graph DGMs.\looseness=-1

\begin{figure}[tb]
\centering
    \begin{subfigure}[\small Learning phase of \model{}.]
    {\includegraphics[width=0.4\linewidth]{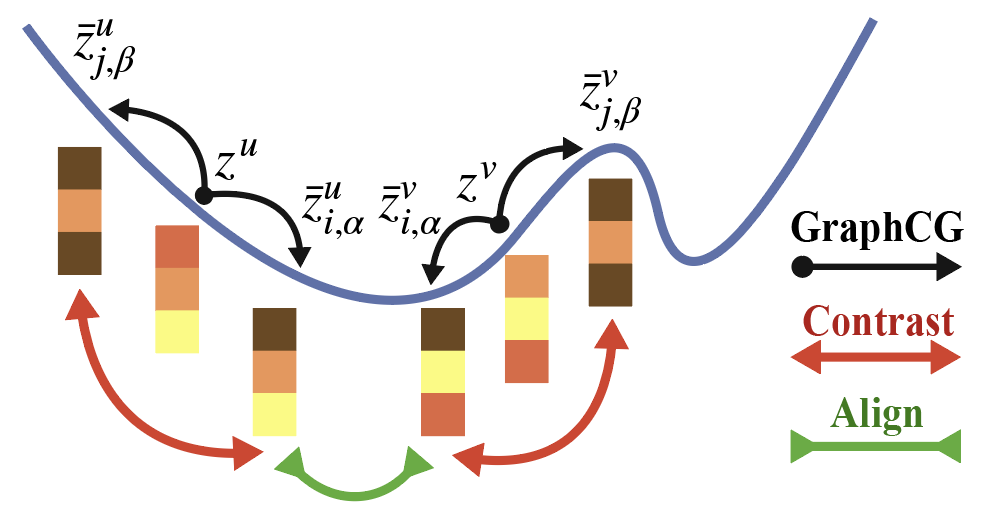}
    \label{fig:GraphCG_training}
    }
    \end{subfigure}
\hfill
     \begin{subfigure}[\small Inference phase of \model{}.]
     {\includegraphics[width=0.55\linewidth]{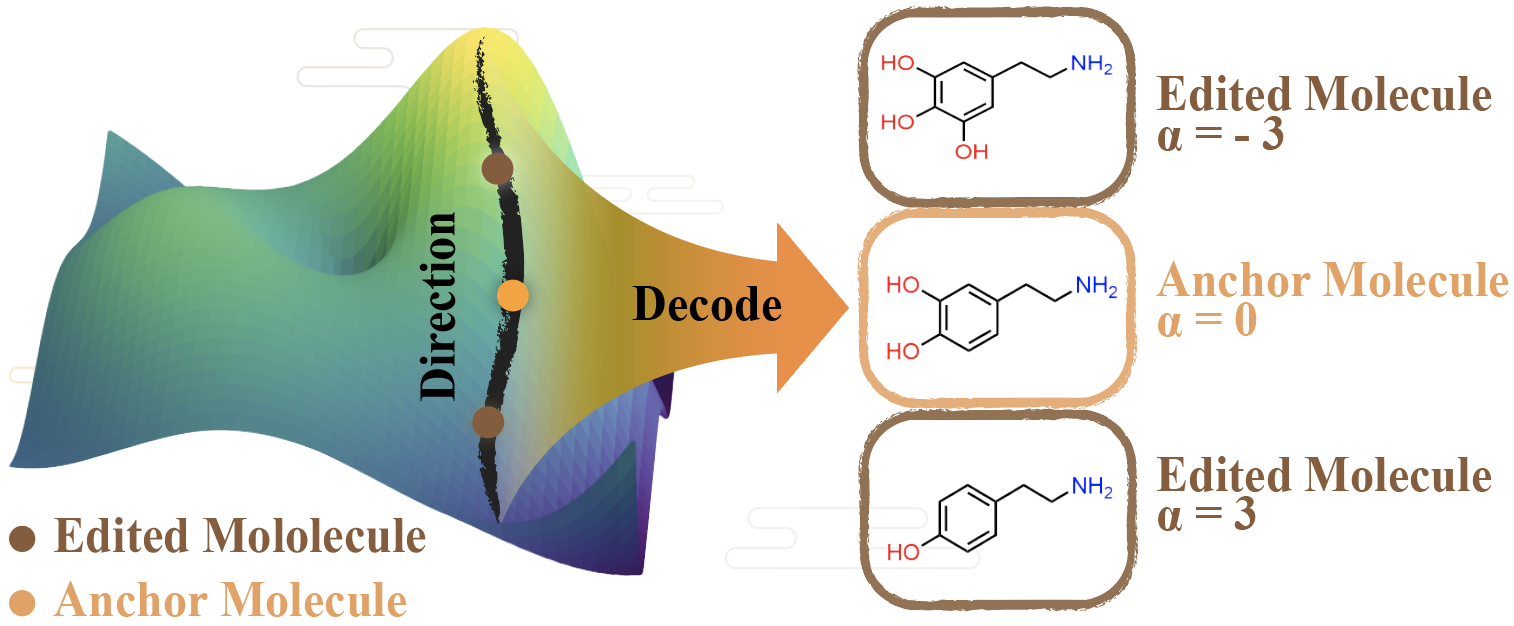}
     \label{fig:GraphCG_test}
     }
     \end{subfigure}
\vspace{-1ex}
\caption{
\small
(a) The learning phase. Given two latent codes $\vz^u$ and $\vz^v$, we edit the four latent representations along $i$-th and $j$-th direction with step size $\alpha$ and $\beta$ respectively. The goal of \model{}, is to align the positive pair ($\mz^u_{i, \alpha}$ and $\mz^v_{i, \alpha}$), and contrast them with $\mz^u_{j, \beta}$ and $\mz^v_{j, \beta}$ respectively. (b) The inference phase. We will first sample an anchor molecule and adopt the learned directions in the learning phase for editing. With step size $\alpha \in [-3, 3]$, we can generate a sequence of molecules. Specifically, after decoding, there is a functional group change shown up: the number of hydroxyl groups decreases along the sequence in the decoded molecules.
\looseness=-1
}
\label{fig:GraphCG_pipeline}
\vspace{-3ex}
\end{figure}

\textbf{Our approach.} This observation then raises the second question: \textit{Given a pretrained yet entangled DGM for graph data, is there a flexible framework enabling the graph controllable generation in an unsupervised manner?} To tackle this problem, we propose a model-agnostic framework coined \model{} for unsupervised graph controllable generation. \model{} has two main phases, as illustrated in~\Cref{fig:GraphCG_pipeline}.
During the learning phase (\Cref{fig:GraphCG_training}), \model{} starts with the assumption that the steerable factors can be learned by maximizing the mutual information (MI) among the semantic directions. We formulate \model{} using an energy-based model (EBM), which offers a large family of solutions. Then during the inference phase, with the learned semantic directions, we can carry out the editing task by moving along the direction with certain step sizes. As the example illustrated in~\Cref{fig:GraphCG_test}, the graph structure (hydroxyl group) changes consistently along the learned editing direction. For evaluation, we qualitatively verify the learned directions of five pretrained graph DGMs. Particularly for the molecular datasets, we propose a novel evaluation metric called sequence monotonic ratio (SMR) to quantitatively measure the structure change over the output sequences.\looseness=-1

\textbf{Our contributions.} (1) We conduct an empirical study on the disentanglement property of three pretrained graph DGMs using six metrics, and we observe that the latent space of these pretrained graph DGMs is entangled. (2) We propose a model-agnostic method called \model{} for the unsupervised graph controllable generation or graph editing. \model{} aims at learning the steerable factors by maximizing the mutual information among corresponding directions, and its outputs are sequences of edited graphs. (3) We quantitatively verify that \model{} outperforms four competitive baselines when evaluated on two pretrained graph DGMs over two molecule datasets. (4) We further qualitatively strengthen the effectiveness of \model{} by illustrating seven semantic factors on five pretrained graph DGMs over five graph datasets, including two for molecular graphs and three for point clouds.\looseness=-1

\textbf{Related work.}
Recent works leverage the DGMs for various controllable generation tasks~\citep{chen2018rise,xia2021gan}, where the inherent assumption is that the learned latent representations encode rich semantics, and thus traversal in the latent space can help steer factors of data~\citep{jahanian2019steerability,shen2020interfacegan,harkonen2020ganspace}. Among them, one research direction~\citep{shen2020interfacegan,nie2021controllable} is using supervised signals to learn the semantic-rich directions, and most works on editing the graph data focus on the supervised setting~\citep{jin2020hierarchical,veber2002molecular,you2018graph}. However, these approaches can not be applied to many realistic scenarios where extracting the supervised labels is difficult. Another research line~\citep{harkonen2020ganspace,shen2021closed,ren2021generative} considers discovering the latent semantics in an unsupervised manner, but these unsupervised methods are designed to be either model-specific or task-specific, making them not directly generalizable to the graph data. A more comprehensive discussion is in~\Cref{sec:app:related_work}.

\section{Background and Problem Formulation} \label{sec:background}

\textbf{Graph and deep generative models (DGMs).} Each graph data (including nodes and edges) is denoted as $\vx \in \mathcal{X}$, where $\mathcal{X}$ is the data space, and DGMs learn the data distribution, {\ie}, $p(\vx)$. Our proposed graph editing method (\model{}) is model-agnostic or DGM-agnostic, so we briefly introduce the mainstream DGMs for graph data as below. Variational auto-encoder (VAE)~\citep{kingma2013auto,higgins2016beta} measures a variational lower bound of $p(\vx)$ by introducing a proposal distribution; flow-based model~\citep{dinh2014nice, rezende2015variational} constructs revertible encoding functions such that the data distribution can be deterministically mapped to a prior distribution. Note that these mainstream DGMs, either explicitly or implicitly, contain an encoder ($f(\cdot)$) and a decoder ($g(\cdot)$) parameterized by neural networks:\looseness=-1
\begin{equation} \label{eq:formuation}
\small{
\vz = f(\vx), \quad\quad\quad\quad \vx' = g(\vz),
}
\end{equation}
where $\vz \in \mathcal{Z}$ is the latent representation, $\mathcal{Z}$ is the latent space, and $\vx'$ is the reconstructed output graph. Since in the literature~\citep{shen2021closed,shen2020interfacegan}, people also call latent representations as latent codes or latent vectors, in what follows, we will use these terms interchangeably. Note that the encoding and decoding functions in~\Cref{eq:formuation} ($f, g$) can be stochastic depending on the DGMs we are using.

\textbf{Steerable factors.} The steerable factors are key attributes of DGMs, referring to the semantic information of data that we can explicitly discover from the pretrained DGMs. For instance, existing works~\citep{shen2021closed,ren2021generative} have shown that using unsupervised methods on facial image DGM can discover factors such as the size of eyes, smiles, noses, etc. In this work, we focus on the steerable factors of graph data, which are data- and task-specific. Yet, there is one category of factor that is commonly shared among all the graph data: the \textbf{structure} factor. Concretely, these steerable factors can be the functional groups in molecular graphs and shapes or sizes in point clouds. The details of these steerable factors are in~\Cref{app:sec:toy_example}.\looseness=-1

\textbf{Semantic direction and step size.} To learn the steerable factors using deep learning tools, we will introduce the semantic directions defined on the latent space of DGM. In such a space $\mathcal{Z}$, we assume there exist $D$ semantically meaningful direction vectors, $\vd_i$ with $i \in \{0, 1, \hdots, D-1\}$.\footnote{In unsupervised editing, the steerable factors on each semantic direction is known by post-training human selection.} There is also a scalar variable, step size $\alpha$, which controls the degree to edit the sampled data with desired steerable factors (as will be introduced below), and we follow the prior work~\citep{shen2021closed} on taking $\alpha \in [-3, 3]$. Each direction corresponds to one or multiple factors, such that by editing the latent vector $\vz$ along $\vd_i$ with step size $\alpha$, the reconstructed graph will possess the desired structural modifications. The editing with a sequence of step sizes $\alpha \in [-3, 3]$ along the same direction $\vd_i$ leads to a sequence of edited graphs.

\textbf{Problem formulation: graph editing or graph controllable generation.}
Given a \textit{pretrained} DGM ({\ie}, the encoder and decoder are fixed), our goal is to learn the most semantically rich directions ($\vd_i$) in the latent space $\mathcal{Z}$. Then for each latent code $\vz$, with the $i$-th semantic direction and a step size $\alpha$, we can get an edited latent vector $\mz_{i, \alpha}$ and edited data $\bar\vx'$ after decoding $\mz_{i, \alpha}$, as:
\begin{equation} \label{eq:editing_function}
\small{
\vz = f(\vx), \quad\quad \mz_{i,\alpha} = h(\vz, \vd_i, \alpha), \quad\quad \bar \vx' = g(\mz_{i,\alpha} ),
}
\end{equation}
where $\vd_i$ and $h(\cdot)$ are the edit direction and edit functions that we want to learn. We expect that $\mz_{i, \alpha}$ can inherently possess certain steerable factors, which can be reflected in the graph structure of $\bar\vx'$.

\textbf{Energy-based model (EBM).}
EBM is a flexible framework for distribution modeling: 
\begin{equation}\label{eq:energy_based_model}
\small{
\begin{aligned}
p(\vx) = \frac{\exp(-E(\vx))}{A} = \frac{\exp(-E(\vx))}{\int_\vx \exp(-E(\vx)) d\vx},
\end{aligned}
}
\end{equation}
where $E(\cdot)$ is the energy function and $A$ is the partition function. In EBM, the bottleneck is the estimation of partition function $A$. It is often intractable due to the high cardinality of $\mathcal{X}$. Various methods have been proposed to handle this issue, including but not limited to contrastive divergence~\citep{hinton2002training}, noise-contrastive estimation~\citep{gutmann2010noise,che2020your}, and score matching~\citep{hyvarinen2005estimation,song2019generative,song2020score}.

\section{Entanglement of Latent Representation for Graph DGMs} \label{sec:entangled_representation}
In this section, we quantify the degree of disentanglement of the existing DGMs for graph data. Recall that the key intuition~\citep{locatello2019challenging} behind disentanglement is that a disentangled representation space should separate the distinct, informative, and steerable factors of variations in the data. In other words, each latent dimension of the disentangled representation corresponds to one or multiple factors. Therefore, the change of the disentangled dimension can lead to the consistent change in the corresponding factors of the data. This good property has become a foundational assumption in many existing controllable generation methods~\citep{shen2020interfacegan,shen2021closed,harkonen2020ganspace}.

In the computer vision domain, StyleGAN~\citep{karras2019style} is one of the most recent works on image generation, and several works have proven its nice disentanglement property~\citep{collins2020editing,shen2020interpreting,harkonen2020ganspace,tewari2020stylerig,wu2021stylespace}. In image generation, \citep{locatello2019challenging} shows that without inductive bias, the representation learned by VAEs is not perfectly disentangled. Then the next question naturally arises: \textit{Is the latent space of pretrained graph DGMs disentangled or not?} To answer this question, we conduct the following experiment.\looseness=-1

\begin{table}[tb]
\setlength{\tabcolsep}{8pt}
\fontsize{10}{8}\selectfont
\centering
\caption{
\small
The six disentanglement metrics on three pretrained DGMs and two graph types. All measures range from 0 to 1, and higher scores mean more disentangled representation.
}
\label{tab:disentanglement_evaluation}
\begin{adjustbox}{max width=\textwidth}
\begin{tabular}{l l l c c c c c c}
\toprule
Graph Type & DGM & Dataset & BetaVAE $\uparrow$ & FactorVAE $\uparrow$ & MIG $\uparrow$ & DCI $\uparrow$ & Modularity $\uparrow$ & SAP $\uparrow$ \\
\midrule
\multirow{2}{*}{Molecular Graph}
& MoFlow & ZINC250K & 0.260 & 0.175 & 0.031 & 0.953 & 0.620 & 0.009\\
& HierVAE & ChEMBL & 0.178 & 0.165 & 0.022 & 0.114 & 0.606 & 0.026\\
\midrule
Point Cloud
& PointFlow & Airplane & 0.022 & 0.025 & 0.029 & 0.160 & 0.745 & 0.022\\
\bottomrule
\end{tabular}
\end{adjustbox}
\vspace{-2ex}
\end{table}

There have been a series of works exploring the disentanglement of the latent space in DGMs, and here we take six widely-used ones: BetaVAE~\citep{higgins2016beta}, FactorVAE~\citep{kim2018disentangling}, MIG~\citep{chen2018isolating}, DCI~\citep{eastwood2018framework}, Modularity~\citep{ridgeway2018learning}, and SAP~\citep{kumar2018variational}. Each measure has its own bias, and we put a detailed comparison in~\Cref{app:sec:toy_example}. Meanwhile, they all share the same high-level idea: given the latent representation from a pretrained DGM, they are proposed to measure how predictive it is to certain steerable factors.

To adapt them to our setting, first, we need to extract the steerable factors in graph DGMs, which requires domain knowledge. For instance, in molecular graphs, we can extract some special substructures called fragments or functional groups. These substructures can be treated as steerable factors since they are the key components of the molecules and are closely related to certain molecular properties~\citep{doi:10.1021/ed064p575}. We use RDKit~\citep{landrum2013rdkit} to extract the 11 most distinguishable fragments as steerable factors for disentanglement measurement. For point clouds, we use PCL tool~\citep{Rusu_ICRA2011_PCL} to extract 75 VFH descriptors~\citep{rusu2010fast} as steerable factors, which depicts the geometries and viewpoints accordingly.\looseness=-1

Then to measure the disentanglement on graph DGMs, we consider six metrics on three datasets and two data types with three backbone models. All the metric values range from 0 to 1, and the higher the value, the more disentangled the DGM is. According to~\Cref{tab:disentanglement_evaluation}, we can observe that most of the disentanglement scores are quite low, except the DCI~\citep{eastwood2018framework} on MoFlow. Thus, we draw the conclusion that, generally, these graph DGMs are entangled. More details of this experiment (the steerable factors on two data types and six disentanglement metrics) can be found in~\Cref{app:sec:toy_example}.

\section{Our Method} \label{sec:method}
The analysis in~\Cref{sec:entangled_representation} naturally raises the next research question: \textit{Given an entangled DGM, is there a flexible way to do the graph data editing in an unsupervised manner?} The answer is positive. We propose \model, a flexible model-agnostic framework to learn the semantic directions in an unsupervised manner. It starts with the assumption that the latent representations edited with the same semantic direction and step size should possess similar information (with respect to the factors) to a certain degree, thus by maximizing the mutual information among them, we can learn the most semantic-rich directions. Then we formulate this editing task as a density estimation problem with the energy-based model (EBM). As introduced in~\Cref{sec:background}, EBM covers a broad range of solutions, and we further propose \model{}-NCE by adopting the noise-contrastive estimation (NCE) solution.

\subsection{GraphCG with Mutual Information}
\textbf{Motivation: learning semantic directions using MI on entangled DGM.} Recall that our ultimate goal is to enable graph editing based on semantic vectors. Existing deep generative models are entangled, thus obtaining such semantic vectors is a nontrivial task. To handle this problem, we propose using mutual information (MI) to learn the semantics. MI measures the non-linear dependency between variables. Here we set the editing condition as containing both the semantic directions and step sizes. We assume that maximizing the MI between different conditions can maximize the shared information within each condition, {\ie}, graphs moving along the same condition share more semantic information. The pipeline is as follows.

We first sample two latent codes in the latent space, $\vz^u$ and $\vz^v$. Such two latent codes will be treated as positive pairs, and their construction will be introduced in~\Cref{sec:GraphCG_with_NCE}. Then we pick up the $i$-th semantic direction and one step size $\alpha$ to obtain the edited latent codes in the latent space $\mathcal{Z}$ as:
\begin{equation}
\small{
\begin{aligned}
\mz^u_{i, \alpha} = h(\vz^u, \vd_i, \alpha), \quad\quad
\mz^v_{i, \alpha} = h(\vz^v, \vd_i, \alpha).
\end{aligned}
}
\end{equation}
Under our assumption, we expect that these two edited latent codes share certain information with respect to the steerable factors. Thus, we want to maximize the MI between $\mz^u_{i, \alpha}$ and $\mz^v_{i, \alpha}$. 
Since the MI is intractable to compute, we adopt the EBM lower bound from \citep{liu2021pre} as:
\begin{equation} \label{eq:MI_lower_bound}
\fontsize{8}{1}\selectfont
\begin{aligned}
\mathcal{L}_{\text{MI}} (\mz^u_{i, \alpha}, \mz^v_{i, \alpha})
= \frac{1}{2} \mathbb{E}
\big[ \log p(\mz^u_{i, \alpha}|\mz^v_{i, \alpha}) + \log p(\mz^v_{i, \alpha}|\mz^u_{i, \alpha}) \big].
\end{aligned}
\end{equation}
The detailed derivation is in~\Cref{app:sec:lower_bound_MI}. Till this step, we have transformed the graph data editing task into the estimation of two conditional log-likelihoods.

\subsection{\model{} with Energy-Based Model}
Following~\Cref{eq:MI_lower_bound}, maximizing the MI between $I\big(\mz^u_{i, \alpha}; \mz^v_{i, \alpha} \big)$ is equivalent to estimating the summation of two conditional log-likelihoods. We then model them using two conditional EBMs. Because these two views are in the mirroring direction, we may as well take one for illustration. For example, for the first conditional log-likelihood, we can model it with EBM as:
\begin{equation}
\small
\begin{aligned}
p(\mz^u_{i, \alpha}|\mz^v_{i, \alpha}) 
= \frac{\exp(-E(\mz^u_{i, \alpha}, \mz^v_{i, \alpha}))}{ \int \exp(-E({\mz^{u'}_{i, \alpha}}, \mz^v_{i, \alpha})) d{\mz^{u'}_{i, \alpha}}} 
= \frac{\exp (f(\mz^u_{i, \alpha},\mz^v_{i, \alpha}))}{A_{ij}},
\end{aligned}
\end{equation}
where $E(\cdot)$ is the energy function, $A_{ij}$ is the intractable partition function, and $f(\cdot)$ is the negative energy. The energy function is flexible and we use the dot-product:
\begin{equation} \label{eq:energy_function}
\small{
\begin{aligned}
f(\mz^u_{i, \alpha}, \mz^v_{i, \alpha})
= \langle h(\vz^u, \vd_i, \alpha), h(\vz^v, \vd_i, \alpha)\rangle,
\end{aligned}
}
\end{equation}
where $h(\cdot)$ is the editing function introduced in~\Cref{eq:editing_function}. Similarly for the other conditional log-likelihood term, and the objective becomes:
\begin{equation} \label{eq:objective_ebm}
\fontsize{7.5}{1}\selectfont
\begin{aligned}
\mathcal{L}_{\text{\model}} =
\mathbb{E}
\big[ \log \frac{ \exp (f(\mz^u_{i, \alpha},\mz^v_{i, \alpha}))}{A_{ij}} + \log \frac{\exp(f(\mz^v_{i, \alpha},\mz^u_{i, \alpha}))}{A_{ji}} \big].
\end{aligned}
\end{equation}
With~\Cref{eq:objective_ebm}, we are able to learn the semantically meaningful direction vectors. We name this unsupervised graph controllable generation framework as \model. In specific, \model{} utilizes EBM for estimation, which yields a wide family of solutions, as introduced below.

\subsection{\model{} with Noise Contrastive Estimation} \label{sec:GraphCG_with_NCE}
We solve~\Cref{eq:objective_ebm} using the noise contrastive estimation (NCE)~\citep{gutmann2010noise}. The high-level idea of NCE is to transform the density estimation problem into a binary classification problem that distinguishes if the data comes from the introduced noise distribution or from the true distribution. NCE has been widely explored for solving EBM~\citep{song2021train}, and we adopt it as \model-NCE by optimizing
the following objective function:
\begin{equation} \label{eq:objective_ebm_nce}
\small
\begin{aligned}
\mathcal{L}_{\text{\model-NCE}}
& = - \Big( \mathbb{E}_{p_n(\mz^u_{j, \beta}|\mz^v_{i, \alpha})} \big[ \log \big(1-\sigma(f(\mz^u_{j, \beta},\mz^v_{i, \alpha}))\big)] + \mathbb{E}_{p_{\text{data}}(\mz^u_{i, \alpha}|\mz^v_{i, \alpha}))} [\log \sigma(f(\mz^u_{i, \alpha},\mz^v_{i, \alpha}))) \big] \\
& \quad\quad + \mathbb{E}_{p_n(\mz^v_{j, \beta}|\mz^u_{i, \alpha})} \big[ \log \big(1-\sigma(f(\mz^v_{j, \beta},\mz^u_{i, \alpha})))\big)] + \mathbb{E}_{p_{\text{data}}(\mz^v_{i, \alpha}|\mz^u_{i, \alpha}))} [\log \sigma(f(\mz^v_{i, \alpha},\mz^u_{i, \alpha}))) \big] \Big),
\end{aligned}
\end{equation}
where $p_{\text{data}}$ is the empirical data distribution and $p_n$ is the noise distribution (derivations are in~\Cref{app:sec:lower_bound_MI}). Recall that the latent code pairs ($\vz^u, \vz^v$) are given in advance, and the noise distribution is on the semantic directions and step sizes.
In specific, the step sizes ($\alpha \ne \beta$) are randomly sampled from [-3, 3], and the latent direction indices ($i \ne j$) are randomly sampled from \{0, 1, ..., D-1\}.
\Cref{eq:objective_ebm_nce} is for one latent code pair, and we take the expectation of it over all the pairs sampled from the dataset. Besides, we would like to consider extra similarity and sparsity constraints as:
\begin{equation} \label{eq:similarity_direction}
\small{
\begin{aligned}
\mathcal{L}_{\text{sim}} = \mathbb{E}_{i,j}[ \text{sim}( \vd_i, \vd_j)], \quad\quad \mathcal{L}_{\text{sparsity}} = \mathbb{E}_i [ \| \vd_i \| ],
\end{aligned}
}
\end{equation}
where $\text{sim}(\cdot)$ is the similarity function between two latent directions, and we use the dot product. By minimizing these two regularization terms, we can make the learned semantic directions more diverse and sparse. Putting them together, the final objective function is:
\begin{equation} \label{eq:final_objective}
\small{
\begin{aligned}
\mathcal{L} = c_1 \cdot \mathbb{E}_{u,v} [\mathcal{L}_{\text{\model-NCE}}] + c_2 \cdot \mathcal{L}_{\text{sim}} + c_3 \cdot \mathcal{L}_{\text{sparsity}},
\end{aligned}
}
\end{equation}
where $c_1, c_2, c_3$ are coefficients, and we treat them as three hyperparameters (check~\Cref{app:sec:experimental_details}). The above pipeline is illustrated in~\Cref{fig:GraphCG_pipeline}, and for the next we will discuss certain key modules.

\textbf{Latent code pairs, positive and negative views.}
We consider two options for obtaining the latent pairs. (1) \textit{Perturbation (\model-P)} is that for each data point $\vx$, we obtain its latent code $\vz = f(\vx)$. Then we apply two perturbations ({\eg}, adding Gaussian noise) on $\vz$ to get two perturbed latent codes as $\vz^u$ and $\vz^v$, respectively. (2) \textit{Random sampling (\model-R)} is that we encode two randomly sampled data points from the empirical data distribution as $\vz^u$ and $\vz^v$ respectively. Perturbation is one of the widely-used strategies~\citep{karras2019style} for data augmentation, and random sampling has been widely used in the NCE~\citep{song2021train} literature.
Then we can define the positive and negative pairs in \model-NCE, where the goal is to align the positives and contrast the negatives. As described in~\Cref{eq:objective_ebm_nce}, the positive pairs are latent pairs moving with the same semantic direction and step size, while the negative pairs are the edited latent codes with different semantic directions and/or step sizes.

\textbf{Semantic direction modeling.}
We first randomly draw a \textit{basis vector} $\ve_i$, and then model the semantic direction~$\vd_i$ as $\vd_i = \text{MLP}(\ve_i)$, where $\text{MLP}(\cdot)$ is the multi-layer perceptron network.

\textbf{Design of editing function.}
Given the semantic direction and two views, the next task is to design the editing function $h(\cdot)$ in~\Cref{eq:editing_function}. Since our proposed \model{} is flexible, and the editing function determines the energy function~\Cref{eq:energy_function}, we consider both the linear and non-linear editing functions as:\looseness=-1
\begin{equation}
\small
\mz_i = \vz + \alpha \cdot \vd_i, \quad\quad \mz_i = \vz + \alpha \cdot \vd_i + \text{MLP}(\vz \oplus \vd_i \oplus [\alpha]),
\end{equation}
where $\oplus$ is the concatenation of two vectors. Noticing that for the non-linear case, we are adding an extra term by mapping from the latent code, semantic direction, and step-size simultaneously. We expect that this could bring in more modeling expressiveness in the editing function. For more details, {\eg}, the ablation study to check the effect on the design of the views and editing functions, please refer to~\Cref{app:sec:molecular_graph_results,app:sec:point_clouds_results}, while more potential explorations are left for future work.

\subsection{Implementations Details}
During training, the goal of \model{} is to learn semantically meaningful direction vectors together with an editing function in the latent space, as in~\Cref{alg:model_learning}. Then we need to manually annotate the semantic directions concerning the corresponding factors using certain post-training evaluation metrics. Finally, for the inference phase, provided with the pretrained graph DGM and a selected semantic direction (together with a step size) learned by \model{}, we can sample a graph -> conduction editing in the latent space -> decoding to generate the edited graph, as described in~\Cref{eq:editing_function}. The detailed algorithm is illustrated in~\Cref{alg:model_inference}. Next, we highlight several key concepts in \model{} and briefly discuss the differences from other related concepts.

\begin{wraptable}[16]{r}{0.6\textwidth}
\begin{minipage}{0.6\textwidth}
\vspace{-5ex}
\begin{algorithm}[H]
\small
\caption{\small{Learning Phase of \model}} \label{alg:model_learning}
\begin{algorithmic}[1]
    \STATE {\bfseries Input:} Given a pretrained generative model encoder, $f(\cdot)$.
    \STATE {\bfseries Output:} Learned direction vector $\vd_i$ and function $h(\cdot)$.
    \STATE Select latent codes $\vz^u, \vz^v \in \mathcal{Z}$ from empirical dataset and $f(\cdot)$.
    \FOR{each step size $\alpha$ and each direction $i$}
        \STATE Set $\mz^u_{i, \alpha} = h(\vz^u, \vd_i, \alpha)$.
        \STATE Set $\mz^v_{i, \alpha} = h(\vz^v, \vd_i, \alpha)$.
        \STATE Assign positive to pair $(\mz^u_{i, \alpha}, \mz^v_{i, \alpha})$.
            \FOR{step size $\beta \ne \alpha$ and direction $j \ne i$}
                \STATE Set $\mz^u_{j, \beta} = h(\vz^u, \vd_j, \beta)$.
                \STATE Set $\mz^v_{j, \beta} = h(\vz^v, \vd_j, \beta)$.
                \STATE Assign negative to pair $(\mz^u_{i, \alpha}, \mz^v_{j, \beta})$.
                \STATE Assign negative to pair $(\mz^u_{j, \beta}, \mz^v_{i, \alpha})$.
            \ENDFOR
        \STATE Do SGD w.r.t. \model{} in~\Cref{eq:final_objective}.
    \ENDFOR
\end{algorithmic}
\end{algorithm}
\end{minipage}
\end{wraptable}
\textbf{NCE and contrastive representation learning.}
\model-NCE is applying EBM-NCE, which is essentially a contrastive learning method, and another dominant contrastive loss is the InfoNCE~\citep{oord2018representation}. We summarize their relations below. (1) Both contrastive methods are doing the same thing: align the positive pairs and contrast the negative pairs. (2) EBM-NCE~\citep{hassani2020contrastive,liu2021pre} has been found to outperform InfoNCE on certain graph applications like representation learning. (3) What we want to propose here is a flexible framework. Specifically, EBM provides a more general framework by designing the energy functions, and EBM-NCE is just one effective solution. Other promising directions include the denoising score matching or denoising diffusion model~\citep{song2020score}, while InfoNCE lacks such a nice extensibility attribute.

\begin{wraptable}[12]{r}{0.6\textwidth}
\begin{minipage}{0.6\textwidth}
\vspace{-5.5ex}
\begin{algorithm}[H]
\caption{\small{Inference Phase of \model{}}} \label{alg:model_inference}
\begin{algorithmic}[1]
    \STATE {\bfseries Input:} Given a pre-trained generative model, $f(\cdot)$ and $g(\cdot)$, a learned direction vector $\vd$.
    \STATE {\bfseries Output:} A sequence of edited graphs.
    \STATE Sample $\vz$ with DGM or $\vx$ from a graph dataset.
    \STATE If the latter, get a latent code $\vz = f(x)$.
    \FOR{step size $\alpha \in [-3, 3]$}
        \STATE Do graph edit in the latent space to get $\bar \vz_{i,\alpha} = h(\vz, \vd, \alpha)$.
        \STATE Decode to the graph space with $\bar \vx' = g(\bar \vz_{i,\alpha})$.
    \ENDFOR
    \STATE Output is thus a sequence of edited graphs, $\{\bar \vx'\}$.
\end{algorithmic}
\end{algorithm}
\end{minipage}
\end{wraptable}
\textbf{\model{} and contrastive self-supervised learning (SSL).}
\model{} shares certain similarities with the self-supervised learning (SSL) method, however, there are some inherent differences, as summarized below. (1) SSL aims at learning the data representation by operating data augmentation on the data space, such as node addition and edge deletion. \model{} aims at learning the semantically meaningful directions by editing on the latent space (the representation function is pretrained and fixed). (2) Based on the first point, SSL aims at using different data points as the negative samples. \model{}, on the other hand, is using different directions and step-sizes as negatives. Namely, SSL is learning data representation in the inter-data level, and \model{} is learning the semantic directions in the inter-direction level.

\textbf{Output sequence in the discrete space.}
Recall that during inference time (\Cref{alg:model_inference}), \model{} takes a DGM and the learned semantic direction to output a sequence of edited graphs. Compared to the vision domain, where certain models~\citep{shen2021closed,shen2020interfacegan} have proven their effectiveness in many tasks, the backbone models in the graph domain have limited discussions. This is challenging because the graph data is in a discrete and structured space, and the evaluation of such space is non-trivial. Meanwhile, \model{} essentially provides another way to verify the quality of graph DGMs. \model{} paves the way for this potential direction, and we would like to leave this for future exploration.

\section{Experiments} \label{sec:experiments}
In this section, we show both the qualitative and quantitative results of \model{}, on two types of graph data: molecular graphs and point clouds. Due to the page limit, We put the experiment and implementation details in~\Cref{app:sec:experimental_details}.

\begin{figure}[b]
\fontsize{8.5}{4}\selectfont
\centering
\vspace{-2ex}
\begin{minipage}[b]{.4\linewidth}
    \raisebox{-1.05\height}
        {\includegraphics[width=\linewidth]{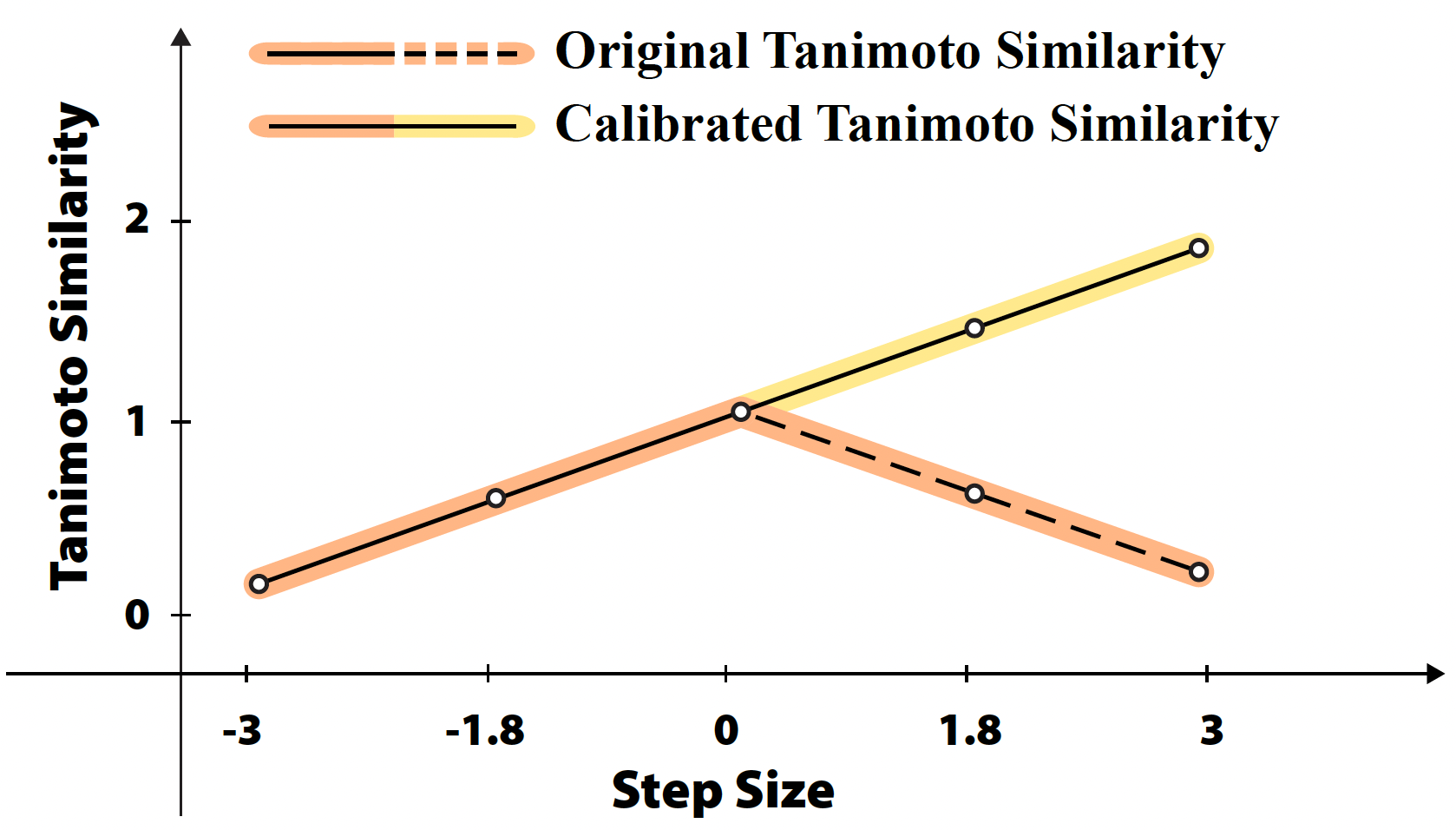}
    }
\end{minipage}
\hfill
\begin{minipage}[t]{.58\linewidth}
    { \fontsize{7.6}{1}\selectfont
    \begin{align}
    & \phi_{\text{SMR}}\big( \{ s(\bar \vx') \}_i^m, \gamma, \tau \big) =
        \begin{cases}
        1, & \text{len}\big(\text{set}\big(\{ s(\bar \vx') \}_i^m\big)\big) \ge \gamma  \\
         & \quad \wedge \text{ monotonic}_\tau \big(\{ s(\bar \vx') \}_i^m\big)\\
        0, & \text{otherwise}
        \end{cases},
    \label{eq:SMR_each_sequence}\\
    & \phi_{\text{SMR}}(\gamma, \tau )_i = \frac{1}{M} \sum_{m=1}^{M} \phi_{\text{SMR}}\big( \{ s(\bar \vx') \}_i^m, \gamma, \tau \big),
    \label{eq:SMR_each_direction} \\
    & \text{top-K}(\gamma, \tau) = \frac{1}{K} \sum_{i \in \text{top-K directions}} \Big( \phi_{\text{SMR}}(\gamma, \tau)_i \Big).
    \label{eq:top_K_SMR}
    \end{align}
    }
\end{minipage}
\vspace{-1ex}
\caption{\small
This illustrates the sequence monotonic ratio (SMR) on calibrated Tanimoto similarity (CTS). \Cref{eq:SMR_each_sequence,eq:SMR_each_direction} are the SMR on each sequence and each direction respectively, where $M$ is the number of generated sequences for the $i$-th direction and $\{ s(\bar \vx') \}_i^m$ is the CTS of the $m$-th generated sequence with the $i$-th direction. \Cref{eq:top_K_SMR} is the average of top-K SMR on $D$ directions.
}
\label{fig:sequence_monotonic_ratio_tanimoto}
\end{figure}

\subsection{Graph Data: Molecular Graphs}
\textbf{Background of molecular graphs.} A molecule can be naturally treated as a graph, where the atoms and bonds are nodes and edges, respectively. The unsupervised graph editing tasks can thus be formulated as editing the substructures of molecular graphs. In practice, people are interested in substructures that are critical components of molecules, which are called the `fragments`. In recent years, graph representation learning has been extensively explored on the molecular graph~\citep{duvenaud2015convolutional,gilmer2017neural,liu2018n,yang2019analyzing,corso2020principal}.

\textbf{Backbone DGMs.}
We consider two state-of-the-art DGMs for molecular graph generation. MoFlow~\citep{zang2020moflow} is a flow-based generative model on molecules that adopts an invertible mapping between the input molecular graphs and a latent prior. HierVAE~\citep{jin2020hierarchical} is a hierarchical VAE model that encodes and decodes molecule atoms and motifs in a hierarchical manner. Besides, the pretrained checkpoints are also provided, on ZINC250K~\citep{irwin2005zinc} and ChEMBL~\citep{mendez2019chembl}, respectively.\looseness=-1

\textbf{Editing sequences and anchor molecule.}
As discussed in~\Cref{sec:method}, the output of the inference in \model{}, is a sequence of edited molecules with the $i$-th semantic direction, $\{\bar \vx'\}_i$. We first randomly generate a molecule using the backbone DGMs (without the editing operation), and we name such molecule as the anchor molecule, $\bar \vx^*$. Then we take 21 step sizes from -3 to 3, with interval 0.3, to obtain a sequence of 21 molecules following~\Cref{eq:editing_function}. Note that the edited molecule with step size 0 under the linear editing function is the same as the anchor molecule, {\ie}, $\bar \vx^*$.

\textbf{Change of structure factors and evaluation metrics.}
We are interested in the change of the graph structure (the steerable factors) along the output sequence edited with the $i$-th semantic direction. To evaluate the structure change, we apply the Tanimoto similarity between each output molecule and the anchor molecule. Besides, for the ease of evaluating the monotonicity, we apply a Tanimoto similarity transformation on the output molecules with positive step sizes by taking the deduction from 2. We call this calibrated Tanimoto similarity (CTS) sequence, marked as $\{ s(\bar \vx') \}_i$. An illustration is shown in~\Cref{fig:sequence_monotonic_ratio_tanimoto}. Further, we propose a metric called Sequence Monotonic Ratio (SMR), $\phi_{\text{SMR}}(\gamma, \tau)_i$, which measures the monotonic ratio of $M$ generated sequences edited with the $i$-th direction. It has two arguments: the diversity threshold $\gamma$ constrains the minimum number of distinct molecules, and the tolerance threshold $\tau$ controls the non-monotonic tolerance ratio along each sequence.

\textbf{Evaluating the diversity of semantic directions.}
SMR can evaluate the monotonic ratio of output sequences generated by one direction. To better illustrate that \model{} is able to learn multiple directions with diverse semantic information, we also consider taking the average of top-$K$ SMR to reveal that all the best $K$ directions are semantically meaningful, as in~\Cref{eq:top_K_SMR}.

\textbf{Baselines.}
For baselines, we consider four unsupervised editing methods. (1) The first is Random. It randomly samples a normalized random vector in the latent space as the semantic direction. (2) The second one is Variance. We analyze the variance on each dimension of the latent space, and select the highest one with one-hot encoding as the semantic direction. (3) The third one is SeFa~\citep{shen2021closed}. It first decomposes the latent space into lower dimensions using PCA, and then takes the most principal components (eigenvectors) as the semantic-rich direction vectors. (4) The last one is DisCo~\citep{ren2021generative}. It maps each latent code back to the data space, followed by an encoder for contrastive learning, so it requires the backbone DGMs to be end-to-end and is infeasible for HierVAE.

\textbf{Quantitative results.}
We take $D=10$ to train \model, and the optimal results on 100 sampled sequences are reported in~\Cref{tab:main_results_molecules}. We can observe that \model{}, can show consistently better structure change with both top-1 and top-3 directions. This can empirically prove the effectiveness of our proposed \model. More comprehensive results are in~\Cref{app:sec:molecular_graph_results}.

\begin{table}[tb!]
\begin{center}
\caption{
\small
This table lists the sequence monotonic ratio (SMR, \%) on calibrated Tanimoto similarity (CTS) for the top-1 and top-3 directions. The best performances are marked in \textbf{bold}.
}
\label{tab:main_results_molecules}
\small
\setlength{\tabcolsep}{10pt}
\vspace{-2ex}
\begin{adjustbox}{max width=\textwidth}
\begin{tabular}{lll cccccccc}
\toprule
\multirow{3}{*}{\makecell{\textbf{Model}}} & \multirow{3}{*}{\makecell{\textbf{Dataset}}} &  & \multicolumn{4}{c}{\textbf{Tanimoto top-1}} & \multicolumn{4}{c}{\textbf{Tanimoto top-3}}\\
\cmidrule(lr){4-7} \cmidrule(lr){8-11}
 & & \multicolumn{1}{c}{diversity $\gamma$} & \multicolumn{2}{c}{3} & \multicolumn{2}{c}{4} & \multicolumn{2}{c}{3} & \multicolumn{2}{c}{4}\\
\cmidrule(lr){4-5} \cmidrule(lr){6-7} \cmidrule(lr){8-9} \cmidrule(lr){10-11}
 & & \multicolumn{1}{c}{tolerance $\tau$} & 0 & 0.2 & 0 & 0.2 & 0 & 0.2 & 0 & 0.2\\
\midrule
\multirow{7}{*}{\makecell{MoFlow}} & \multirow{7}{*}{\makecell{ZINC250k}}
& Random & 23.0 & 25.0 & 12.0 & 15.0 & 22.0 & 24.0 & 11.0 & 13.7\\
& & Variance & 24.0 & 28.0 & 12.0 & 16.0 & 20.0 & 25.0 & 10.0 & 15.0\\
& & SeFa~\cite{shen2021closed} & 4.0 & 4.0 & 0.0 & 0.0 & 3.3 & 3.3 & 0.0 & 0.0\\
& & DisCo~\cite{ren2021generative} & 7.0 & 14.0 & 2.0 & 8.0 & 5.3 & 11.7 & 2.0 & 7.7\\
\cmidrule(lr){3-11}
 & & \model-P & \textbf{32.0} & \textbf{34.0} & \textbf{16.0} & \textbf{18.0} & \textbf{29.0} & \textbf{31.0} & \textbf{13.7} & \textbf{16.3}\\
 & & \model-R & 25.0 & 26.0 & 11.0 & 14.0 & 22.0 & 24.3 & 10.3 & 13.3\\
\midrule
\multirow{6}{*}{\makecell{HierVAE}} & \multirow{6}{*}{\makecell{ChEMBL}}
& Random & 14.0 & 45.0 & 14.0 & 43.0 & 10.0 & 42.3 & 9.3 & 41.7\\
& & Variance & 23.0 & 59.0 & 19.0 & 55.0 & 18.3 & 52.7 & 15.7 & 50.3\\
& & SeFa~\cite{shen2021closed} & 4.0 & 41.0 & 4.0 & 41.0 & 2.3 & 36.0 & 2.3 & 36.0\\
\cmidrule(lr){3-11}
 & & \model-P & 40.0 & \textbf{73.0} & \textbf{32.0} & \textbf{65.0} & 36.0 & \textbf{64.3} & 26.3 & \textbf{57.7}\\
 & & \model-R & \textbf{42.0} & 67.0 & 30.0 & 55.0 & \textbf{38.0} & 62.3 & \textbf{28.7} & 53.7\\
\bottomrule
\end{tabular}
\end{adjustbox}
\end{center}
\vspace{-2ex}
\end{table}

\begin{figure}[tb!]
    \begin{subfigure}[\small Steerable factor: number of halogens.]
    {\includegraphics[width=0.47\linewidth]{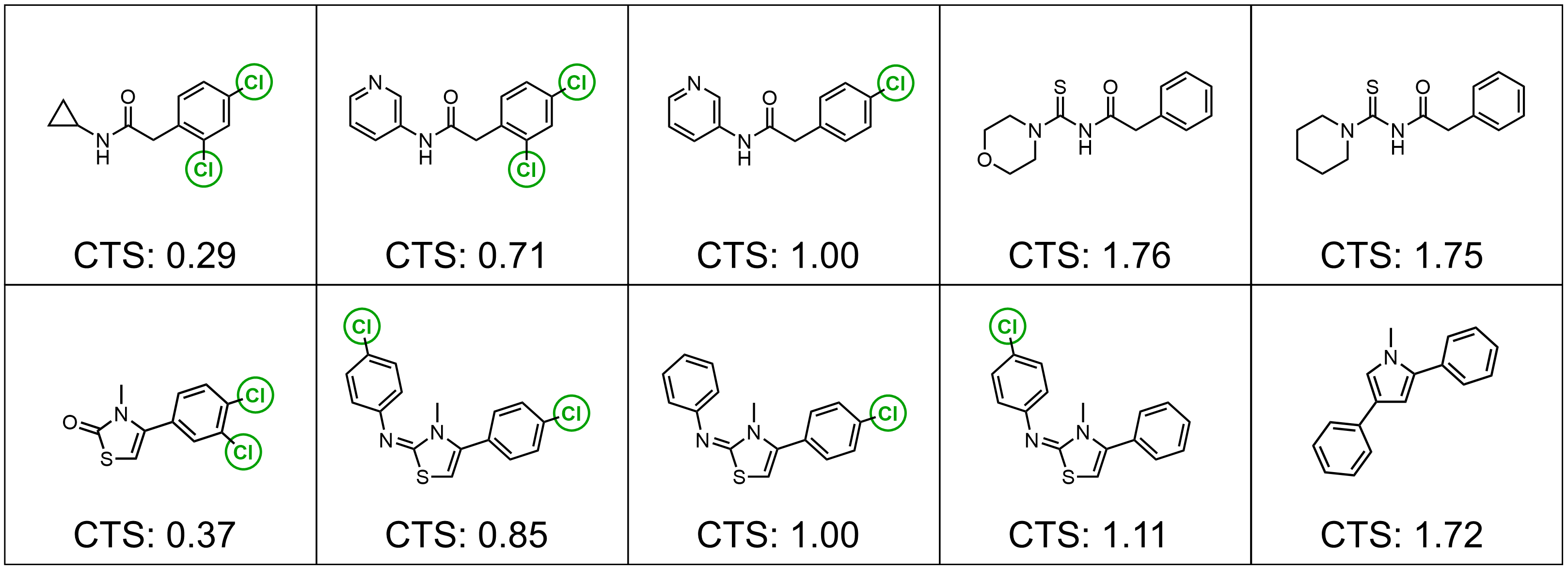}
    \label{fig:path_generation_example_hiervae_main_a}
    }
    \end{subfigure}
\hfill
     \begin{subfigure}[\small Steerable factor: number of hydroxyls.]
     {\includegraphics[width=0.47\linewidth]{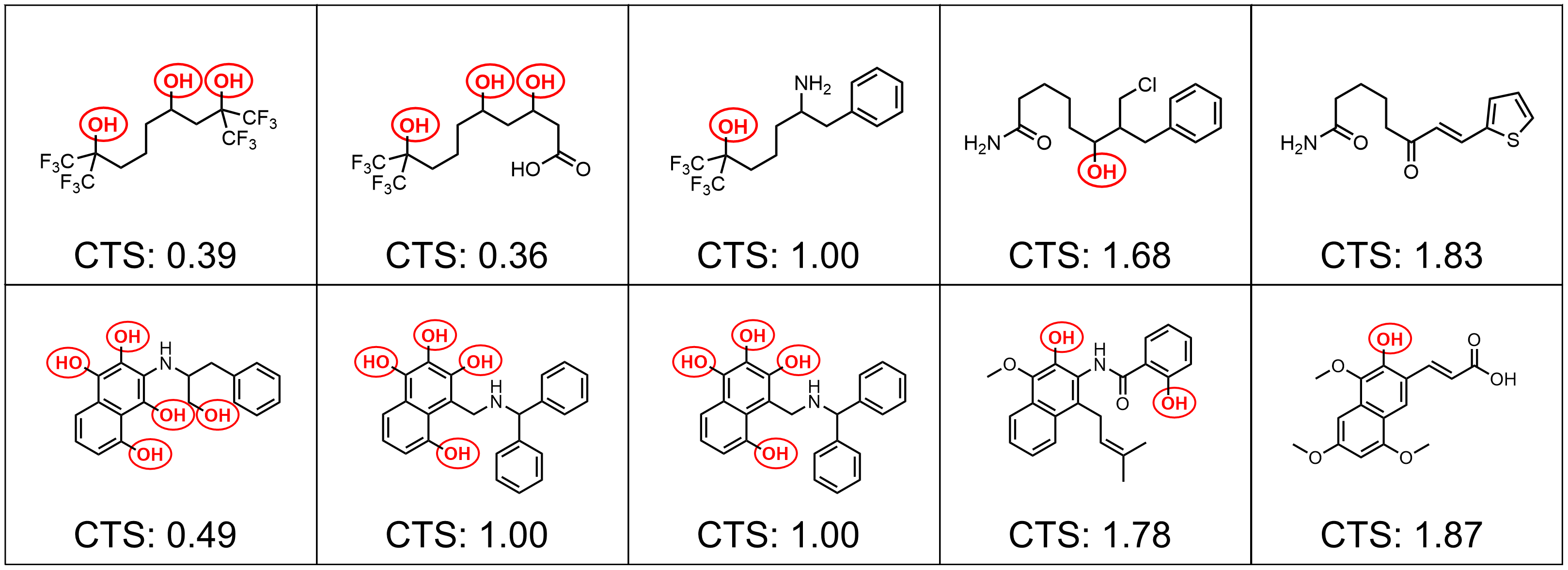}
     \label{fig:path_generation_example_hiervae_main_b}
     }
     \end{subfigure}
\\
\begin{subfigure}[\small Steerable factor: number of amides.]
    {\includegraphics[width=0.47\linewidth]{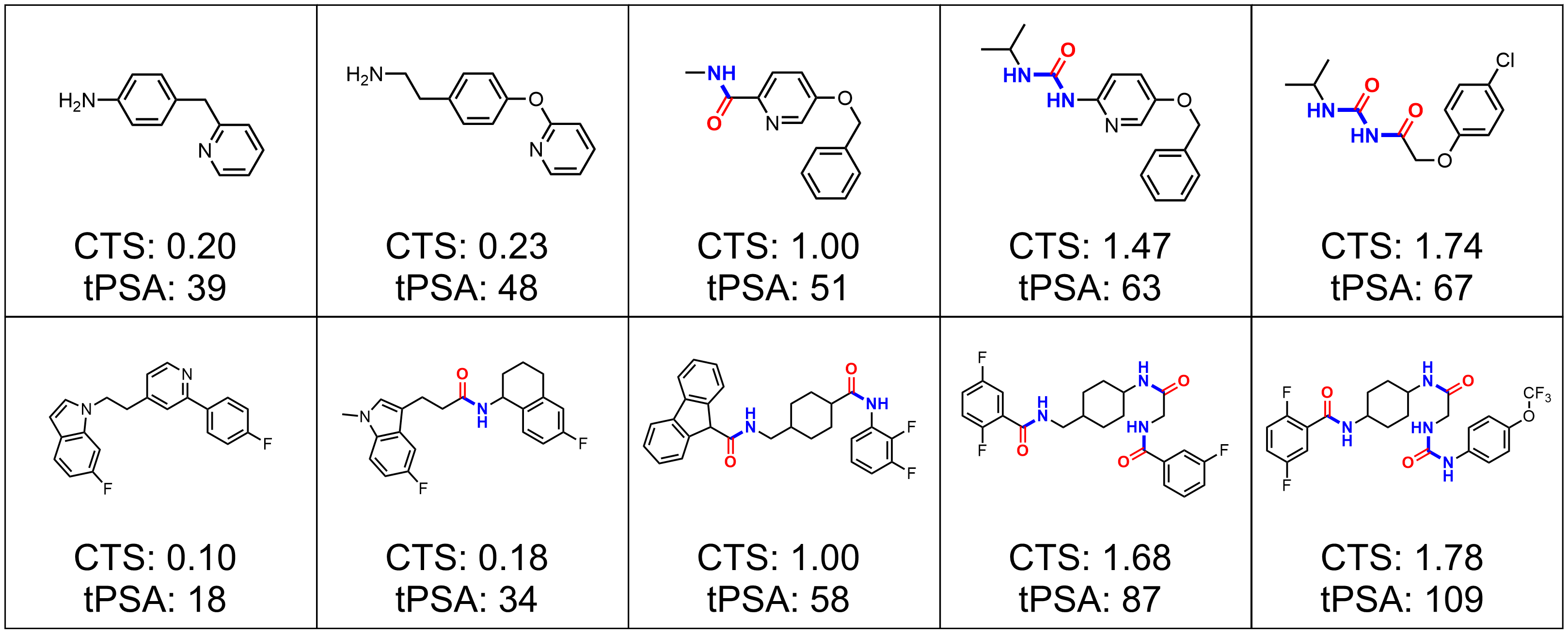}
    \label{fig:path_generation_example_hiervae_main_c}
    }
    \end{subfigure}
\hfill
\begin{subfigure}[\small Steerable factor: chain length.]
    {\includegraphics[width=0.47\linewidth]{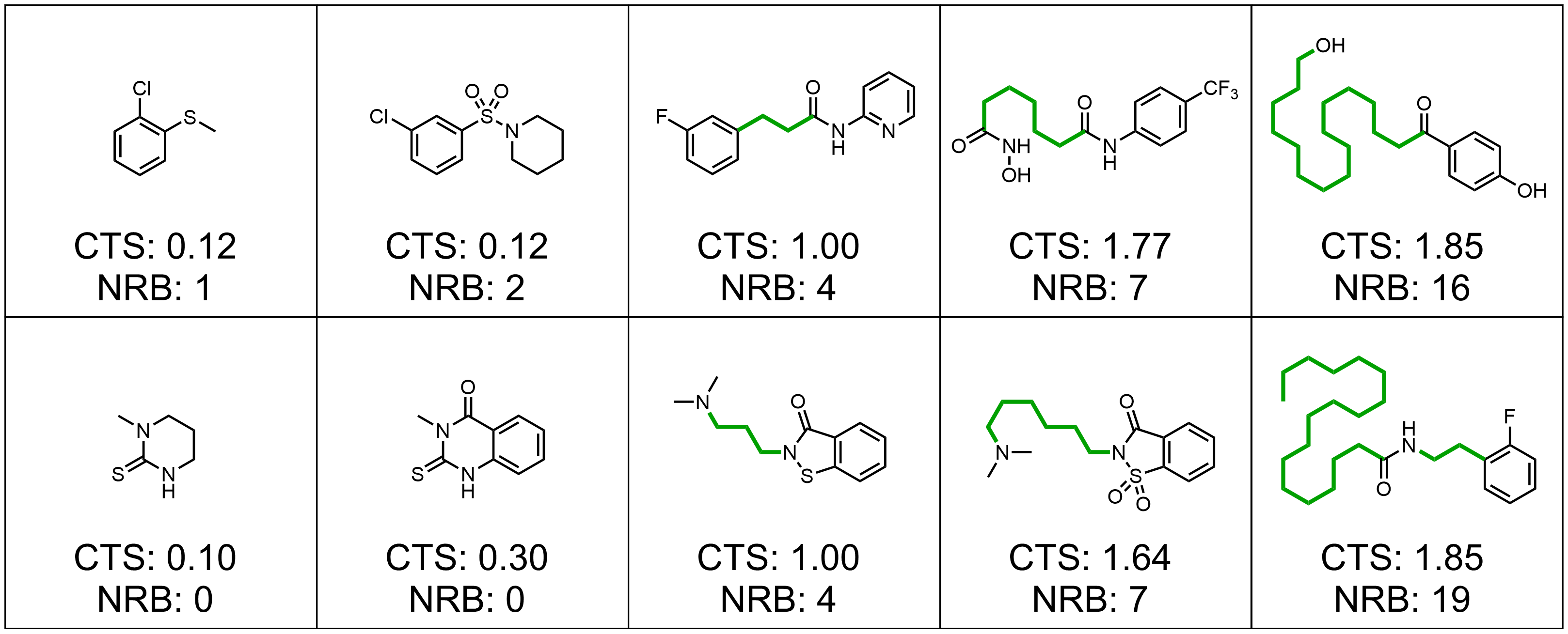}
    \label{fig:path_generation_example_hiervae_main_d}
    }
    \end{subfigure}
\\
\vspace{-2ex}
\caption{
\small
\model{} for molecular graph editing. We visualize the output molecules and CTS in four directions with two sequences each, where each sequence consists of five steps. The five steps correspond to five step sizes: -3, -1.8, 0, 1.8, and 3, where 0 marks the anchor molecule (center point of reach sequence). \Cref{fig:path_generation_example_hiervae_main_a} visualizes how the number of halogens (marked in green) decreses from -3 to 3. \Cref{fig:path_generation_example_hiervae_main_b} visualizes how the number of hydroxyls (marked in red) decreases from -3 to 3. \Cref{fig:path_generation_example_hiervae_main_c} visualizes how the number of amides (marked in red and blue) increases from -3 to 3. \Cref{fig:path_generation_example_hiervae_main_d} visualizes how the number of chains (marked in green) increases from -3 to 3. Notably, certain properties change with molecular structures accordingly, like topological polar surface area (tPSA) and the number of rotatable bonds (NRB).
}
\vspace{-2ex}
\label{fig:path_generation_molecule}
\end{figure}

\textbf{Analysis on steerable factors in molecules: functional group change.}
For visualization, we sample 8 molecular graph sequences along 4 selected directions in~\Cref{fig:path_generation_molecule}, and the backbone DGM is HierVAE pretrained on ChEMBL. The CTS holds a good monotonic trend, and each direction shows certain unique changes in the molecular structures, {\ie}, the steerable factors in molecules. Some structural changes are reflected in molecular properties. We expand all the details below. In~\Cref{fig:path_generation_example_hiervae_main_a,fig:path_generation_example_hiervae_main_b}, the number of halogen atoms and hydroxyl groups (in alcohols and phenols) in the molecules decrease from left to right, respectively. In~\Cref{fig:path_generation_example_hiervae_main_c}, the number of amides in the molecules increases along the path. Because amides are polar functional groups, the topological polar surface area (tPSA) of the molecules also increases, which is a key molecular property for the prediction of drug transport properties, {\eg}, permeability~\citep{ertl2000fast}. In~\Cref{fig:path_generation_example_hiervae_main_d}, the flexible chain length, marked by the number of ethylene (CH$_2$CH$_2$) units, increases from left to right. Since the number of rotatable bonds (NRB) measures the molecular flexibility, it also increases accordingly~\citep{veber2002molecular}.
\vspace{-1ex}

\begin{figure}[tb!]
\centering
    \begin{subfigure}[Steerable factor: engine.]
    {\includegraphics[width=0.48\linewidth]{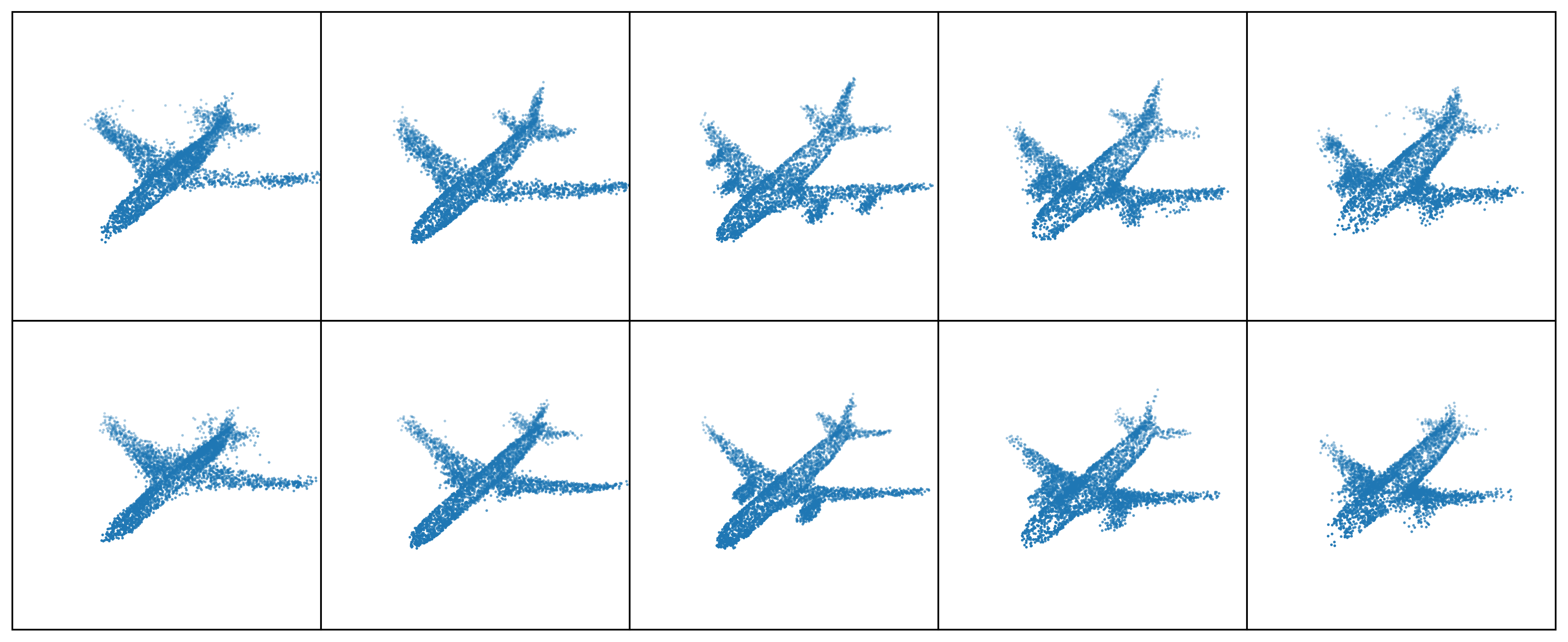}
    \label{fig:airplane_data_2_dir_14_engine_removal}}
    \end{subfigure}
\hfill
    \begin{subfigure}[Steerable factor: engine.]
    {\includegraphics[width=0.48\linewidth]{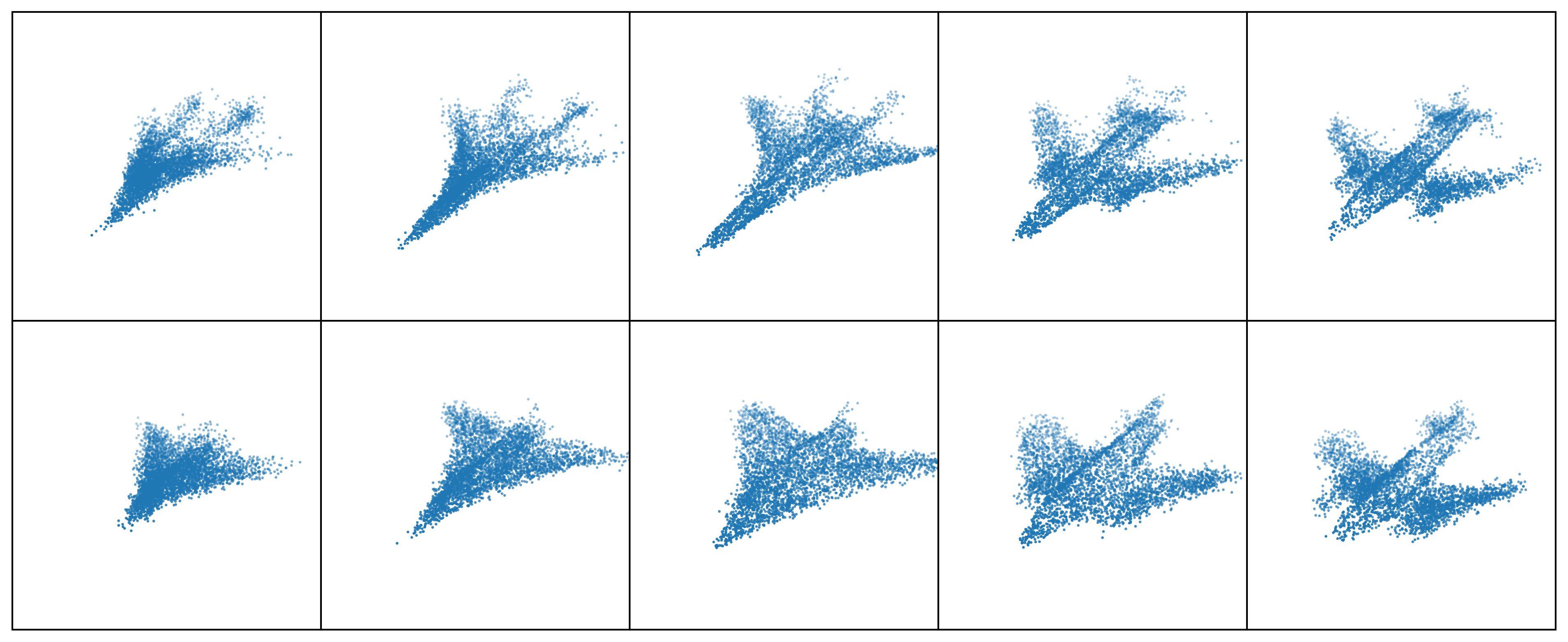}
    \label{fig:airplane_data_2_dir_14_engine_add}}
    \end{subfigure}
\\
\vspace{-2ex}
    \begin{subfigure}[Steerable factor: size.]
    {\includegraphics[width=0.48\linewidth]{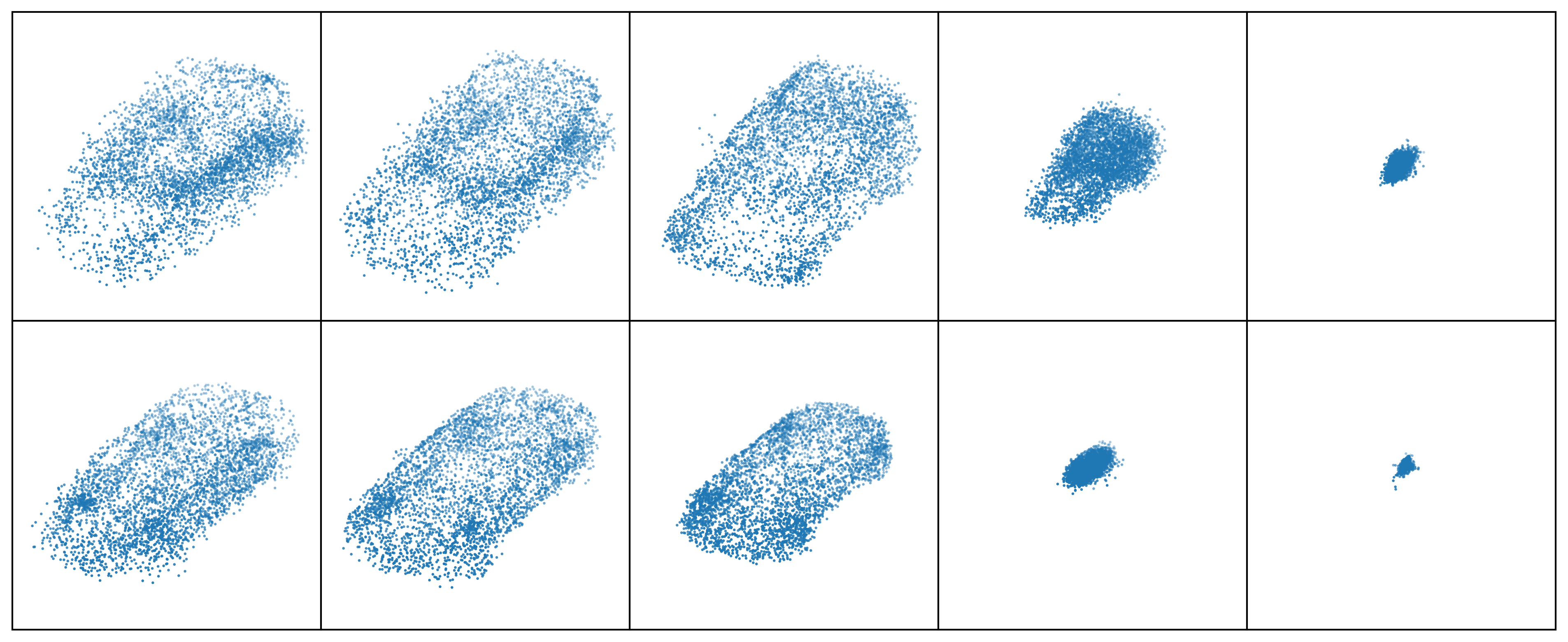}
    \label{fig:car_data_2_dir_8}}
    \end{subfigure}
\hfill
    \begin{subfigure}[Steerable factor: leg height.]
    {\includegraphics[width=0.48\linewidth]{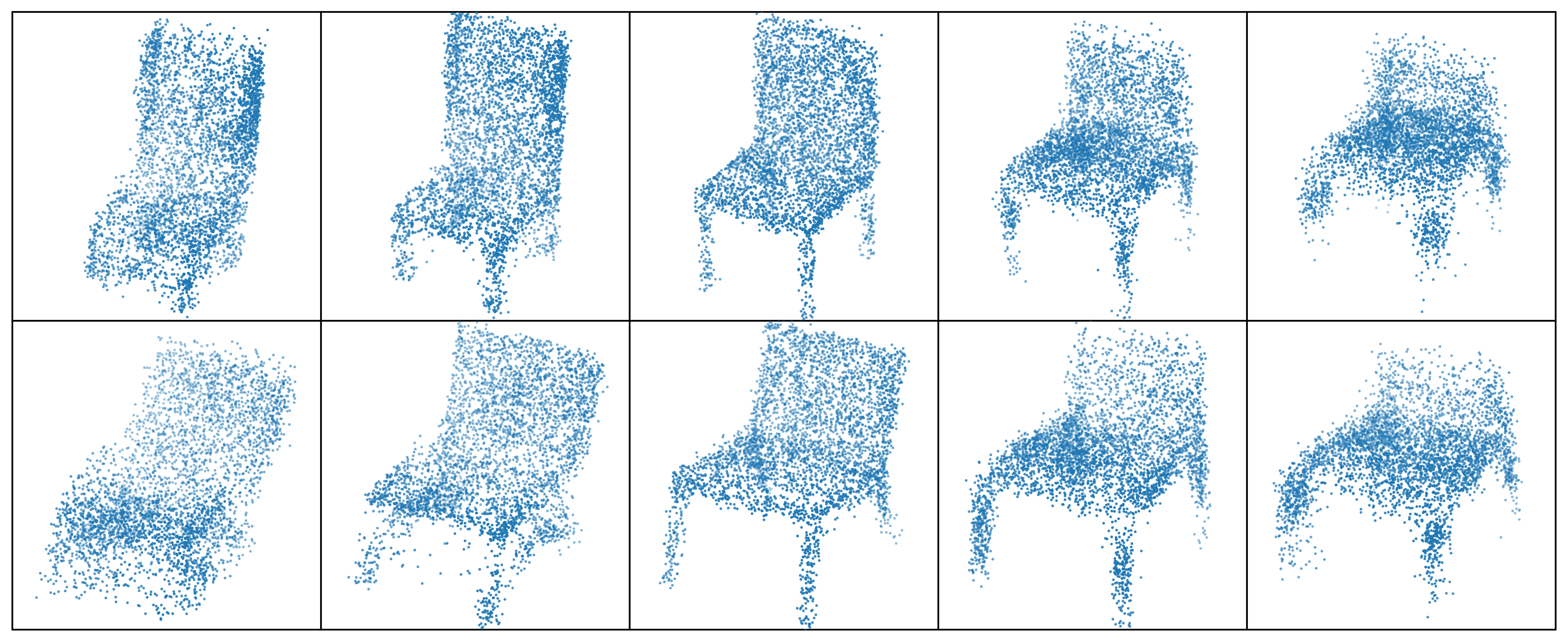}
    \label{fig:chair_data_2_dir_8}}
    \end{subfigure}
\\
\vspace{-2ex}
\caption{
\small
\model{} for point clouds editing. We show four editing sequences, where each sequence consists of five point clouds, and the center one is the anchor point clouds, {\ie}, with step size 0. The other four point clouds correspond to step size with -3, -1.8, 1.8, and 3, respectively. \Cref{fig:airplane_data_2_dir_14_engine_removal} and~\Cref{fig:airplane_data_2_dir_14_engine_add} refer the same semantic direction, and they are showing how to steer the factor engine: the number of engines will be decreased and increased with the negative (left) and positive (right) step size respectively. Similarly, \Cref{fig:car_data_2_dir_8,fig:chair_data_2_dir_8} illustrate the effect of the steerable factors on the car size and the chair leg height.
}
\label{fig:point_clouds_main}
\vspace{-3ex}
\end{figure}

\subsection{Graph Data: Point Clouds}

\textbf{Background of point clouds.}
A point cloud is represented as a set of points, where each point $P_{i}$ is described by a vector of 3D Euclidean coordinates possibly with extra channels ({\eg}, colors, surface normals, and returned laser intensity). In 2017, Qi et. al~\citep{qi2017pointnet} designed a deep learning framework called PointNet that directly consumes unordered point sets as inputs and can be used for various tasks such as classification and segmentation.
 
\textbf{Backbone DGMs.} We consider one of the latest DGMs on point clouds, PointFlow~\citep{yang2019pointflow}. It is using the normalizing flow model for estimating the 3D point cloud distribution. Then we consider PointFlow pretrained on three datasets in ShapeNet~\citep{chang2015shapenet}: Airplane, Car, and Chair. All point clouds are obtained by sampling points uniformly from the mesh surface.

\textbf{Analysis on steerable factors in point clouds: shape change.}
To train \model, we take $D=10$ directions, and we sample 8 point cloud sequences along 3 directions for visualization in~\Cref{fig:point_clouds_main}. More results are in~\Cref{app:sec:point_clouds_results}. It is observed that \model{}, can steer the shape of the point clouds, {\eg}, the size of cars and the height of chair legs. We also find it interesting that \model{}, can steer more finger-trained factors, like modifying the number of jet engines.
\section{Conclusion and Discussion}
\vspace{-1ex}

In this work, we are interested in unsupervised graph editing. It is a well-motivated task for various real-world applications, and we discuss two mainstream data types: molecular graphs and point clouds. We start with exploring the latent space of mainstream deep generative models and propose \model{}, a model-agnostic unsupervised method for graph data editing. The key component of \model{}, is EBM, and we take the \model{}-NCE as the solution for now. For future work, we may as well extend it to more advanced solutions like denoising diffusion model~\citep{ho2020denoising}.

One limitation of \model{}, (as well as the solutions to general unsupervised data editing)~\citep{harkonen2020ganspace,shen2021closed,ren2021generative} is that we may need some post-training human selection (as shown in~\Cref{alg:model_inference}) to select the most promising semantic vectors to steer factors. Another issue is the lack of open-sourced evaluation metrics. This requires both a deep understanding of the representation space of deep generative models and domain knowledge of the problem. For instance, activity cliff is a challenging task~\citep{hu2012extending} for editing, while current measures fail to capture it. To set up constructive evaluation metrics can help augment our understandings from both the domain and technique perspectives. This is beyond the scope of our work, yet would be interesting to explore as a future direction.

\section*{Code and Data Availability}
The codes and data download scripts are available at this \href{https://github.com/chao1224/GraphCG}{GitHub repository}.

\bibliography{reference}
\bibliographystyle{tmlr}

\newpage
\appendix

\appendix

\section*{Reproducibility Statement}
\vspace{-1ex}
To ensure the reproducibility of the empirical results, we provide the implementation details (hyperparameters, dataset specifications, pretrained checkpoints, etc.) in~\Cref{sec:experiments,app:sec:toy_example,app:sec:experimental_details,app:sec:molecular_graph_results,app:sec:point_clouds_results}, and the source code will be released in the future.
Besides, the complete derivations of equations and clear explanations are given in~\Cref{sec:method,app:sec:lower_bound_MI}.

Specifically, we provide the details for reproducing the results:
\begin{itemize}[noitemsep,topsep=0pt]
    \item In~\Cref{app:tab:SMR_CTS_MoFlow}, GraphCG-P with~\Cref{app:eq:non_linear} and GraphCG-R with \Cref{app:eq:linear_01} are reported in~\Cref{tab:main_results_molecules}.
    \item In~\Cref{app:tab:SMR_CTS_HierVAE}, GraphCG-P with \Cref{app:eq:linear_02} and GraphCG-R with \Cref{app:eq:linear_02} are reported in~\Cref{tab:main_results_molecules}.
\end{itemize}
For the visualization in~\Cref{fig:path_generation_molecule}, we take the GraphCG-P with \Cref{app:eq:linear_02}, and the backbone generative model is HierVAE pretrained on ChEMBL. Further, we provide an anonymous link \href{https://anonymous.4open.science/r/GraphCG_outputs-075A}{here}. In these CSV files:
\begin{itemize}[noitemsep,topsep=0pt]
    \item Direction 0 is the halogen fragment (data 4, 71).
    \item Direction 5 is the amide fragment (data 95, 61).
    \item Direction 6 is the chain length (data 57, 14).
    \item Direction 7 is the alcohol and phenol fragments (data 10, 8).
\end{itemize}
For the visualization in~\Cref{fig:point_clouds_main}, we take the GraphCG-R with~\Cref{app:eq:linear_02} on PointFlow, for all three datasets.

\section*{Scope of GraphCG: Why Not Image}
In the current main story, we do not include \model{} for image editing. We would like to highlight that image editing is indeed a different story in terms of editing with pretrained DGMs, and the details reasons are as follows:
\begin{itemize}[noitemsep,topsep=0pt]
    \item We want to clarify that graph is a very general data structure, including the molecular graph and point clouds.
    \item In this sense, the image is a very special case of graph/structured data. It has a very tight spatial correlation, which is in a rigid form. Feel free to check~\citep{han2022vision,corso2020principal} for recent explorations on GNNs on images. Such a nice spatial correlation enables disentangled generative model. For example, in vision, StyleGAN~\citep{karras2019style} has proven with nice disentanglement property~\citep{collins2020editing,shen2020interpreting,harkonen2020ganspace,tewari2020stylerig,wu2021stylespace}.
    \item But possessing such a nice generative model is not the case for general graph data. I.e., we typically don’t have such a nice disentangled generative model, as discussed in~\Cref{sec:intro,sec:entangled_representation}.
    \begin{itemize}[noitemsep,topsep=0pt]
        \item If the pretrained graph DGMs have the nice disentanglement property, like StyleGAN, then the unsupervised steerable exploration can be quite straightforward, like SeFa. SeFa is the SOTA unsupervised editing method on the image, and it only needs a simple PCA operation.
        \item But, in reality, and in the general setting, the generative models fail to provide nice properties like disentanglement (for the general graph data). Then we need GraphCG. It is a general framework for less ordered graph data (compared to images).
        \item Further, when we compare SeFa with GraphCG, we can observe that GraphCG is better by a large margin. This also verifies that the general graph controllable generation is more challenging.
    \end{itemize}
    \item So to get back to this question, GraphCG is indeed a general graph-specific method, aiming at exploring the steerable factors in the entangled generative model setting. And obviously, the image generation task is not included in the current draft because the SOTA image generative model is good enough (well-disentangled~\citep{collins2020editing,shen2020interpreting,harkonen2020ganspace,tewari2020stylerig,wu2021stylespace}) with simple yet efficient methods like SeFa.
\end{itemize}


\section{Graph Data}
This section will be merged with Section 5 in the final version.

\subsection{Molecules} \label{app:sec:structured_data_molecules}
Molecules can be naturally represented as the 2D molecular graphs, where atoms and bonds are nodes and edges in the graph, respectively. For the recent years, graph representation learning has been extensively explored on the molecular graph~\citep{duvenaud2015convolutional,gilmer2017neural,liu2018n,yang2019analyzing,corso2020principal}. Based on the graph representation of molecules, a variety of work have been done for molecule generation. The state-of-the-art ones include MoFlow~\citep{zang2020moflow} and HierVAE~\citep{jin2020hierarchical}.

\subsection{Point Clouds} \label{app:sec:structured_data_point_cloud}
A point cloud is represented as a set of points, where each point $P_{i}$ is described by a vector of 3D Euclidean coordinates possibly with extra channels (e.g., colors, surface normals and returned laser intensity). In 2017, Qi et. al~\citep{qi2017pointnet} designed a deep learning framework called PointNet that directly consumes unordered point sets as inputs and can be used for various tasks such as classification and segmentation. For the generative models on point clouds, we consider one of the latest work, PointFlow~\citep{yang2019pointflow}.

\section{Related Work} \label{sec:app:related_work}
\paragraph{Image editing and image controllable generation}
Many existing works on controllable generation with DGMs mainly focus on image data. 
With the assumption that the learned latent space already includes rich semantic information~\citep{karras2019style,jahanian2019steerability,shen2020interfacegan,harkonen2020ganspace}, the question then becomes how to identify semantic-rich directions from the latent space of DGMs. 
Depending on whether or using supervised signals to discover the semantic directions, existing works can be divided into two settings: \textit{supervised} and \textit{unsupervised}.

The \textit{supervised} setting relies on the supervised signals to learn the pre-defined semantic-rich directions. For instance, InterfaceGAN~\citep{shen2020interfacegan} identifies the semantic directions in the latent space via linear models that recognize semantic boundaries among data. LACE~\citep{nie2021controllable} uses energy-based models to learn the joint distribution of data and properties ({\ie}, semantics) and formulate the sampling process as to solve an ordinary differential equation. 

As supervised signals usually require domain knowledge and laborious annotations, the latest works are more focused on the \textit{unsupervised} setting, either model-specific or data-specific. Specifically, GANSpace~\citep{harkonen2020ganspace} applies PCA on the intermediate layers of GANs (instead of latent space) for learning the semantic-rich directions. SeFa~\citep{shen2021closed} exploits the pretrained GANs and extracts the semantic-rich directions by using PCA on the backbone layers. Nevertheless, as both methods are specifically designed for StyleGAN~\citep{karras2019style}, it is nontrivial to generalize them to other DGMs. A more recent work DisCo~\citep{ren2021generative} employs a different pipeline: it trains a new encoder after reconstruction and maps the generated images to another latent space for editing. However, training a new encoder introduces extra complexities.

\paragraph{Graph editing and graph controllable generation}
This is an emerging field with many downstream applications~\citep{drews2000drug, shi2020point}. However, existing works are mainly focusing on the supervised setting. For example, conventional approaches, such as genetic algorithms (GAs), edit the molecule graphs in the data space via hand-crafted heuristics with the guidance of molecular property predictors. More recent learning-based methods perform the latent direction discovery, either by training a classifier on the latent space of DGMs on the graph data~\citep{jin2020hierarchical} or by learning to retrieve exemplar samples from a retrieval database for guidance~\citep{wang2022retrieval}. Other works fine-tune a pre-trained graph DGM using the supervised property annotations as rewards, resulting in a controllable DGM specifically for the considered task~\citep{olivecrona2017molecular, you2018graph}.
To the best of our knowledge, our work is the first to explore unsupervised graph editing in an unsupervised manner. Besides, different from many previous approaches that may only work for molecule graphs or point cloud graphs, our method is generic and thus can be applied to various graph modalities.

\section{Analysis Experiments on Disentanglement} \label{app:sec:toy_example}
In~\Cref{sec:entangled_representation}, we conduct three analysis experiments to conclude that the representation space is not perfectly disentangled in such setting. In this section, we provide more details and complementary information of these experiments.

\begin{table}[h]
\centering
\small
\caption{
The six mainstream disentanglement metrics on five DGMs and three data types. All measures range from 0 to 1, and higher scores mean more disentangled representation. MoFlow and HierVAE are for molecular graphs, PointFlow is for point clouds.
}
\label{app:table:disentanglement_measures}
\begin{adjustbox}{max width=\textwidth}
\begin{tabular}{l l l c c c c c c}
\toprule
Data Type & Model & Dataset & BetaVAE $\uparrow$ & FactorVAE $\uparrow$ & MIG $\uparrow$ & DCI $\uparrow$ & Modularity $\uparrow$ & SAP $\uparrow$ \\
\midrule
\multirow{4}{*}{Molecular Graphs}
& \multirow{2}{*}{MolFlow} & QM9 & 0.251 & 0.165 & 0.038 & 0.727 & 0.599 & 0.017\\
& & ZINC250k & 0.264 & 0.175 & 0.030 & 0.958 & 0.620 & 0.009 \\
\cmidrule(lr){2-9}
& \multirow{2}{*}{HierVAE} & QM9 & 0.165 & 0.135 & 0.044 & 0.241 & 0.626 & 0.076 \\
& & ChEMBL & 0.159 & 0.130 & 0.023 & 0.113 & 0.604 & 0.026 \\
\midrule
\multirow{3}{*}{Point Clouds} & \multirow{3}{*}{PointFlow} 
& Airplane & 0.022 & 0.025 & 0.029 & 0.160 & 0.745 & 0.022\\
& & Chair & 0.019 & 0.014 & 0.032 & 0.149 & 0.721 & 0.019\\
& & Car & 0.019 & 0.023 & 0.021 & 0.120 & 0.713 & 0.021\\
\bottomrule
\end{tabular}
\end{adjustbox}
\end{table}


\subsection{Steerable Factors}
As mentioned in~\Cref{sec:entangled_representation}, we consider measuring the disentanglement with respect to three types of structured data: molecular graphs and point clouds. Recall that we need to define steerable factors first, so as to measure the disentanglement.

\paragraph{Molecular Graph}
For molecular graphs, we treat the important substructures (a.k.a., motifs or fragments) as factors, and they are extracted using RDKit. To calculate the disentanglement for molecules, we randomly sample 10k molecules on QM9-MolFlow, ZINC250K-MolFlow, QM9-HierVAE, and ChEMBL-HierVAE, respectively. Most of the fragments do not show up or with very few times (less than 1\% occurrence frequency). Removing these fragments leads to the following 11 motifs, and we consider the existence of them as binary labels:
\begin{itemize}[noitemsep,topsep=0pt]
    \item aliphatic hydroxyl groups.
    \item aliphatic hydroxyl groups excluding tert-OH.
    \item aromatic nitrogens.
    \item aromatic amines.
    \item carbonyl O.
    \item carbonyl O, excluding COOH.
    \item Tertiary amines.
    \item Secondary amines.
    \item amides.
    \item ether oxygens (including phenoxy).
    \item nitriles.
\end{itemize}

\paragraph{Point Clouds} For point clouds, we adopt the viewpoint feature histogram (VFH)~\citep{rusu2010fast} implemented in PCL~\citep{Rusu_ICRA2011_PCL}. There are 308 bins in total, where each bin is a histogram of the angles that viewpoint direction makes with each normal. VFH has been widely used as point cloud descriptors, and here we binarize it into factors with:
\begin{itemize}[noitemsep,topsep=0pt]
    \item We collect the shared non-zero bins from all three datasets (Airplane, Car, and Chair), and ignore the bins where the values distribution are highly skewed. This can give us 75 bins.
    \item Then for each of these selected bins, we use the median value as the threshold to generate the binary factor labels.
\end{itemize}

\subsection{Disentanglement Measures}
We follow the~\citep{locatello2019challenging} on considering the following six disentanglement measures:
\begin{itemize}[noitemsep,topsep=0pt]
    \item $\beta$-VAE~\citep{higgins2016beta} evaluates the prediction accuracy of a linear classifier for the index of a fixed factor of variation.
    \item FactorVAE~\citep{kim2018disentangling} addresses the limitations (i.e. corner case) of $\beta$-VAE via introducing a majority voting classifier on a different feature vector. 
    \item MIG~\citep{chen2018isolating} measures the normalized difference between the highest and second highest mutual information between latent dimensions and each steerable factor.
    \item DCI~\citep{eastwood2018framework} disentanglement score measures the average difference from one of the entropy of probability that a latent dimension is useful for predicting a steerable factor (computed by the relative importance score).
    \item Modularity~\citep{ridgeway2018learning} measures whether each latent dimension is dependent on at most one single steerable factor. It computes the average normalized squared difference over the highest and second-highest mutual information between each steerable factor and each latent dimension.
    \item SAP~\citep{kumar2018variational} calculates the $R^2$ score of the linear models trained to predict each steerable factor from each latent dimension. 
\end{itemize}

Recall that all six disentanglement measures range from 0 to 1, and a higher value means the corresponding space is more disentangled. The results can be found in~\Cref{app:table:disentanglement_measures}. We can conclude that all the values are indeed low on all datasets and models, revealing that DGMs are entangled in general.

\section{Mutual Information} \label{app:sec:lower_bound_MI}
In this section, we will briefly introduce mutual information (MI), and also a lower bound for maximizing MI. This has been previously proposed in~\citep{liu2021pre} for self-supervised learning, and the comprehensive derivations are as follows. First for notation, without loss of generality, we use $X$ and $Y$ as the two views.

\subsection{Formulation}
The standard formulation for mutual information (MI) is
\begin{equation} \label{eq:app:MI}
\small{
\begin{aligned}
I(X;Y)
& = \mathbb{E}_{p(\vx,\vy)} \big[ \log \frac{p(\vx,\vy)}{p(\vx) p(\vy)} \big].
\end{aligned}
}
\end{equation}

Mutual information (MI) between random variables measures the corresponding non-linear dependence. As can be seen in the first equation in~\Cref{eq:app:MI}, the larger the divergence between the joint ($p(\vx, \vy$) and the product of the marginals $p(\vx) p(\vy)$, the stronger the dependence between $X$ and $Y$.

\subsection{A Lower Bound to MI}
First we can get a lower bound of MI. Assuming that there exist (possibly negative) constants $a$ and $b$ such that $a \le H(X)$ and $b \le H(Y)$, {\ie}, the lower bounds to the (differential) entropies, then we have:
\begin{equation}
\small
\begin{aligned}
I(X;Y)
& = \frac{1}{2} \big( H(X) + H(Y) - H(Y|X) - H(X|Y) \big)\\
& \ge \frac{1}{2} \big( a + b - H(Y|X) - H(X|Y) \big)\\
& \ge \frac{1}{2} \big( a + b \big) + \mathcal{L}_{\text{MI}}.\\
\end{aligned}
\end{equation}
Thus, we transform the MI maximization problem into maximizing the following objective:
\begin{equation} \label{eq:app:MI_objective}
\small{
\begin{aligned}
\mathcal{L}_{\text{MI}} 
& = \frac{1}{2} \mathbb{E}_{p(\vx,\vy)} \Big[ \log p(\vx|\vy) \Big] + \frac{1}{2} \mathbb{E}_{p(\vx,\vy)} \Big[ \log p(\vy|\vx) \Big].
\end{aligned}
}
\end{equation}
Empirically, we use energy-based models to model the distributions. The existence of $a$ and $b$ can be understood as the requirements that the two distributions ($p_\vx, p_\vy$) are not collapsed. For the next, we will try to model the two conditional data distributions $p$ with model distributions $p_\theta$.

\subsection{Derivation of conditional EBM with NCE}
WLOG, let's consider modeling the $p_\theta(\vx|\vy)$ first, and by EBM it is as follows:
\begin{equation} \label{eq:app:EBM_SSL}
\small
p_\theta(\vx|\vy) = \frac{\exp(-E(\vx|\vy))}{ \int \exp(-E({\tilde \vx}|\vy)) d{\tilde \vx}} = \frac{\exp(f_\vx(\vx, \vy))}{\int \exp(f_\vx({\tilde \vx}|\vy)) d{\tilde \vx}} = \frac{\exp(f_\vx(\vx, \vy))}{A_{\vx|\vy}}.
\end{equation}

Then we solve this using NCE. NCE handles the intractability issue by transforming it as a binary classification task. We take the partition function $A_{\vx|\vy}$ as a parameter, and introduce a noise distribution $p_n$. Based on this, we introduce a mixture model: $\vz=0$ if the conditional $\vx|\vy$ is from $p_n(\vx|\vy)$, and $\vz=1$ if $\vx|\vy$ is from $p_{\text{data}}(\vx|\vy)$. So the joint distribution is:
\begin{equation*}
\small
    p_{n,\text{\text{data}}}(\vx|\vy) = p(\vz=1) p_{\text{data}}(\vx|\vy) + p(\vz=0) p_n(\vx|\vy)
\end{equation*}

The posterior of $p(\vz=0|\vx,\vy)$ is
\begin{equation*}
\small
p_{n,\text{\text{data}}}(\vz=0|\vx,\vy) = \frac{p(\vz=0) p_n(\vx|\vy)}{p(z=0) p_n(\vx|\vy) + p(\vz=1) p_{\text{data}}(\vx|\vy)} = \frac{\nu \cdot p_n(\vx|\vy)}{\nu \cdot p_n(\vx|\vy) + p_{\text{data}}(\vx|\vy)},
\end{equation*}
where $\nu = \frac{p(\vz=0)}{p(\vz=1)}$.

Similarly, we can have the joint distribution under EBM framework as:
\begin{equation*}
\small
p_{n, \theta}(\vx) = p(z=0) p_n(\vx|\vy) + p(z=1) p_{\theta}(\vx|\vy)
\end{equation*}
And the corresponding posterior is:
\begin{equation*}
p_{n,\theta}(\vz=0|\vx,\vy) = \frac{p(\vz=0) p_n(\vx|\vy)}{p(\vz=0) p_n(\vx|\vy) + p(\vz=1) p_{\theta}(\vx|\vy)} = \frac{\nu \cdot p_n(\vx|\vy)}{\nu \cdot p_n(\vx|\vy) + p_{\theta}(\vx|\vy)}
\end{equation*}

We indirectly match $p_\theta(\vx|\vy)$ to $p_{\text{data}}(\vx|\vy)$ by fitting $p_{n,\theta}(\vz|\vx,\vy)$ to $p_{n,\text{\text{data}}}(\vz|\vx,\vy)$ by minimizing their KL-divergence:
\begin{equation} \label{eq:app:EBM_01}
\small
\begin{aligned}
& \min_\theta  D_{\text{KL}}(p_{n,\text{\text{data}}}(\vz|\vx,\vy) || p_{n,\theta}(\vz|\vx,\vy)) \\
& =  \mathbb{E}_{p_{n,\text{\text{data}}}(\vx,\vz|\vy)} [\log p_{n,\theta}(\vz|\vx,\vy)] \\
& =   \int \sum_\vz p_{n,\text{\text{data}}}(\vx,\vz|\vy) \cdot \log p_{n,\theta}(\vz|\vx,\vy) d \vx\\
& =  \int \Big\{ p(\vz=0) p_{n,\text{\text{data}}}(\vx|\vy,\vz=0) \log p_{n,\theta}(\vz=0|\vx,\vy) \\
 & \quad\quad\quad\quad + p(\vz=1) p_{n,\text{\text{data}}}(\vx|\vz=1,\vy) \log p_{n,\theta}(\vz=1|\vx,\vy) \Big\} d\vx \\
& =  \nu \cdot \mathbb{E}_{p_{n}(\vx|\vy)} \Big[\log p_{n,\theta}(\vz=0|\vx,\vy) \Big] + \mathbb{E}_{p_{\text{data}}(\vx|\vy )} \Big[\log p_{n,\theta}(\vz=1|\vx,\vy) \Big] \\
& =  \nu \cdot \mathbb{E}_{p_{n}(\vx|\vy)} \Big[ \log \frac{\nu \cdot p_n(\vx|\vy)}{\nu \cdot p_n(\vx|\vy) + p_{\theta}(\vx|\vy)} \Big] + \mathbb{E}_{p_{\text{data}}(\vx|\vy )} \Big[ \log \frac{p_\theta(\vx|\vy)}{\nu \cdot p_n(\vx|\vy) + p_{\theta}(\vx|\vy)} \Big].\\
\end{aligned}
\end{equation}
This optimal distribution is an estimation to the actual distribution (or data distribution), {\ie}, $p_\theta(\vx|\vy) \approx p_{\text{data}}(\vx|\vy)$. We can follow the similar steps for $p_\theta(\vy|\vx) \approx p_{\text{data}}(\vy|\vx)$. Thus following~\Cref{eq:app:EBM_01}, the objective function is to maximize
\begin{equation} \label{eq:app:EBM_02}
\small
\begin{aligned}
\nu \cdot \mathbb{E}_{p_{\text{data}}(\vy)} \mathbb{E}_{p_{n}(\vx|\vy)} \Big[ \log \frac{\nu \cdot p_n(\vx|\vy)}{\nu \cdot p_n(\vx|\vy) + p_{\theta}(\vx|\vy)} \Big] + \mathbb{E}_{p_{\text{data}}(\vy)} \mathbb{E}_{p_{\text{data}}(\vx|\vy )} \Big[ \log \frac{p_\theta(\vx|\vy)}{\nu \cdot p_n(\vx|\vy) + p_{\theta}(\vx|\vy)} \Big].
\end{aligned}
\end{equation}

The we will adopt three strategies to approximate~\Cref{eq:app:EBM_02}:
\begin{enumerate}
    \item \textbf{Self-normalization.} When the EBM is very expressive, {\ie}, using deep neural network for modeling, we can assume it is able to approximate the normalized density directly~\citep{mnih2012fast,song2021train}. In other words, we can set the partition function $A=1$. This is a self-normalized EBM-NCE, with normalizing constant close to 1, {\ie}, $p(\vx) = \exp(-E(\vx)) = \exp(f(\vx))$.
    \item \textbf{Exponential tilting term.} Exponential tilting term~\citep{arbel2020generalized} is another useful trick. It models the distribution as $\tilde p_\theta(\vx) = q(\vx) \exp(-E_\theta(\vx)) $, where $q(\vx)$ is the reference distribution. If we use the same reference distribution as the noise distribution, the tilted probability is $\tilde p_\theta(\vx) = p_n(\vx) \exp(-E_\theta(\vx))$.
    \item \textbf{Sampling.} For many cases~\citep{liu2021pre}, we only need to sample 1 negative points for each data, {\ie}, $\nu=1$.
\end{enumerate}

Following these three disciplines, the objective function to optimize $p_{\theta}(\vx|\vy)$ becomes
\begin{equation}
\small
\begin{aligned}
& \mathbb{E}_{p_{n}(\vx|\vy)} \Big[ \log \frac{ p_n(\vx|\vy)}{ p_n(\vx|\vy) + \tilde p_{\theta}(\vx|\vy)} \Big] + \mathbb{E}_{p_{\text{data}}(\vx|\vy )} \Big[ \log \frac{\tilde p_\theta(\vx|\vy)}{ p_n(\vx|\vy) + \tilde p_{\theta}(\vx|\vy)} \Big]\\
= &   \mathbb{E}_{p_{n}(\vx|\vy)} \Big[ \log \frac{1}{1 + p_{\theta}(\vx|\vy)} \Big] + \mathbb{E}_{p_{\text{data}}(\vx|\vy )} \Big[ \log \frac{p_\theta(\vx|\vy)}{1 + p_{\theta}(\vx|\vy)} \Big]\\
= &   \mathbb{E}_{p_{n}(\vx|\vy)} \Big[ \log \frac{\exp (-  f_\vx(\vx, \vy))}{\exp (-  f_\vx(\vx, \vy)) + 1} \Big] + \mathbb{E}_{p_{\text{data}}(\vx|\vy )} \Big[ \log \frac{1}{\exp (-  f_\vx(\vx, \vy)) + 1} \Big]\\
= &   \mathbb{E}_{p_{n}(\vx|\vy)} \Big[ \log \big(1-\sigma(  f_\vx(\vx, \vy))\big) \Big] + \mathbb{E}_{p_{\text{data}}(\vx|\vy )} \Big[ \log \sigma(  f_\vx(\vx, \vy)) \Big].
\end{aligned}
\end{equation}

Thus, the final EBM-NCE contrastive SSL objective is
\begin{equation}
\begin{aligned}
\mathcal{L}_{\text{EBM-NCE}}
& = -\frac{1}{2} \mathbb{E}_{p_{\text{data}}(\vy)} \Big[\mathbb{E}_{p_{n}(\vx|\vy)} \log \big(1-\sigma(  f_\vx(\vx, \vy))\big) + \mathbb{E}_{p_{\text{data}}(\vx|\vy )} \log \sigma(  f_\vx(\vx, \vy)) \Big]\\
& ~~~~ - \frac{1}{2} \mathbb{E}_{p_{\text{data}}(\vx)} \Big[\mathbb{E}_{p_{n}(\vy|\vx)} \log \big(1-\sigma(  f_\vy(\vy,\vx))\big) + \mathbb{E}_{p_{\text{data}}(\vy,\vx)} \log \sigma(  f_\vy(\vy,\vx)) \Big].
\end{aligned}
\end{equation}

\clearpage
\newpage
\section{Implementation and Experiment Details} \label{app:sec:experimental_details}

\textbf{Editing function.}
For the editing function, we consider both linear (\Cref{app:eq:linear_01,app:eq:linear_02}) and non-linear (\Cref{app:eq:non_linear}) cases as below, {\ie}, for $\mz_{i,\alpha} = h(\vz, \vd_i, \alpha)$:
{
{
\fontsize{7.8}{1}\selectfont
\begin{align}
& \mz_{i,\alpha} = \vz + \alpha \cdot \vd_i, ~~~ \vd_i = \text{norm} \circ \text{Linear}(\ve_i), \label{app:eq:linear_01}
\\
& \mz_{i,\alpha} = \vz + \alpha \cdot \vd_i, ~~~ \vd_i = \text{sqrt} \circ \text{norm} \circ \text{ReLU} \circ \text{Linear}(\ve_i), \label{app:eq:linear_02}
\\
& \mz_{i,\alpha} = \vz + \alpha \cdot \vd_i + \text{norm} \circ \text{Linear} \circ \text{ReLU} \circ \text{Linear} (\vz \oplus \vd_i \oplus [\alpha]), ~~~ \vd_i = \text{norm} \circ \text{Linear} \circ \text{ReLU} \circ \text{Linear}(\ve_i), \label{app:eq:non_linear}
\end{align}
}
}%
where $\circ$ means the composition of two functions.

\textbf{Objective function.}
The objective function is given as:
\begin{equation}
\small{
\mathcal{L} = c_1 \cdot \mathbb{E}_{u,v} [\mathcal{L}_{\text{\model-NCE}}] + c_2 \cdot \mathcal{L}_{\text{sim}} + c_3 \cdot \mathcal{L}_{\text{sparsity}},
}
\end{equation}
where $\mathcal{L}_{\text{\model-NCE}}$ is the MI estimation defined in~\Cref{eq:objective_ebm_nce}, $\mathcal{L}_{\text{sim}}$ is the direction similarity defined in~\Cref{eq:similarity_direction}, and $\mathcal{L}_{\text{sparsity}}$.  $c_1$, $c_2$ and $c_3$ are three coefficients accordingly.

\textbf{Hyperparameters.}
We list the key hyperparameters in~\Cref{tab:app:hyperparameter_specifications}, and all the results are evaluated on 100 sampled sequences. We also want to highlight that DisCo is an unstable baseline, in the sense that once we add more training data ({\eg}, from 100 to 500) or more training epochs ({\eg}, from 1 epoch to 5 epochs), the model will collapse, with the nan loss. Thus, here we are reporting the most reasonable results for DisCo, {\ie}, 100 training data with 1 epoch.

\begin{table}[htb]
\centering
\setlength{\tabcolsep}{5pt}
\fontsize{9}{9}\selectfont
\caption{
\small{
Hyperparameter specifications.
}
}
\label{tab:app:hyperparameter_specifications}
\begin{adjustbox}{max width=\textwidth}
\begin{tabular}{l l l}
\toprule
 & Hyperparameter & Value\\
\midrule

\multirow{2}{*}{Random}
& D & \{10\} \\
& $\alpha$ & \{-3, -2.7, -2.4, ..., 2.7, 3\}\\
\midrule

\multirow{3}{*}{Variance}
& D & \{10\} \\
& $\alpha$ & \{-3, -2.7, -2.4, ..., 2.7, 3\}\\
& \# training data & \{100, 500\}\\
\midrule

\multirow{3}{*}{SeFa}
& D & \{10\} \\
& $\alpha$ & \{-3, -2.7, -2.4, ..., 2.7, 3\}\\
& \# training data & \{100, 500\}\\
\midrule

\multirow{3}{*}{DisCo}
& D & \{10\} \\
& $\alpha$ & \{-3, -2.7, -2.4, ..., 2.7, 3\}\\
& \# training data & \{100\}\\
& epochs & \{1\}\\
\midrule

\multirow{6}{*}{\model}
& D & \{10\} \\
& $\alpha$ & \{-3, -2.7, -2.4, ..., 2.7, 3\}\\
& \# training data & \{100, 500\}\\
& epochs & \{20, 100\}\\
& coefficient $c_1$ & \{1, 2\} \\
& coefficient $c_2$ & \{0, 1\} \\
& coefficient $c_3$ & \{0, 1\} \\
\bottomrule
\end{tabular}
\end{adjustbox}
\end{table}

\textbf{Hardware.}
We use V100 GPU cards, and each job (w.r.t. different hyperparameters) for \model\, can be finished within 3 hours on a single GPU card.

\textbf{Time complexity.}
The time complexity of GraphCG is $O(B \times D^2)$ for GraphCG-P and $O(B^2 \times D^2)$ for GraphCG-R, where $B$ is the number of data points for each batch. Here we omit the constants for step-sizes.

\clearpage
\newpage
\section{Results: Molecular Graph} \label{app:sec:molecular_graph_results}

\subsection{Evaluation Metrics}

\begin{figure}[h]
\centering
\begin{subfigure}[Tanimoto Similarity Sequence]{
    \includegraphics[width=0.45\linewidth]{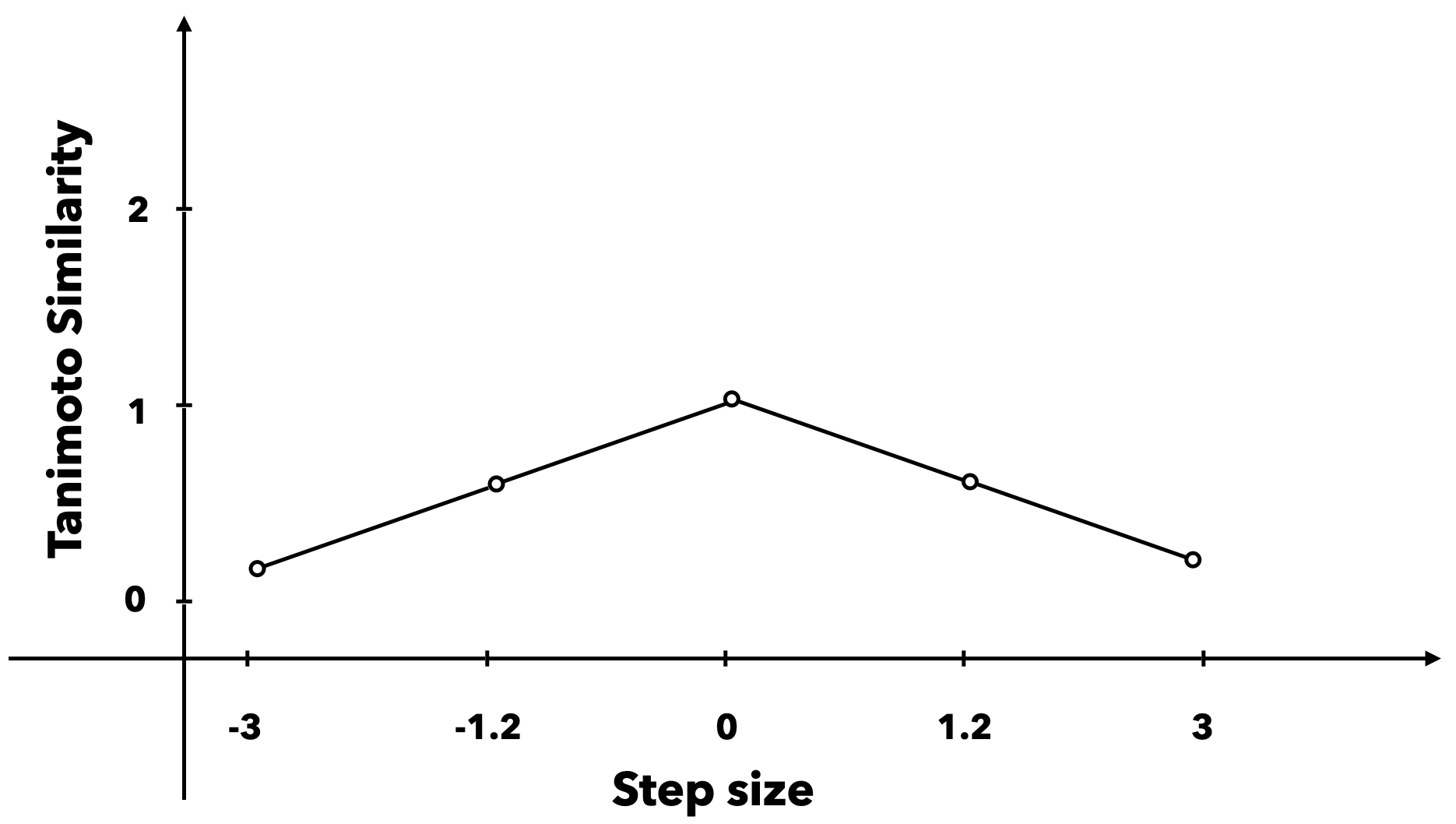}
    \label{fig:original_tanimoto_similarity_sequence}
}
\end{subfigure}
\hfill
\begin{subfigure}[Calibrated Tanimoto Similarity Sequence]{
    \includegraphics[width=0.45\linewidth]{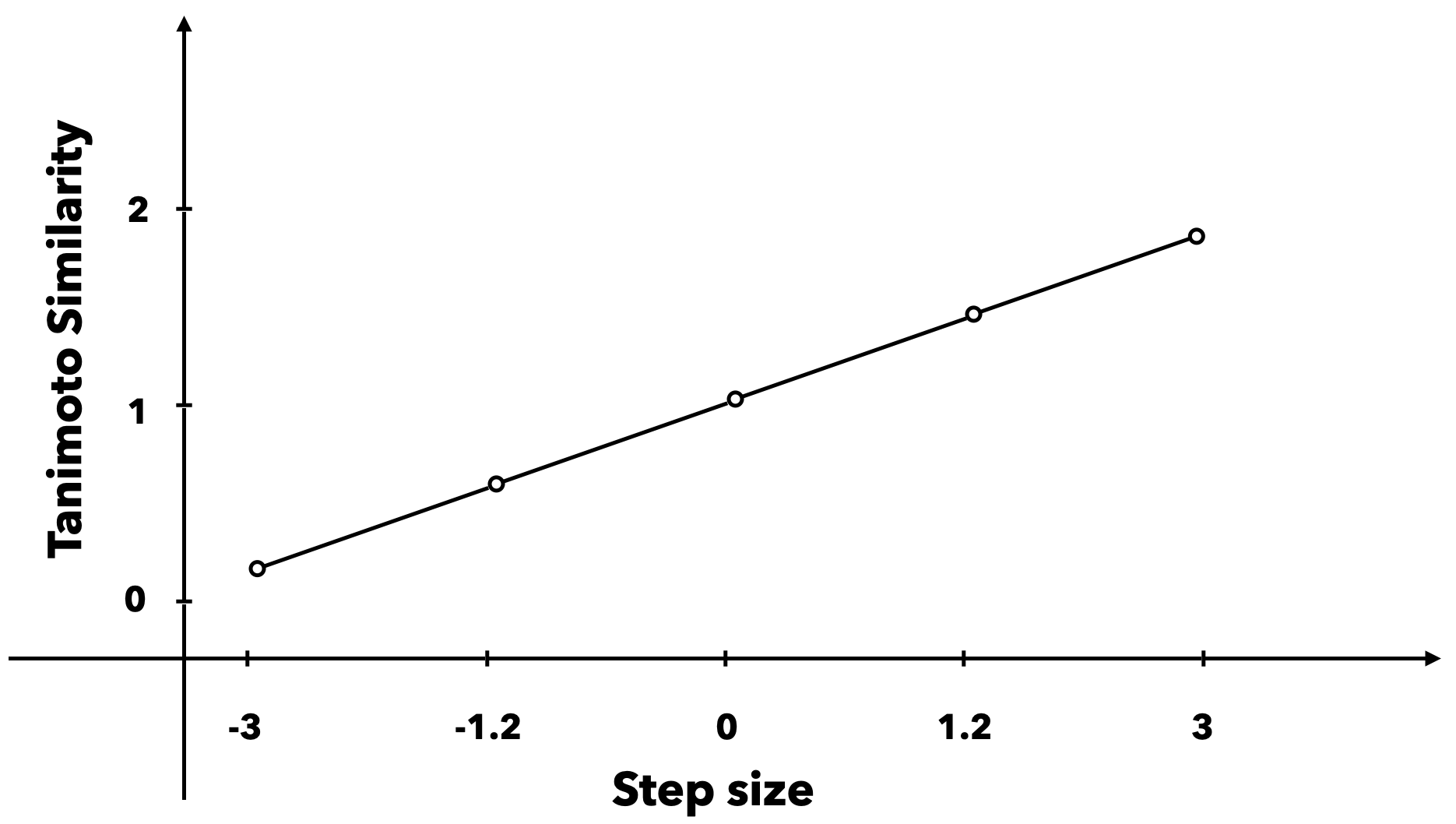}
    \label{fig:calibrated_tanimoto_similarity_sequence}
}
\end{subfigure}
\caption{
\small
\Cref{fig:original_tanimoto_similarity_sequence} is the original Tanimoto similarity sequence w.r.t. the anchor molecule, {\ie}, step size with 0 in the figure. Yet, this is not easy to compute the monotonicity. We thus propose the calibrated Tanimoto similarity sequence, by taking the deduction from 2 for output molecules with positive step size, as shown in~\Cref{fig:calibrated_tanimoto_similarity_sequence}.
}
\label{fig:tanimoto_similarity_sequence}
\end{figure}

\textbf{Change of Structure Factors and Calibrated Tanimoto Similarity (CTS).}
We are interested in the change of the graph structure (the steerable factors) along the output sequence edited with the $i$-th direction.
To evaluate the structure change, we apply the Tanimoto similarity between each output molecule and the anchor molecule, as shown in~\Cref{fig:original_tanimoto_similarity_sequence}.
Besides, for the ease of evaluating the monotonicity, we utilize a transformation (on the Tanimoto similarity) of output molecules with positive step size by taking the deduction from 2. We call this calibrated Tanimoto similarity sequence (CTS), {\ie}, $\{ s(\bar \vx') \}_i$, as shown in~\Cref{fig:calibrated_tanimoto_similarity_sequence}.

\textbf{Sequence Monotonic Ratio (SMR)}.
For evaluation, we propose a metric called Sequence Monotonic Ratio (SMR), $\phi_{\text{SMR}}(\gamma, \tau)_i$. It measures the monotonic ratio of $M$ generated sequences edited with the $i$-th direction. It has two arguments: the diversity threshold $\gamma$ constrains the minimum number of distinct molecules, and the tolerance threshold $\tau$ controls the non-monotonic tolerance ratio along each sequence.

In specific, for each learned semantic direction $i$, we will generate $M$ sequences of edited molecules, and the calibrated Tanimoto similarity for each sequence is marked as $\{ s(\bar \vx') \}_i^m$. Then we can define the SMR on each direction as: \looseness=-1
\begin{equation} \label{app:eq:SMR_each_direction}
\fontsize{7.8}{1}\selectfont
\begin{aligned}
& \phi_{\text{SMR}}(\gamma, \tau )_i = \frac{1}{M} \sum_{m=1}^{M} \phi_{\text{SMR}}\big( \{ s(\bar \vx') \}_i^m, \gamma, \tau \big),
\\
& \phi_{\text{SMR}}\big( \{ s(\bar \vx') \}_i^m, \gamma, \tau \big) =
\begin{cases}
1, & \text{len}\big(\text{set}\big\{ s(\bar \vx') \}_i^m\big)\big) \ge \gamma \wedge \text{ monotonic}_\tau \big(\{ s(\bar \vx') \}_i^m\big)\\
0, & \text{otherwise}
\end{cases}.
\end{aligned}
\end{equation}

\textbf{Evaluating the Diversity of Semantic Directions.}
SMR can evaluate all the output sequences generated by one direction. To better illustrate that \model\, is able to learn multiple directions with various semantic information, we also consider taking the average of top-$K$ SMR w.r.t. directions to reveal that all the best $K$ directions are semantically meaningful, as in~\Cref{app:eq:top_K_SMR}:
\begin{equation} \label{app:eq:top_K_SMR}
\fontsize{7.8}{1}\selectfont
\begin{aligned}
& \text{top-K}(\gamma, \tau) = \frac{1}{K} \sum_{i \in \text{top-K directions}} \Big( \phi_{\text{SMR}}(\gamma, \tau)_i \Big).
\end{aligned}
\end{equation}

\newpage
\subsection{Results on Molecular Structures}
Next we would like to show the comprehensive SMR on the CTS results with respect to different backbone models, as in~\Cref{app:tab:SMR_CTS_MoFlow,app:tab:SMR_CTS_HierVAE}.

\begin{table}[ht]
\begin{center}
\vspace{-3ex}
\caption{
\small
This table lists the sequence monotonic ratio (SMR, \%) on calibrated Tanimoto similarity (CST) w.r.t. the top-1, top-2, and top-3 directions. The backbone model is the pretrained MoFlow on ZINC250k.
}
\label{app:tab:SMR_CTS_MoFlow}
\small
\setlength{\tabcolsep}{10pt}
\begin{adjustbox}{max width=\textwidth}
\begin{tabular}{ll cccccc cccccc cccccc}
\toprule
\multirow{3}{*}{Edit Method} & top-K
 & \multicolumn{6}{c}{\textbf{Tanimoto top-1}}
 & \multicolumn{6}{c}{\textbf{Tanimoto top-2}}
 & \multicolumn{6}{c}{\textbf{Tanimoto top-3}}
\\
\cmidrule(lr){3-8} \cmidrule(lr){9-14} \cmidrule(lr){15-20}
 & $\gamma$ 
 & \multicolumn{2}{c}{2} & \multicolumn{2}{c}{3} & \multicolumn{2}{c}{4}
 & \multicolumn{2}{c}{2} & \multicolumn{2}{c}{3} & \multicolumn{2}{c}{4}
 & \multicolumn{2}{c}{2} & \multicolumn{2}{c}{3} & \multicolumn{2}{c}{4}
\\
\cmidrule(lr){3-4} \cmidrule(lr){5-6} \cmidrule(lr){7-8} \cmidrule(lr){9-10} \cmidrule(lr){11-12} \cmidrule(lr){13-14} \cmidrule(lr){15-16} \cmidrule(lr){17-18} \cmidrule(lr){19-20}
 & $\tau$
 & 0 & 0.2 & 0 & 0.2 & 0 & 0.2
 & 0 & 0.2 & 0 & 0.2 & 0 & 0.2
 & 0 & 0.2 & 0 & 0.2 & 0 & 0.2
\\
\midrule
Random & & 35.0 & 36.0 & 23.0 & 25.0 & 12.0 & 15.0 & 34.5 & 36.0 & 22.5 & 25.0 & 11.5 & 14.0 & 34.0 & 36.0 & 22.0 & 24.0 & 11.0 & 13.7\\
Variance & & 32.0 & 36.0 & 24.0 & 28.0 & 12.0 & 16.0 & 31.5 & 35.5 & 21.0 & 26.5 & 10.5 & 16.0 & 30.3 & 35.3 & 20.0 & 25.0 & 10.0 & 15.0\\
SeFa & & 23.0 & 23.0 & 4.0 & 4.0 & 0.0 & 0.0 & 19.0 & 19.0 & 4.0 & 4.0 & 0.0 & 0.0 & 17.3 & 17.3 & 3.3 & 3.3 & 0.0 & 0.0\\
DisCo & & 8.0 & 15.0 & 7.0 & 14.0 & 2.0 & 8.0 & 7.5 & 13.5 & 6.0 & 12.5 & 2.0 & 8.0 & 7.0 & 13.0 & 5.3 & 11.7 & 2.0 & 7.7\\
\midrule
\multirow{3}{*}{GraphCG-P}
 & \Cref{app:eq:linear_01} & 39.0 & 40.0 & 27.0 & 28.0 & 15.0 & 18.0 & 38.5 & 40.0 & 26.0 & 28.0 & 15.0 & 18.0 & 37.0 & 39.0 & 24.7 & 27.3 & 14.3 & 17.3\\
 & \Cref{app:eq:linear_02} & 35.0 & 37.0 & 19.0 & 22.0 & 8.0 & 11.0 & 33.5 & 36.5 & 18.5 & 21.5 & 7.0 & 10.5 & 31.7 & 34.7 & 17.7 & 20.7 & 6.3 & 9.3\\
 & \Cref{app:eq:non_linear} & 44.0 & 46.0 & 32.0 & 34.0 & 16.0 & 18.0 & 42.5 & 44.5 & 30.0 & 32.0 & 15.0 & 17.5 & 41.7 & 44.0 & 29.0 & 31.0 & 13.7 & 16.3\\
\midrule
\multirow{3}{*}{GraphCG-R}
 & \Cref{app:eq:linear_01} & 37.0 & 42.0 & 25.0 & 26.0 & 11.0 & 14.0 & 37.0 & 40.0 & 23.0 & 25.5 & 11.0 & 13.5 & 36.3 & 39.0 & 22.0 & 24.3 & 10.3 & 13.3\\
 & \Cref{app:eq:linear_02} & 35.0 & 37.0 & 19.0 & 22.0 & 8.0 & 11.0 & 33.5 & 36.5 & 18.5 & 21.5 & 7.0 & 10.5 & 31.7 & 34.7 & 17.7 & 20.7 & 6.3 & 9.3\\
 & \Cref{app:eq:non_linear} & 0.0 & 0.0 & 0.0 & 0.0 & 0.0 & 0.0 & 0.0 & 0.0 & 0.0 & 0.0 & 0.0 & 0.0 & 0.0 & 0.0 & 0.0 & 0.0 & 0.0 & 0.0\\
\bottomrule
\end{tabular}
\end{adjustbox}
\end{center}
\end{table}

\begin{table}[ht]
\begin{center}
\vspace{-3ex}
\caption{
\small 
This table lists the sequence monotonic ratio (SMR, \%) on calibrated Tanimoto similarity (CST) w.r.t. the top-1, top-2, and top-3 directions. The backbone model is the pretrained HierVAE on ChEMBL.
}
\label{app:tab:SMR_CTS_HierVAE}
\small
\setlength{\tabcolsep}{10pt}
\begin{adjustbox}{max width=\textwidth}
\begin{tabular}{ll cccccc cccccc cccccc}
\toprule
\multirow{3}{*}{Edit Method} & top-K
 & \multicolumn{6}{c}{\textbf{Tanimoto top-1}}
 & \multicolumn{6}{c}{\textbf{Tanimoto top-2}}
 & \multicolumn{6}{c}{\textbf{Tanimoto top-3}}
\\
\cmidrule(lr){3-8} \cmidrule(lr){9-14} \cmidrule(lr){15-20}
 & $\gamma$ 
 & \multicolumn{2}{c}{2} & \multicolumn{2}{c}{3} & \multicolumn{2}{c}{4}
 & \multicolumn{2}{c}{2} & \multicolumn{2}{c}{3} & \multicolumn{2}{c}{4}
 & \multicolumn{2}{c}{2} & \multicolumn{2}{c}{3} & \multicolumn{2}{c}{4}
\\
\cmidrule(lr){3-4} \cmidrule(lr){5-6} \cmidrule(lr){7-8} \cmidrule(lr){9-10} \cmidrule(lr){11-12} \cmidrule(lr){13-14} \cmidrule(lr){15-16} \cmidrule(lr){17-18} \cmidrule(lr){19-20}
 & $\tau$
 & 0 & 0.2 & 0 & 0.2 & 0 & 0.2
 & 0 & 0.2 & 0 & 0.2 & 0 & 0.2
 & 0 & 0.2 & 0 & 0.2 & 0 & 0.2
\\
\midrule
Random & & 14.0 & 45.0 & 14.0 & 45.0 & 14.0 & 43.0 & 11.0 & 43.5 & 11.0 & 43.5 & 11.0 & 42.5 & 10.0 & 42.3 & 10.0 & 42.3 & 9.3 & 41.7\\
Variance & & 28.0 & 64.0 & 23.0 & 59.0 & 19.0 & 55.0 & 22.5 & 59.5 & 19.5 & 57.0 & 17.5 & 55.0 & 20.3 & 54.3 & 18.3 & 52.7 & 15.7 & 50.3\\
SeFa & & 4.0 & 41.0 & 4.0 & 41.0 & 4.0 & 41.0 & 3.0 & 41.0 & 3.0 & 41.0 & 3.0 & 41.0 & 2.3 & 36.0 & 2.3 & 36.0 & 2.3 & 36.0\\
\midrule
\multirow{3}{*}{GraphCG-P}
 & \Cref{app:eq:linear_01} & 23.0 & 61.0 & 19.0 & 57.0 & 15.0 & 53.0 & 19.0 & 59.0 & 16.5 & 56.5 & 13.5 & 53.0 & 17.0 & 55.3 & 15.3 & 53.7 & 12.7 & 50.7\\
 & \Cref{app:eq:linear_02} & 62.0 & 77.0 & 40.0 & 73.0 & 32.0 & 65.0 & 60.5 & 74.0 & 38.5 & 67.5 & 29.0 & 61.0 & 59.3 & 70.3 & 36.0 & 64.3 & 26.3 & 57.7\\
 & \Cref{app:eq:non_linear} & 29.0 & 71.0 & 28.0 & 70.0 & 27.0 & 69.0 & 22.0 & 62.0 & 21.5 & 61.5 & 20.5 & 61.0 & 18.7 & 57.7 & 18.3 & 57.3 & 17.7 & 56.7\\
\midrule
\multirow{3}{*}{GraphCG-R}
 & \Cref{app:eq:linear_01} & 16.0 & 56.0 & 16.0 & 56.0 & 15.0 & 55.0 & 13.5 & 47.0 & 13.5 & 47.0 & 12.0 & 47.0 & 11.7 & 44.7 & 11.7 & 44.7 & 10.7 & 43.3\\
 & \Cref{app:eq:linear_02} & 61.0 & 74.0 & 42.0 & 67.0 & 30.0 & 55.0 & 59.0 & 69.5 & 40.0 & 64.0 & 29.5 & 54.5 & 57.7 & 67.7 & 38.0 & 62.3 & 28.7 & 53.7\\
 & \Cref{app:eq:non_linear} & 25.0 & 57.0 & 24.0 & 57.0 & 21.0 & 55.0 & 20.0 & 55.0 & 19.5 & 54.5 & 17.0 & 52.0 & 17.3 & 52.7 & 17.0 & 52.3 & 15.0 & 50.0\\
\bottomrule
\end{tabular}
\end{adjustbox}
\end{center}
\end{table}


\clearpage

\subsection{Results on Molecular Properties}

By far, we have been mainly focusing on the structure change of the output sequence of molecules. Yet, some works~\citep{doi:10.1021/ed064p575} also proved that certain key components of the molecules can be closely related to the molecular properties. We summarize the properties into roughly three categories, and the SMR on 8 properties with different $\gamma$, $\tau$ are listed below.
\begin{enumerate}[noitemsep,topsep=0pt]
    \item Physical properties, including water-octanol partition coefficient (LogP)~\citep{comer2001lipophilicity}, topological polar surface area (tPSA)~\citep{ertl2000fast}), and molecular weight (MW).
    \item Drug-related molecular properties, including drug-likeness (QED)~\citep{bickerton2012quantifying} and synthetic accessibility (SA)~\citep{ertl2009estimation}.
    \item Biological properties, including three binding affinity tasks (DRD2, JNK3, GSK3$\beta$). All the oracle labels are provided from the previous works~\citep{olivecrona2017molecular,jin2020multi,sun2017excape}.
\end{enumerate}

\begin{table}[htb!]
\begin{center}
\vspace{-3ex}
\caption{
\small
This table lists the sequence monotonic ratio (SMR, \%) on 8 molecule properties with $\gamma = 2, \tau=0$.
}
\vspace{-2ex}
\setlength{\tabcolsep}{10pt}
\begin{adjustbox}{max width=\textwidth}
\begin{tabular}{l l c c c c c c c c c c c}
\toprule
\multirow{2}{*}{\makecell{Model \&\\Dataset}} & \multirow{2}{*}{Edit Method} & \multirow{2}{*}{$h$} & \multicolumn{3}{c}{Physical} & \multicolumn{2}{c}{Drug-related} & \multicolumn{3}{c}{Bioactivity}\\
\cmidrule(lr){4-6} \cmidrule(lr){7-8} \cmidrule(lr){9-11}
 & & & LogP $\uparrow$ & tPSA $\uparrow$ & MW $\uparrow$ & QED $\uparrow$ & SA $\uparrow$ & DRD2 $\uparrow$ & JNK3 $\uparrow$ & GSK3$\beta$ $\uparrow$\\
\midrule

\multirow{10}{*}{\makecell{MoFlow\\ZINC250k}}
& Random & -- & 22.0 & 28.0 & 23.0 & 24.0 & 24.0 & 21.0 & 27.0 & 24.0\\
& Variance & -- & 24.0 & 24.0 & 18.0 & 21.0 & 19.0 & 20.0 & 26.0 & 22.0\\
& SeFa & -- & 20.0 & 20.0 & 23.0 & 21.0 & 22.0 & 22.0 & 21.0 & 22.0\\
& DisCo & -- & 5.0 & 5.0 & 5.0 & 4.0 & 4.0 & 5.0 & 6.0 & 5.0\\
\cmidrule(lr){2-11}
& \multirow{3}{*}{GraphCG-P}
 & \Cref{app:eq:linear_01} & 26.0 & 30.0 & 24.0 & 25.0 & 22.0 & 23.0 & 27.0 & 26.0\\
&  & \Cref{app:eq:linear_02} & 25.0 & 29.0 & 25.0 & 24.0 & 27.0 & 24.0 & 25.0 & 27.0\\
&  & \Cref{app:eq:non_linear} & 27.0 & 39.0 & 28.0 & 27.0 & 28.0 & 26.0 & 39.0 & 32.0\\
\cmidrule(lr){2-11}
& \multirow{3}{*}{GraphCG-R}
 & \Cref{app:eq:linear_01} & 27.0 & 27.0 & 26.0 & 26.0 & 25.0 & 23.0 & 33.0 & 28.0\\
&  & \Cref{app:eq:linear_02} & 25.0 & 29.0 & 25.0 & 24.0 & 27.0 & 24.0 & 25.0 & 27.0\\
&  & \Cref{app:eq:non_linear} & 0.0 & 0.0 & 0.0 & 0.0 & 0.0 & 0.0 & 0.0 & 0.0\\
\midrule
\multirow{9}{*}{\makecell{HierVAE\\ChEMBL}}
& Random & -- & 3.0 & 5.0 & 4.0 & 2.0 & 2.0 & 1.0 & 11.0 & 8.0\\
& Variance & -- & 9.0 & 9.0 & 8.0 & 11.0 & 9.0 & 7.0 & 25.0 & 22.0\\
& SeFa & -- & 1.0 & 0.0 & 2.0 & 0.0 & 0.0 & 0.0 & 1.0 & 2.0\\
\cmidrule(lr){2-11}
& \multirow{3}{*}{GraphCG-P}
 & \Cref{app:eq:linear_01} & 7.0 & 12.0 & 9.0 & 6.0 & 5.0 & 7.0 & 21.0 & 15.0\\
&  & \Cref{app:eq:linear_02} & 52.0 & 56.0 & 50.0 & 50.0 & 54.0 & 52.0 & 55.0 & 53.0\\
&  & \Cref{app:eq:non_linear} & 7.0 & 12.0 & 3.0 & 3.0 & 4.0 & 2.0 & 30.0 & 20.0\\
\cmidrule(lr){2-11}
& \multirow{3}{*}{GraphCG-R}
 & \Cref{app:eq:linear_01} & 2.0 & 6.0 & 3.0 & 2.0 & 2.0 & 2.0 & 11.0 & 9.0\\
&  & \Cref{app:eq:linear_02} & 52.0 & 54.0 & 49.0 & 49.0 & 52.0 & 51.0 & 54.0 & 53.0\\
&  & \Cref{app:eq:non_linear} & 6.0 & 10.0 & 4.0 & 6.0 & 3.0 & 3.0 & 24.0 & 19.0\\
\bottomrule
\end{tabular}
\end{adjustbox}
\end{center}
\end{table}

\begin{table}[htb!]
\begin{center}
\vspace{-3ex}
\caption{
\small
This table lists the sequence monotonic ratio (SMR, \%) on 8 molecule properties with $\gamma = 2, \tau=0.2$.
}
\vspace{-2ex}
\setlength{\tabcolsep}{10pt}
\begin{adjustbox}{max width=\textwidth}
\begin{tabular}{l l c c c c c c c c c c c}
\toprule
\multirow{2}{*}{\makecell{Model \&\\Dataset}} & \multirow{2}{*}{Edit Method} & \multirow{2}{*}{$h$} & \multicolumn{3}{c}{Physical} & \multicolumn{2}{c}{Drug-related} & \multicolumn{3}{c}{Bioactivity}\\
\cmidrule(lr){4-6} \cmidrule(lr){7-8} \cmidrule(lr){9-11}
 & & & LogP $\uparrow$ & tPSA $\uparrow$ & MW $\uparrow$ & QED $\uparrow$ & SA $\uparrow$ & DRD2 $\uparrow$ & JNK3 $\uparrow$ & GSK3$\beta$ $\uparrow$\\
\midrule
\multirow{10}{*}{\makecell{MoFlow\\ZINC250k}}
& Random & -- & 24.0 & 28.0 & 26.0 & 27.0 & 25.0 & 22.0 & 29.0 & 26.0\\
& Variance & -- & 25.0 & 25.0 & 22.0 & 23.0 & 21.0 & 22.0 & 27.0 & 23.0\\
& SeFa & -- & 20.0 & 20.0 & 23.0 & 21.0 & 22.0 & 22.0 & 21.0 & 22.0\\
& DisCo & -- & 7.0 & 7.0 & 8.0 & 7.0 & 5.0 & 8.0 & 9.0 & 9.0\\
\cmidrule(lr){2-11}
& \multirow{3}{*}{GraphCG-P}
 & \Cref{app:eq:linear_01} & 31.0 & 32.0 & 27.0 & 27.0 & 26.0 & 28.0 & 29.0 & 27.0\\
&  & \Cref{app:eq:linear_02} & 28.0 & 29.0 & 27.0 & 25.0 & 28.0 & 26.0 & 27.0 & 29.0\\
&  & \Cref{app:eq:non_linear} & 29.0 & 39.0 & 31.0 & 30.0 & 29.0 & 27.0 & 39.0 & 33.0\\
\cmidrule(lr){2-11}
& \multirow{3}{*}{GraphCG-R}
 & \Cref{app:eq:linear_01} & 28.0 & 29.0 & 28.0 & 28.0 & 27.0 & 25.0 & 34.0 & 29.0\\
&  & \Cref{app:eq:linear_02} & 28.0 & 29.0 & 27.0 & 25.0 & 28.0 & 26.0 & 27.0 & 29.0\\
&  & \Cref{app:eq:non_linear} & 0.0 & 0.0 & 0.0 & 0.0 & 0.0 & 0.0 & 0.0 & 0.0\\
\midrule
\multirow{9}{*}{\makecell{HierVAE\\ChEMBL}}
& Random & -- & 12.0 & 14.0 & 13.0 & 10.0 & 11.0 & 9.0 & 18.0 & 16.0\\
& Variance & -- & 24.0 & 16.0 & 20.0 & 24.0 & 24.0 & 18.0 & 31.0 & 26.0\\
& SeFa & -- & 18.0 & 4.0 & 35.0 & 1.0 & 14.0 & 3.0 & 7.0 & 3.0\\
\cmidrule(lr){2-11}
& \multirow{3}{*}{GraphCG-P}
 & \Cref{app:eq:linear_01} & 19.0 & 20.0 & 24.0 & 15.0 & 16.0 & 13.0 & 25.0 & 20.0\\
&  & \Cref{app:eq:linear_02} & 53.0 & 56.0 & 52.0 & 51.0 & 56.0 & 53.0 & 57.0 & 54.0\\
&  & \Cref{app:eq:non_linear} & 16.0 & 29.0 & 15.0 & 16.0 & 17.0 & 22.0 & 32.0 & 25.0\\
\cmidrule(lr){2-11}
& \multirow{3}{*}{GraphCG-R}
 & \Cref{app:eq:linear_01} & 14.0 & 14.0 & 26.0 & 8.0 & 15.0 & 12.0 & 20.0 & 16.0\\
&  & \Cref{app:eq:linear_02} & 53.0 & 54.0 & 51.0 & 50.0 & 54.0 & 52.0 & 56.0 & 54.0\\
&  & \Cref{app:eq:non_linear} & 21.0 & 21.0 & 16.0 & 18.0 & 21.0 & 17.0 & 29.0 & 27.0\\
\bottomrule
\end{tabular}
\end{adjustbox}
\end{center}
\end{table}
\clearpage

\begin{table}[htb!]
\begin{center}
\vspace{-3ex}
\caption{
\small
This table lists the sequence monotonic ratio (SMR, \%) on 8 molecule properties with $\gamma = 3, \tau=0$.
}
\vspace{-2ex}
\setlength{\tabcolsep}{10pt}
\begin{adjustbox}{max width=\textwidth}
\begin{tabular}{l l c c c c c c c c c c c}
\toprule
\multirow{2}{*}{\makecell{Model \&\\Dataset}} & \multirow{2}{*}{Edit Method} & \multirow{2}{*}{$h$} & \multicolumn{3}{c}{Physical} & \multicolumn{2}{c}{Drug-related} & \multicolumn{3}{c}{Bioactivity}\\
\cmidrule(lr){4-6} \cmidrule(lr){7-8} \cmidrule(lr){9-11}
 & & & LogP $\uparrow$ & tPSA $\uparrow$ & MW $\uparrow$ & QED $\uparrow$ & SA $\uparrow$ & DRD2 $\uparrow$ & JNK3 $\uparrow$ & GSK3$\beta$ $\uparrow$\\
\midrule

\multirow{10}{*}{\makecell{MoFlow\\ZINC250k}}
& Random & -- & 10.0 & 17.0 & 9.0 & 10.0 & 10.0 & 9.0 & 15.0 & 11.0\\
& Variance & -- & 10.0 & 10.0 & 6.0 & 7.0 & 6.0 & 6.0 & 18.0 & 11.0\\
& SeFa & -- & 1.0 & 1.0 & 4.0 & 2.0 & 3.0 & 3.0 & 2.0 & 3.0\\
& DisCo & -- & 3.0 & 3.0 & 3.0 & 2.0 & 2.0 & 3.0 & 4.0 & 4.0\\
\cmidrule(lr){2-11}
& \multirow{3}{*}{GraphCG-P}
 & \Cref{app:eq:linear_01} & 11.0 & 16.0 & 9.0 & 10.0 & 7.0 & 9.0 & 15.0 & 13.0\\
&  & \Cref{app:eq:linear_02} & 10.0 & 13.0 & 9.0 & 8.0 & 11.0 & 8.0 & 11.0 & 11.0\\
&  & \Cref{app:eq:non_linear} & 14.0 & 26.0 & 14.0 & 14.0 & 16.0 & 10.0 & 26.0 & 16.0\\
\cmidrule(lr){2-11}
& \multirow{3}{*}{GraphCG-R}
 & \Cref{app:eq:linear_01} & 10.0 & 17.0 & 8.0 & 13.0 & 10.0 & 10.0 & 16.0 & 12.0\\
&  & \Cref{app:eq:linear_02} & 10.0 & 13.0 & 9.0 & 8.0 & 11.0 & 8.0 & 11.0 & 11.0\\
&  & \Cref{app:eq:non_linear} & 0.0 & 0.0 & 0.0 & 0.0 & 0.0 & 0.0 & 0.0 & 0.0\\
\midrule
\multirow{9}{*}{\makecell{HierVAE\\ChEMBL}}
& Random & -- & 3.0 & 5.0 & 4.0 & 2.0 & 2.0 & 1.0 & 11.0 & 8.0\\
& Variance & -- & 5.0 & 6.0 & 4.0 & 7.0 & 6.0 & 4.0 & 21.0 & 17.0\\
& SeFa & -- & 1.0 & 0.0 & 2.0 & 0.0 & 0.0 & 0.0 & 1.0 & 2.0\\
\cmidrule(lr){2-11}
& \multirow{3}{*}{GraphCG-P}
 & \Cref{app:eq:linear_01} & 5.0 & 8.0 & 8.0 & 2.0 & 2.0 & 3.0 & 17.0 & 13.0\\
&  & \Cref{app:eq:linear_02} & 11.0 & 16.0 & 14.0 & 11.0 & 16.0 & 8.0 & 26.0 & 20.0\\
&  & \Cref{app:eq:non_linear} & 6.0 & 11.0 & 2.0 & 3.0 & 3.0 & 2.0 & 30.0 & 19.0\\
\cmidrule(lr){2-11}
& \multirow{3}{*}{GraphCG-R}
 & \Cref{app:eq:linear_01} & 2.0 & 6.0 & 3.0 & 2.0 & 2.0 & 2.0 & 11.0 & 9.0\\
&  & \Cref{app:eq:linear_02} & 14.0 & 17.0 & 15.0 & 12.0 & 14.0 & 13.0 & 24.0 & 21.0\\
&  & \Cref{app:eq:non_linear} & 6.0 & 9.0 & 3.0 & 5.0 & 3.0 & 3.0 & 23.0 & 19.0\\
\bottomrule
\end{tabular}
\end{adjustbox}
\end{center}
\end{table}

\begin{table}[htb!]
\begin{center}
\vspace{-3ex}
\caption{
\small
This table lists the sequence monotonic ratio (SMR, \%) on 8 molecule properties with $\gamma = 3, \tau=0.2$.
}
\vspace{-2ex}
\setlength{\tabcolsep}{10pt}
\begin{adjustbox}{max width=\textwidth}
\begin{tabular}{l l c c c c c c c c c c c}
\toprule
\multirow{2}{*}{\makecell{Model \&\\Dataset}} & \multirow{2}{*}{Edit Method} & \multirow{2}{*}{$h$} & \multicolumn{3}{c}{Physical} & \multicolumn{2}{c}{Drug-related} & \multicolumn{3}{c}{Bioactivity}\\
\cmidrule(lr){4-6} \cmidrule(lr){7-8} \cmidrule(lr){9-11}
 & & & LogP $\uparrow$ & tPSA $\uparrow$ & MW $\uparrow$ & QED $\uparrow$ & SA $\uparrow$ & DRD2 $\uparrow$ & JNK3 $\uparrow$ & GSK3$\beta$ $\uparrow$\\
\midrule

\multirow{10}{*}{\makecell{MoFlow\\ZINC250k}}
& Random & -- & 13.0 & 17.0 & 12.0 & 14.0 & 11.0 & 12.0 & 16.0 & 12.0\\
& Variance & -- & 12.0 & 11.0 & 9.0 & 9.0 & 8.0 & 8.0 & 19.0 & 14.0\\
& SeFa & -- & 1.0 & 1.0 & 4.0 & 2.0 & 3.0 & 3.0 & 2.0 & 3.0\\
& DisCo & -- & 5.0 & 6.0 & 6.0 & 5.0 & 5.0 & 6.0 & 8.0 & 8.0\\
\cmidrule(lr){2-11}
& \multirow{3}{*}{GraphCG-P}
 & \Cref{app:eq:linear_01} & 16.0 & 18.0 & 11.0 & 13.0 & 11.0 & 13.0 & 15.0 & 17.0\\
&  & \Cref{app:eq:linear_02} & 12.0 & 13.0 & 11.0 & 9.0 & 12.0 & 10.0 & 13.0 & 13.0\\
&  & \Cref{app:eq:non_linear} & 16.0 & 26.0 & 16.0 & 17.0 & 17.0 & 12.0 & 26.0 & 17.0\\
\cmidrule(lr){2-11}
& \multirow{3}{*}{GraphCG-R}
 & \Cref{app:eq:linear_01} & 12.0 & 17.0 & 10.0 & 14.0 & 12.0 & 11.0 & 18.0 & 12.0\\
&  & \Cref{app:eq:linear_02} & 12.0 & 13.0 & 11.0 & 9.0 & 12.0 & 10.0 & 13.0 & 13.0\\
&  & \Cref{app:eq:non_linear} & 0.0 & 0.0 & 0.0 & 0.0 & 0.0 & 0.0 & 0.0 & 0.0\\
\midrule
\multirow{9}{*}{\makecell{HierVAE\\ChEMBL}}
& Random & -- & 12.0 & 14.0 & 13.0 & 10.0 & 11.0 & 9.0 & 18.0 & 15.0\\
& Variance & -- & 20.0 & 15.0 & 16.0 & 20.0 & 20.0 & 14.0 & 27.0 & 21.0\\
& SeFa & -- & 18.0 & 4.0 & 35.0 & 1.0 & 14.0 & 3.0 & 7.0 & 3.0\\
\cmidrule(lr){2-11}
& \multirow{3}{*}{GraphCG-P}
 & \Cref{app:eq:linear_01} & 15.0 & 17.0 & 23.0 & 14.0 & 15.0 & 12.0 & 21.0 & 19.0\\
&  & \Cref{app:eq:linear_02} & 23.0 & 24.0 & 23.0 & 22.0 & 28.0 & 21.0 & 30.0 & 31.0\\
&  & \Cref{app:eq:non_linear} & 15.0 & 28.0 & 14.0 & 16.0 & 16.0 & 21.0 & 32.0 & 24.0\\
\cmidrule(lr){2-11}
& \multirow{3}{*}{GraphCG-R}
 & \Cref{app:eq:linear_01} & 14.0 & 14.0 & 26.0 & 8.0 & 15.0 & 12.0 & 20.0 & 16.0\\
&  & \Cref{app:eq:linear_02} & 23.0 & 27.0 & 23.0 & 23.0 & 23.0 & 24.0 & 33.0 & 32.0\\
&  & \Cref{app:eq:non_linear} & 20.0 & 20.0 & 15.0 & 17.0 & 20.0 & 16.0 & 29.0 & 27.0\\
\bottomrule
\end{tabular}
\end{adjustbox}
\end{center}
\end{table}
\clearpage

\begin{table}[htb!]
\begin{center}
\vspace{-3ex}
\caption{
\small
This table lists the sequence monotonic ratio (SMR, \%) on 8 molecule properties with $\gamma = 4, \tau=0$.
}
\vspace{-2ex}
\setlength{\tabcolsep}{10pt}
\begin{adjustbox}{max width=\textwidth}
\begin{tabular}{l l c c c c c c c c c c c}
\toprule
\multirow{2}{*}{\makecell{Model \&\\Dataset}} & \multirow{2}{*}{Edit Method} & \multirow{2}{*}{$h$} & \multicolumn{3}{c}{Physical} & \multicolumn{2}{c}{Drug-related} & \multicolumn{3}{c}{Bioactivity}\\
\cmidrule(lr){4-6} \cmidrule(lr){7-8} \cmidrule(lr){9-11}
 & & & LogP $\uparrow$ & tPSA $\uparrow$ & MW $\uparrow$ & QED $\uparrow$ & SA $\uparrow$ & DRD2 $\uparrow$ & JNK3 $\uparrow$ & GSK3$\beta$ $\uparrow$\\
\midrule

\multirow{10}{*}{\makecell{MoFlow\\ZINC250k}}
& Random & -- & 3.0 & 7.0 & 4.0 & 4.0 & 3.0 & 3.0 & 6.0 & 5.0\\
& Variance & -- & 3.0 & 4.0 & 2.0 & 2.0 & 2.0 & 2.0 & 7.0 & 4.0\\
& SeFa & -- & 0.0 & 0.0 & 0.0 & 0.0 & 0.0 & 0.0 & 0.0 & 0.0\\
& DisCo & -- & 1.0 & 1.0 & 1.0 & 1.0 & 2.0 & 1.0 & 2.0 & 2.0\\
\cmidrule(lr){2-11}
& \multirow{3}{*}{GraphCG-P}
 & \Cref{app:eq:linear_01} & 6.0 & 6.0 & 4.0 & 5.0 & 2.0 & 5.0 & 6.0 & 7.0\\
&  & \Cref{app:eq:linear_02} & 3.0 & 4.0 & 2.0 & 1.0 & 3.0 & 2.0 & 5.0 & 3.0\\
&  & \Cref{app:eq:non_linear} & 4.0 & 12.0 & 5.0 & 4.0 & 4.0 & 3.0 & 11.0 & 6.0\\
\cmidrule(lr){2-11}
& \multirow{3}{*}{GraphCG-R}
 & \Cref{app:eq:linear_01} & 4.0 & 6.0 & 2.0 & 3.0 & 4.0 & 3.0 & 7.0 & 3.0\\
&  & \Cref{app:eq:linear_02} & 3.0 & 4.0 & 2.0 & 1.0 & 3.0 & 2.0 & 5.0 & 3.0\\
&  & \Cref{app:eq:non_linear} & 0.0 & 0.0 & 0.0 & 0.0 & 0.0 & 0.0 & 0.0 & 0.0\\
\midrule
\multirow{9}{*}{\makecell{HierVAE\\ChEMBL}}
& Random & -- & 2.0 & 4.0 & 3.0 & 2.0 & 1.0 & 1.0 & 9.0 & 7.0\\
& Variance & -- & 2.0 & 6.0 & 3.0 & 5.0 & 4.0 & 2.0 & 17.0 & 14.0\\
& SeFa & -- & 1.0 & 0.0 & 2.0 & 0.0 & 0.0 & 0.0 & 1.0 & 2.0\\
\cmidrule(lr){2-11}
& \multirow{3}{*}{GraphCG-P}
 & \Cref{app:eq:linear_01} & 4.0 & 6.0 & 5.0 & 1.0 & 2.0 & 2.0 & 15.0 & 11.0\\
&  & \Cref{app:eq:linear_02} & 5.0 & 7.0 & 6.0 & 5.0 & 5.0 & 4.0 & 17.0 & 15.0\\
&  & \Cref{app:eq:non_linear} & 5.0 & 10.0 & 1.0 & 2.0 & 2.0 & 2.0 & 28.0 & 18.0\\
\cmidrule(lr){2-11}
& \multirow{3}{*}{GraphCG-R}
 & \Cref{app:eq:linear_01} & 1.0 & 5.0 & 3.0 & 2.0 & 1.0 & 2.0 & 10.0 & 9.0\\
&  & \Cref{app:eq:linear_02} & 4.0 & 9.0 & 6.0 & 5.0 & 5.0 & 6.0 & 17.0 & 15.0\\
&  & \Cref{app:eq:non_linear} & 4.0 & 7.0 & 2.0 & 4.0 & 2.0 & 2.0 & 20.0 & 18.0\\
\bottomrule
\end{tabular}
\end{adjustbox}
\end{center}
\end{table}

\begin{table}[htb!]
\begin{center}
\vspace{-3ex}
\caption{
\small
This table lists the sequence monotonic ratio (SMR, \%) on 8 molecule properties with $\gamma = 4, \tau=0.2$.
}
\vspace{-2ex}
\setlength{\tabcolsep}{10pt}
\begin{adjustbox}{max width=\textwidth}
\begin{tabular}{l l c c c c c c c c c c c}
\toprule
\multirow{2}{*}{\makecell{Model \&\\Dataset}} & \multirow{2}{*}{Edit Method} & \multirow{2}{*}{$h$} & \multicolumn{3}{c}{Physical} & \multicolumn{2}{c}{Drug-related} & \multicolumn{3}{c}{Bioactivity}\\
\cmidrule(lr){4-6} \cmidrule(lr){7-8} \cmidrule(lr){9-11}
 & & & LogP $\uparrow$ & tPSA $\uparrow$ & MW $\uparrow$ & QED $\uparrow$ & SA $\uparrow$ & DRD2 $\uparrow$ & JNK3 $\uparrow$ & GSK3$\beta$ $\uparrow$\\
\midrule

\multirow{10}{*}{\makecell{MoFlow\\ZINC250k}}
& Random & -- & 6.0 & 8.0 & 7.0 & 8.0 & 6.0 & 6.0 & 7.0 & 6.0\\
& Variance & -- & 6.0 & 6.0 & 5.0 & 4.0 & 4.0 & 4.0 & 8.0 & 7.0\\
& SeFa & -- & 0.0 & 0.0 & 0.0 & 0.0 & 0.0 & 0.0 & 0.0 & 0.0\\
& DisCo & -- & 4.0 & 4.0 & 4.0 & 4.0 & 5.0 & 6.0 & 6.0 & 5.0\\
\cmidrule(lr){2-11}
& \multirow{3}{*}{GraphCG-P}
 & \Cref{app:eq:linear_01} & 10.0 & 11.0 & 6.0 & 9.0 & 9.0 & 8.0 & 7.0 & 7.0\\
&  & \Cref{app:eq:linear_02} & 4.0 & 6.0 & 5.0 & 5.0 & 7.0 & 3.0 & 5.0 & 5.0\\
&  & \Cref{app:eq:non_linear} & 8.0 & 12.0 & 8.0 & 7.0 & 6.0 & 6.0 & 11.0 & 10.0\\
\cmidrule(lr){2-11}
& \multirow{3}{*}{GraphCG-R}
 & \Cref{app:eq:linear_01} & 6.0 & 6.0 & 5.0 & 6.0 & 6.0 & 5.0 & 8.0 & 4.0\\
&  & \Cref{app:eq:linear_02} & 4.0 & 6.0 & 5.0 & 5.0 & 7.0 & 3.0 & 5.0 & 5.0\\
&  & \Cref{app:eq:non_linear} & 0.0 & 0.0 & 0.0 & 0.0 & 0.0 & 0.0 & 0.0 & 0.0\\
\midrule
\multirow{9}{*}{\makecell{HierVAE\\ChEMBL}}
& Random & -- & 10.0 & 14.0 & 13.0 & 10.0 & 11.0 & 9.0 & 18.0 & 15.0\\
& Variance & -- & 16.0 & 15.0 & 14.0 & 16.0 & 19.0 & 11.0 & 23.0 & 18.0\\
& SeFa & -- & 18.0 & 4.0 & 35.0 & 1.0 & 14.0 & 3.0 & 7.0 & 3.0\\
\cmidrule(lr){2-11}
& \multirow{3}{*}{GraphCG-P}
 & \Cref{app:eq:linear_01} & 14.0 & 16.0 & 20.0 & 13.0 & 15.0 & 12.0 & 19.0 & 17.0\\
&  & \Cref{app:eq:linear_02} & 17.0 & 19.0 & 17.0 & 20.0 & 17.0 & 19.0 & 27.0 & 28.0\\
&  & \Cref{app:eq:non_linear} & 14.0 & 27.0 & 13.0 & 15.0 & 15.0 & 21.0 & 30.0 & 24.0\\
\cmidrule(lr){2-11}
& \multirow{3}{*}{GraphCG-R}
 & \Cref{app:eq:linear_01} & 13.0 & 13.0 & 25.0 & 8.0 & 14.0 & 11.0 & 19.0 & 16.0\\
&  & \Cref{app:eq:linear_02} & 19.0 & 20.0 & 21.0 & 19.0 & 18.0 & 18.0 & 24.0 & 26.0\\
&  & \Cref{app:eq:non_linear} & 20.0 & 18.0 & 14.0 & 16.0 & 20.0 & 16.0 & 29.0 & 26.0\\
\bottomrule
\end{tabular}
\end{adjustbox}
\end{center}
\end{table}

\clearpage

\subsection{Visualization}

\begin{figure}[htb]
\centering
    \begin{subfigure}[\small Steerable factor: number of halogens.]
    {\includegraphics[width=0.47\linewidth]{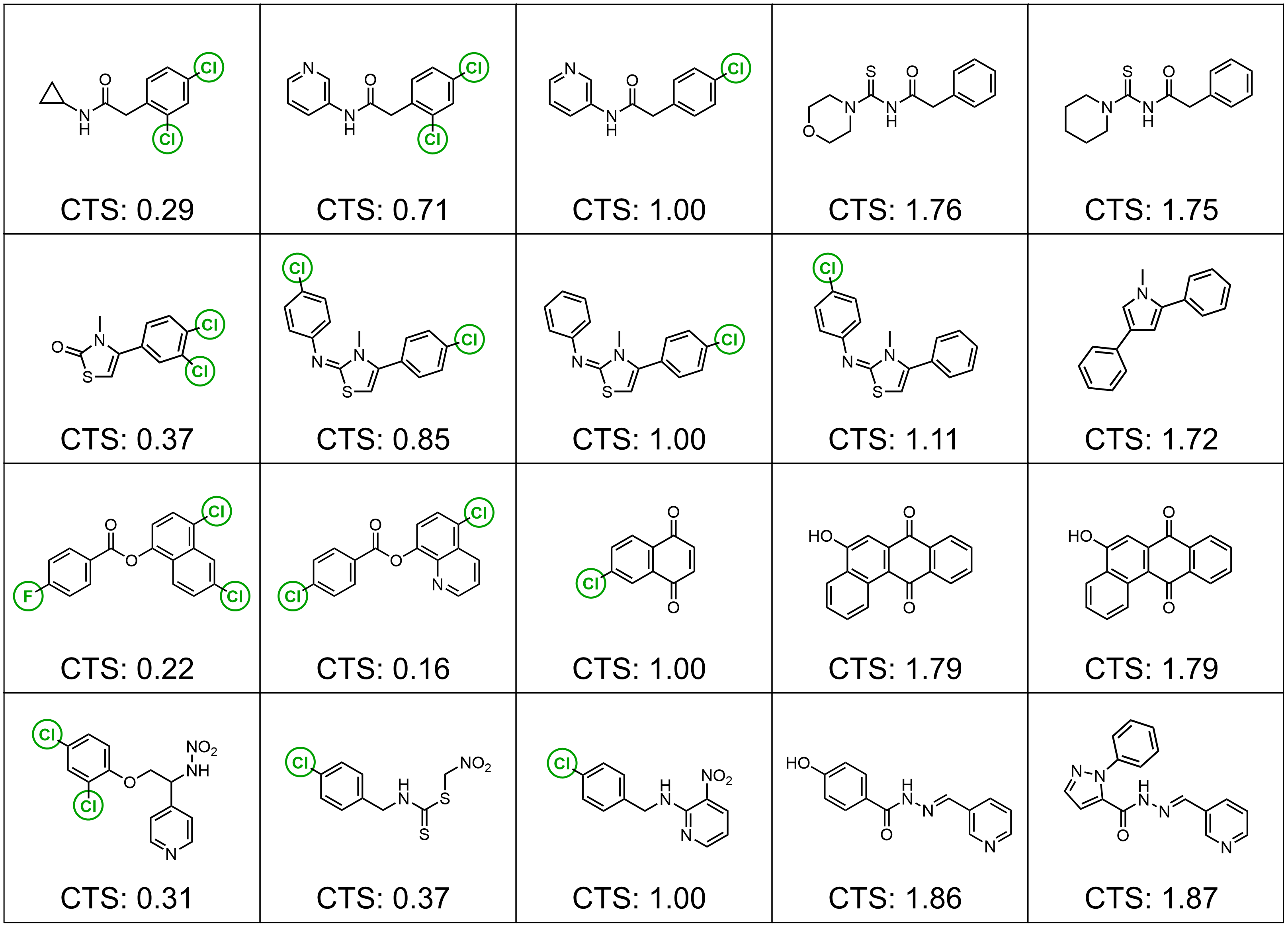}
    \label{fig:path_generation_example_hiervae_appd_a}
    }
    \end{subfigure}
\hfill
\begin{subfigure}[\small Steerable factor: number of hydroxyls.]
     {\includegraphics[width=0.47\linewidth]{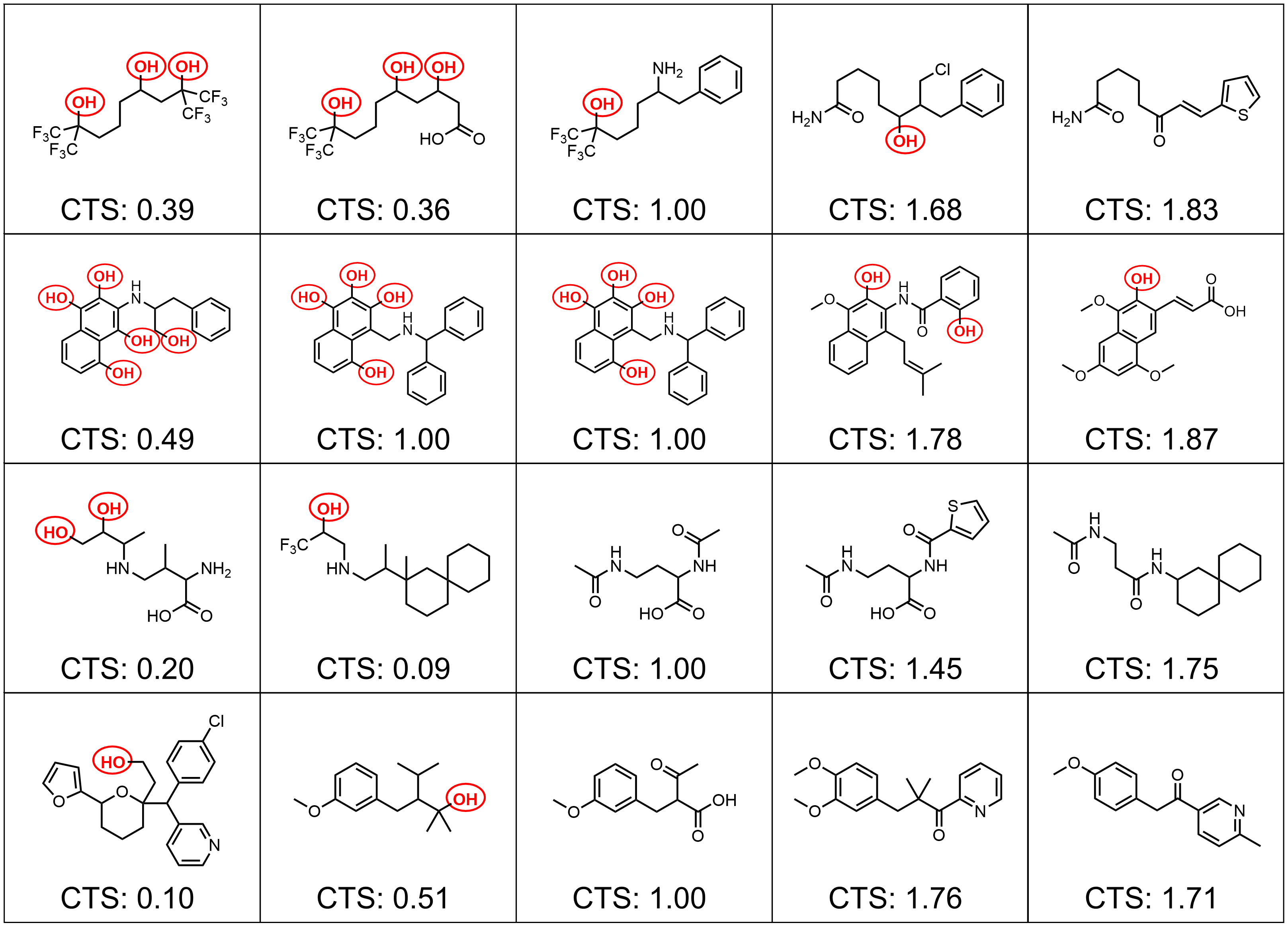}
     \label{fig:path_generation_example_hiervae_appd_b}
     }
     \end{subfigure}
\\
\vspace{-2ex}
\begin{subfigure}[\small Steerable factor: number of amides.]
    {\includegraphics[width=0.47\linewidth]{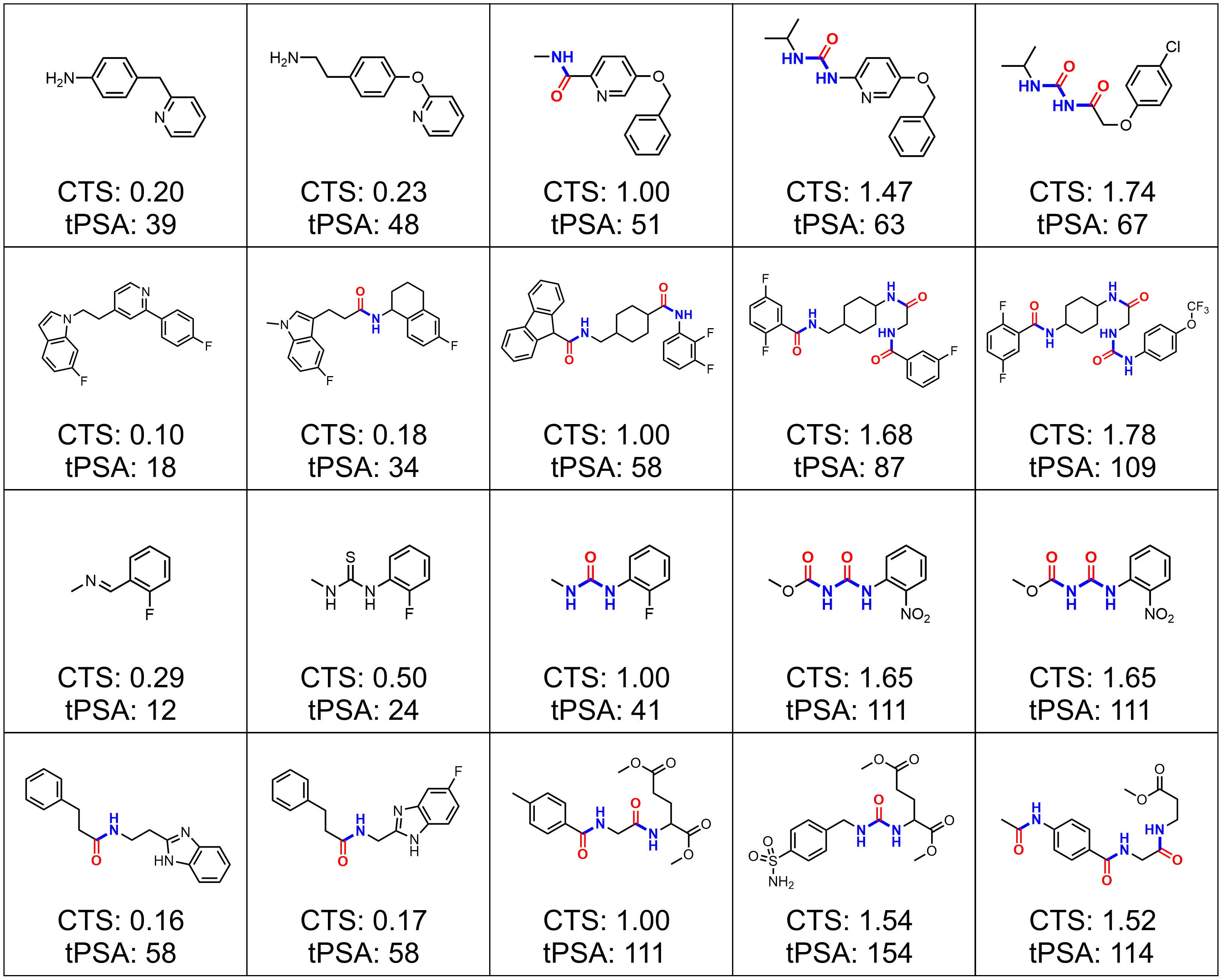}
    \label{fig:path_generation_example_hiervae_appd_c}
    }
    \end{subfigure}
\hfill
\begin{subfigure}[\small Steerable factor: chain length.]
    {\includegraphics[width=0.47\linewidth]{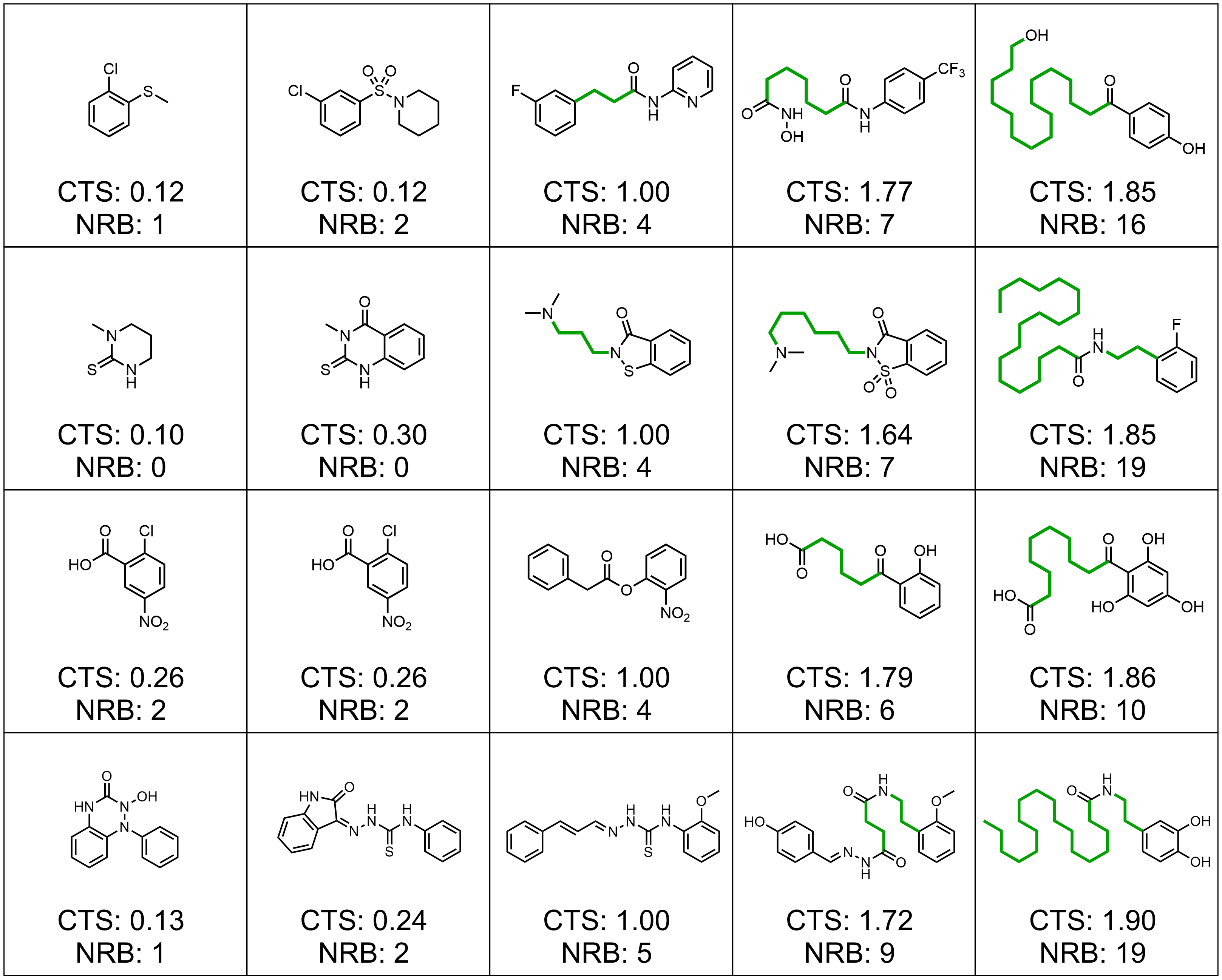}
    \label{fig:path_generation_example_hiervae_appd_d}
    }
    \end{subfigure}
\\
\vspace{-2ex}
\caption{
\small
\model\, for molecular graph editing. 
We visualize the output molecules and CTS on four directions with four sequences each, where each sequence consists of five steps. The center point is the anchor molecule, and the other four points correspond to step size with -3, -1.8, 1.8, and 3 respectively.
\Cref{fig:path_generation_example_hiervae_appd_a}~to~\Cref{fig:path_generation_example_hiervae_appd_c} show how functional groups in the molecules can be viewed as the steerable factors as they change along the path, such as halogen atoms, hydroxyl groups and amides. \Cref{fig:path_generation_example_hiervae_appd_d} illustrates the effect on the steerable factor on the length of flexible chains in the molecules. Notebly, certain properties change together with molecular structures, like topological polar surface area (tPSA) and number of rotatable bonds (NRB).}
\label{fig:path_generation_molecule_appd}
\vspace{-2ex}
\end{figure}

For a more comprehensive visualization of the steerable factors in molecular graphs, we demonstrate 16 molecular graph paths along the 4 selected directions in~\Cref{fig:path_generation_molecule_appd}, and the backbone DGM is HierVAE pretrained on ChEMBL. The CTS holds good monotonic trend in all these sequences. Each direction shows certain unique changes in the molecular structures, {\ie}, the steerable factors in molecules. Some structural changes are reflected in molecular properties. We expand all the details below.
In~\Cref{fig:path_generation_example_hiervae_appd_a} and~\Cref{fig:path_generation_example_hiervae_appd_b}, the number of halogen atoms and hydroxyl groups (in alcohols and phenols) in the molecules decrease from left to right, respectively. 
In~\Cref{fig:path_generation_example_hiervae_appd_c}, the number of amides in the molecules increases along the path. As a result, the topological polar surface area (tPSA) of the molecules increase accordingly, which is a key molecular property for the prediction of drug transport properties, {\eg}, permeability~\citep{ertl2000fast}.
In~\Cref{fig:path_generation_example_hiervae_main_d}, the flexible chain length, marked by the number of ethylene (CH$_2$CH$_2$) units, increases from left to right. Since the number of rotatable bonds (NRB) measures the molecular flexibility, it also increases accordingly~\citep{veber2002molecular}.

\newpage
Since we have determined four directions with semantic information matching with the domain knowledge (fragments), then we can check if the disentanglement measure changes before and after editing. We show the results in~\Cref{tab:app:GraphCG_disentanglement_evaluation}.

\begin{table}[htb]
\caption{\small Disentanglement measure before and after editing. The corresponding model is the GraphCG-P with \Cref{app:eq:linear_02}, and the backbone generative model is HierVAE pretrained on ChEMBL. The better performance is marked in \textbf{bold}.}
\label{tab:app:GraphCG_disentanglement_evaluation}
\centering
\begin{adjustbox}{max width=\textwidth}
\begin{tabular}{l l c c c c c c c}
\toprule
Fragment & & BetaVAE $\uparrow$ & MIG $\uparrow$ & SAP $\uparrow$ \\
\midrule
\multirow{2}{*}{Halogen}
& after editing & 0.617 &  0.010 & \textbf{0.017}\\
& before editing & \textbf{0.950} &  \textbf{0.062} & \textbf{0.017}\\
\midrule

\multirow{2}{*}{Hydroxyls}
& after editing & 0.833 &  0.031 & 0.017\\
& before editing & \textbf{0.933} &  \textbf{0.113} & \textbf{0.067}\\
\midrule

\multirow{2}{*}{Amide}
& after editing & 0.400 &  0.041 & \textbf{0.017}\\
& before editing & \textbf{0.933} &  \textbf{0.136} & \textbf{0.017}\\
\midrule

\multirow{2}{*}{Chain length}
& after editing & 0.400 &  \textbf{0.051} & 0.000\\
& before editing & \textbf{0.700} &  0.020 & \textbf{0.017}\\
\bottomrule
\end{tabular}
\end{adjustbox}
\end{table}

\clearpage

\subsection{Ablation Studies on Coefficients $c_1, c_2, c_3$}

\begin{table}[htb!]
\begin{center}
\vspace{-3ex}
\caption{
\small
Ablation studies on coefficients $c_1, c_2, c_3$. Backbone DGM is MoFlow, dataset is ZINC250K, and the editing model is GraphCG-R with \Cref{app:eq:linear_01}. The optimal values are $c_1=2, c_2=1, c_3=1$, and they are reported in~\Cref{tab:main_results_molecules,app:tab:SMR_CTS_MoFlow}.
}
\label{tab:main_results_molecules_ablation_MoFlow}
\small
\setlength{\tabcolsep}{10pt}
\begin{adjustbox}{max width=\textwidth}
\begin{tabular}{lll cccccccc}
\toprule
\multirow{3}{*}{$c_1$} & \multirow{3}{*}{$c_2$} & \multirow{3}{*}{$c_3$} & \multicolumn{4}{c}{\textbf{Tanimoto top-1}} & \multicolumn{4}{c}{\textbf{Tanimoto top-3}}\\
\cmidrule(lr){4-7} \cmidrule(lr){8-11}
 & & & \multicolumn{2}{c}{3} & \multicolumn{2}{c}{4} & \multicolumn{2}{c}{3} & \multicolumn{2}{c}{4}\\
\cmidrule(lr){4-5} \cmidrule(lr){6-7} \cmidrule(lr){8-9} \cmidrule(lr){10-11}
 & & & 0 & 0.2 & 0 & 0.2 & 0 & 0.2 & 0 & 0.2\\
\midrule

1 & 0 & 0 & 22.0 & 23.0 & 11.0 & 13.0 & 19.7 & 21.0 & 10.7 & 12.7\\
1 & 0 & 1 & 19.0 & 20.0 & 10.0 & 12.0 & 18.0 & 19.0 & 9.0 & 10.7\\
1 & 1 & 0 & 24.0 & 26.0 & 11.0 & 13.0 & 21.3 & 24.7 & 9.3 & 12.3\\
1 & 1 & 1 & 25.0 & 26.0 & 11.0 & 14.0 & 21.7 & 24.3 & 10.3 & 13.3\\
2 & 0 & 0 & 20.0 & 22.0 & 9.0 & 11.0 & 19.0 & 20.7 & 8.7 & 10.7\\
2 & 0 & 1 & 21.0 & 21.0 & 9.0 & 12.0 & 19.0 & 20.0 & 8.7 & 10.3\\
2 & 1 & 0 & 24.0 & 26.0 & 11.0 & 13.0 & 21.3 & 24.0 & 9.3 & 12.0\\
2 & 1 & 1 & 25.0 & 26.0 & 11.0 & 14.0 & 22.0 & 24.3 & 10.3 & 13.3\\

\bottomrule

\end{tabular}
\end{adjustbox}
\end{center}
\vspace{-4ex}
\end{table}

\begin{table}[htb!]
\begin{center}
\caption{
\small
Ablation studies on coefficients $c_1, c_2, c_3$. Backbone DGM is HierVAE, dataset is ChEMBL, and the editing model is GraphCG-P with \Cref{app:eq:linear_02}. The optimal values are $c_1=2, c_2=1, c_3=1$, and they are reported in~\Cref{tab:main_results_molecules,app:tab:SMR_CTS_HierVAE}.
}
\label{tab:main_results_molecules_ablation_HierVAE}
\small
\setlength{\tabcolsep}{10pt}
\begin{adjustbox}{max width=\textwidth}
\begin{tabular}{lll cccccccc}
\toprule
\multirow{3}{*}{$c_1$} & \multirow{3}{*}{$c_2$} & \multirow{3}{*}{$c_3$} & \multicolumn{4}{c}{\textbf{Tanimoto top-1}} & \multicolumn{4}{c}{\textbf{Tanimoto top-3}}\\
\cmidrule(lr){4-7} \cmidrule(lr){8-11}
 & & & \multicolumn{2}{c}{3} & \multicolumn{2}{c}{4} & \multicolumn{2}{c}{3} & \multicolumn{2}{c}{4}\\
\cmidrule(lr){4-5} \cmidrule(lr){6-7} \cmidrule(lr){8-9} \cmidrule(lr){10-11}
 & & & 0 & 0.2 & 0 & 0.2 & 0 & 0.2 & 0 & 0.2\\
\midrule

1 & 0 & 0 & 28.0 & 55.0 & 16.0 & 55.0 & 23.3 & 54.3 & 16.0 & 54.3\\
1 & 0 & 1 & 28.0 & 55.0 & 16.0 & 55.0 & 23.3 & 54.3 & 16.0 & 54.3\\
1 & 1 & 0 & 34.0 & 61.0 & 26.0 & 55.0 & 32.0 & 59.0 & 23.3 & 53.3\\
1 & 1 & 1 & 34.0 & 61.0 & 26.0 & 55.0 & 32.0 & 59.0 & 23.3 & 53.3\\
2 & 0 & 0 & 28.0 & 55.0 & 16.0 & 55.0 & 23.3 & 54.3 & 16.0 & 54.3\\
2 & 0 & 1 & 28.0 & 55.0 & 16.0 & 55.0 & 23.3 & 54.3 & 16.0 & 54.3\\
2 & 1 & 0 & 40.0 & 73.0 & 32.0 & 65.0 & 36.0 & 64.3 & 26.3 & 57.7\\
2 & 1 & 1 & 40.0 & 73.0 & 32.0 & 65.0 & 36.0 & 64.3 & 26.3 & 57.7\\

\bottomrule

\end{tabular}
\end{adjustbox}
\end{center}
\vspace{-4ex}
\end{table}

\clearpage
\newpage
\section{Results: Point Clouds} \label{app:sec:point_clouds_results}

Here we compare two editing functions, {\ie}, the key function design in \model{}. We provide the visualizations for the linear editing function in~\Cref{app:eq:linear_02}, and non-linear editing function in~\Cref{app:eq:non_linear}.

First, we want to highlight that all the samples are generated randomly. In the linear case in~\Cref{app:sec:point_clouds_linear_results}, we can observe that the shape of the airplanes, cars, and chairs, are steerable using \model. We also find it interesting that \model{} can steer more finger-trained factors, like modifying the airplane engines. However, in the non-linear case, the diversity of the edited data is smaller. This can be observed from the middle columns in~\Cref{app:sec:point_clouds_nonlinear_results}. We conjecture that this is also related to the backbone DGMs and datasets since \model{} on molecular data does not have this issue. We will leave this for future exploration.

\subsection{Linear Editing Function} \label{app:sec:point_clouds_linear_results}

\begin{figure}[H]
\centering
    \begin{subfigure}[Steerable factor: size (direction 1).]
    {\includegraphics[width=0.47\linewidth]{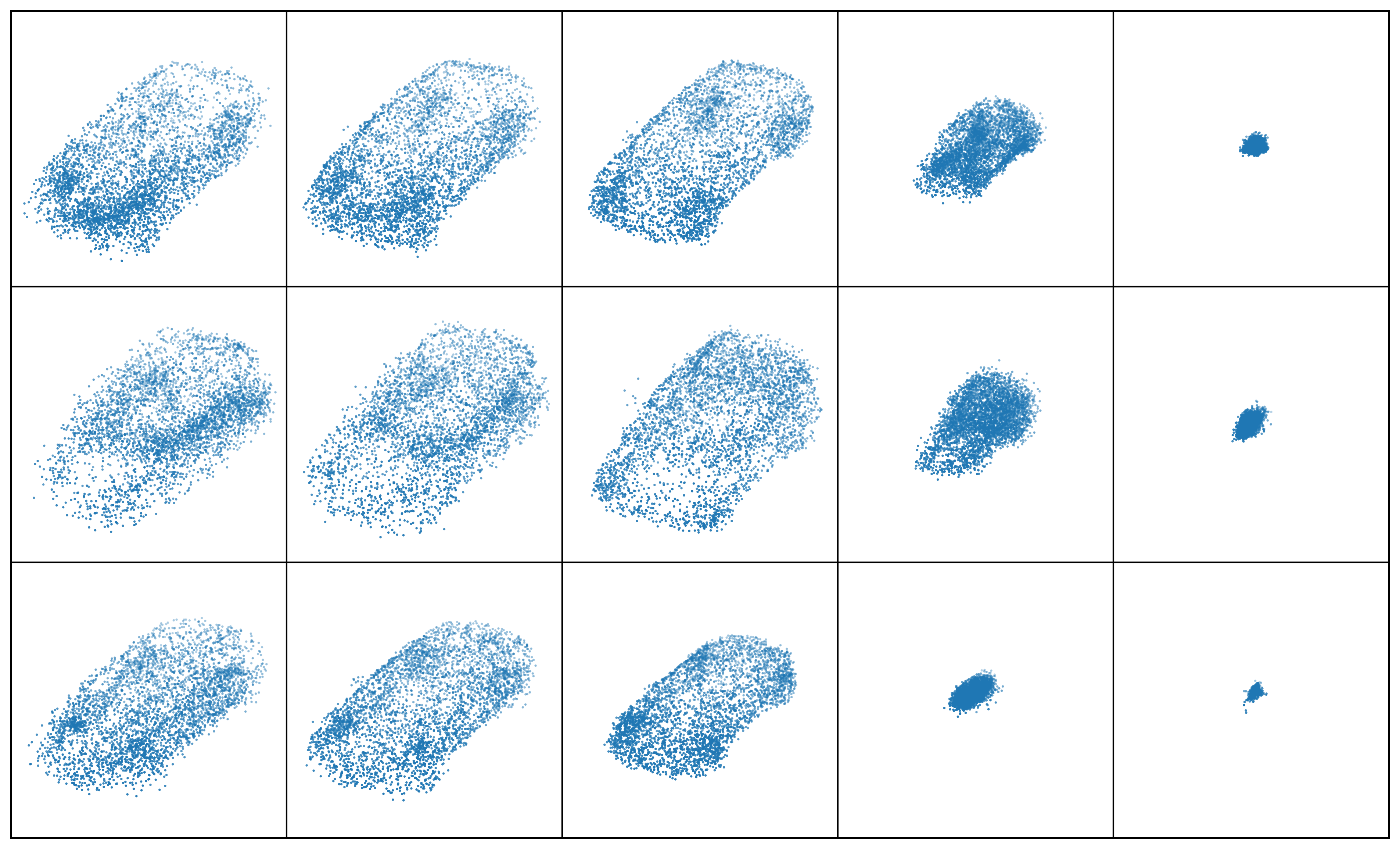}
    }
    \end{subfigure}
\hfill
    \begin{subfigure}[Steerable factor: size (direction 2).]
    {\includegraphics[width=0.47\linewidth]{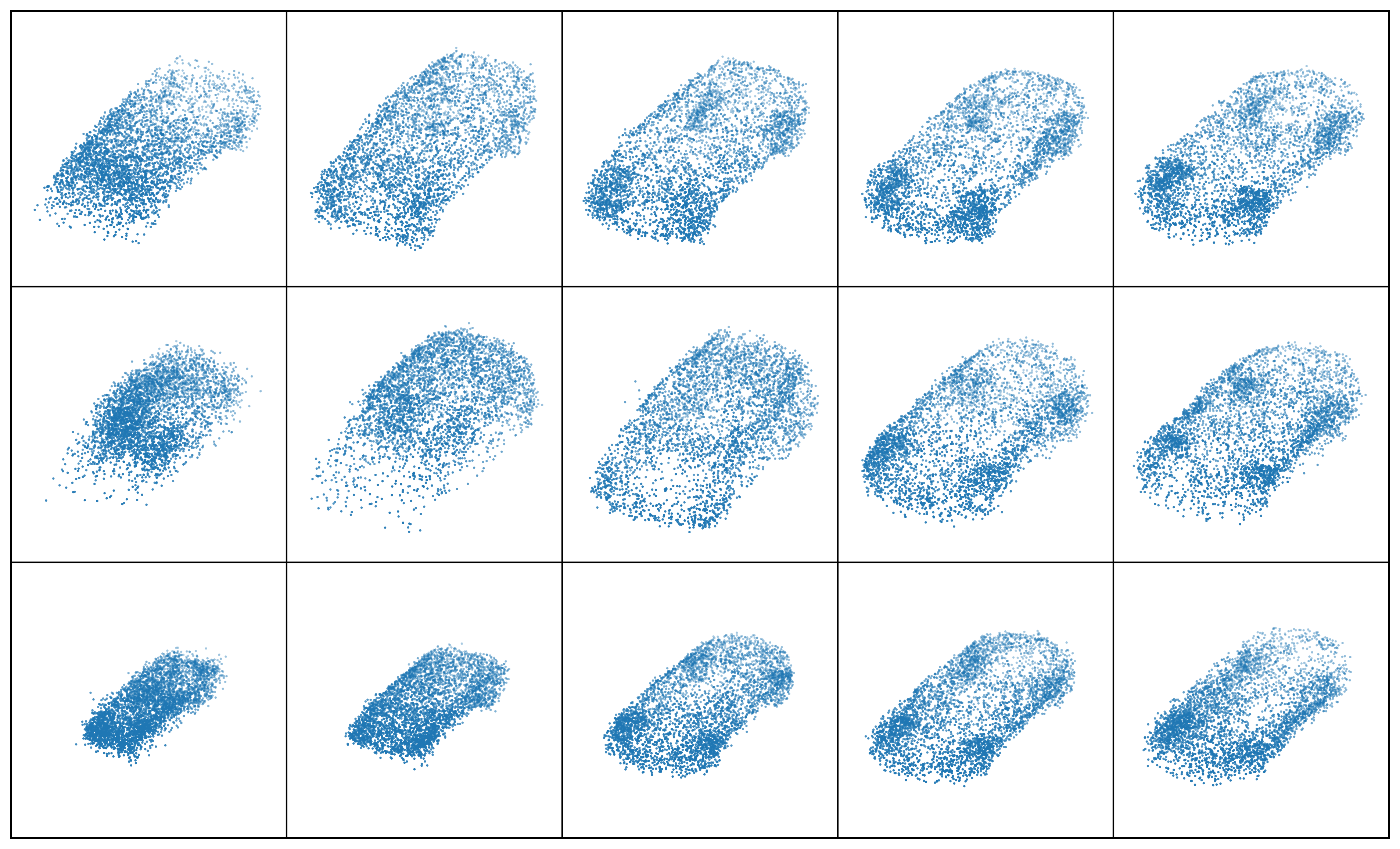}
    }
    \end{subfigure}
\\
\vspace{-2ex}
\caption{
\model{} for point clouds (Car) editing. The steerable factors on this dataset are not obvious, and here we only plot the car size editable with two directions.
}
\end{figure}

\begin{figure}[H]
\centering
    \begin{subfigure}[Steerable factor: leg height (direction 1).]
    {\includegraphics[width=0.47\linewidth]{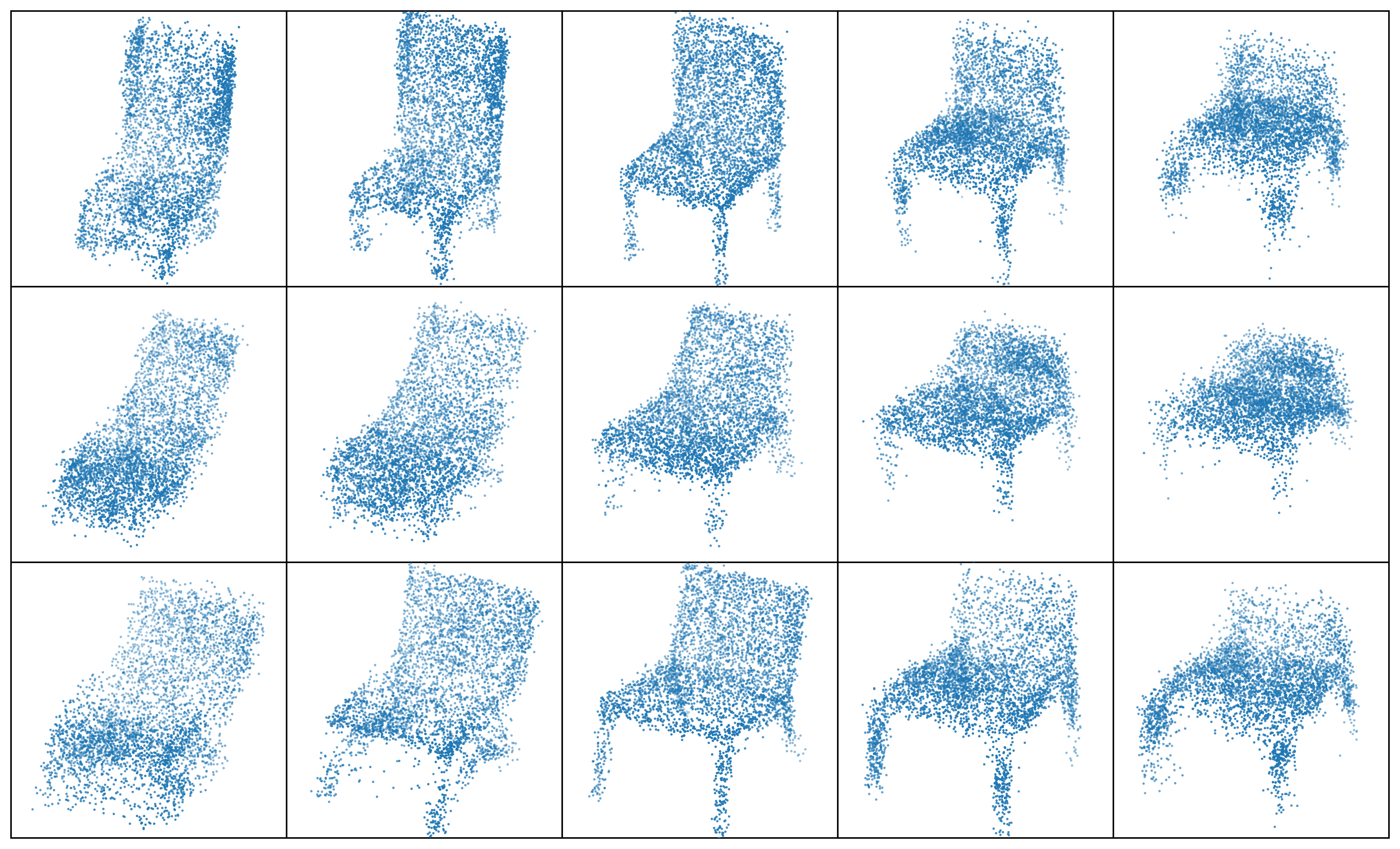}
    }
    \end{subfigure}
\hfill
    \begin{subfigure}[Steerable factor: seat size (direction 2).]
    {\includegraphics[width=0.47\linewidth]{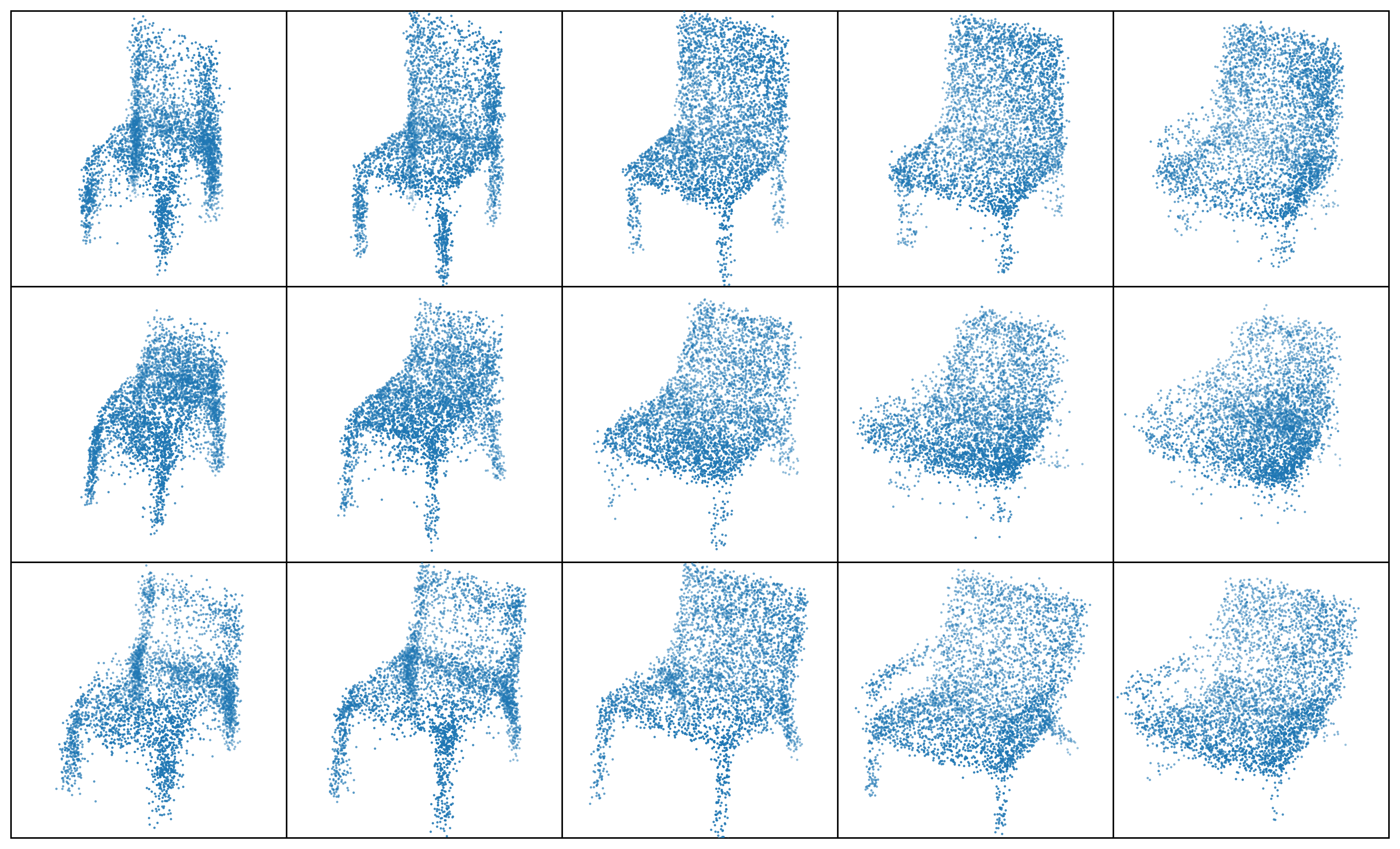}
    }
    \end{subfigure}
\\
\vspace{-2ex}
\caption{
\model{} for point clouds (Chair) editing. It can successfully reflect these steerable factors: leg height, and seat size.
}
\end{figure}
\clearpage

\begin{figure}[H]
\centering
    \begin{subfigure}[Steerable factor: engine (direction 1).]
    {\includegraphics[width=0.47\linewidth]{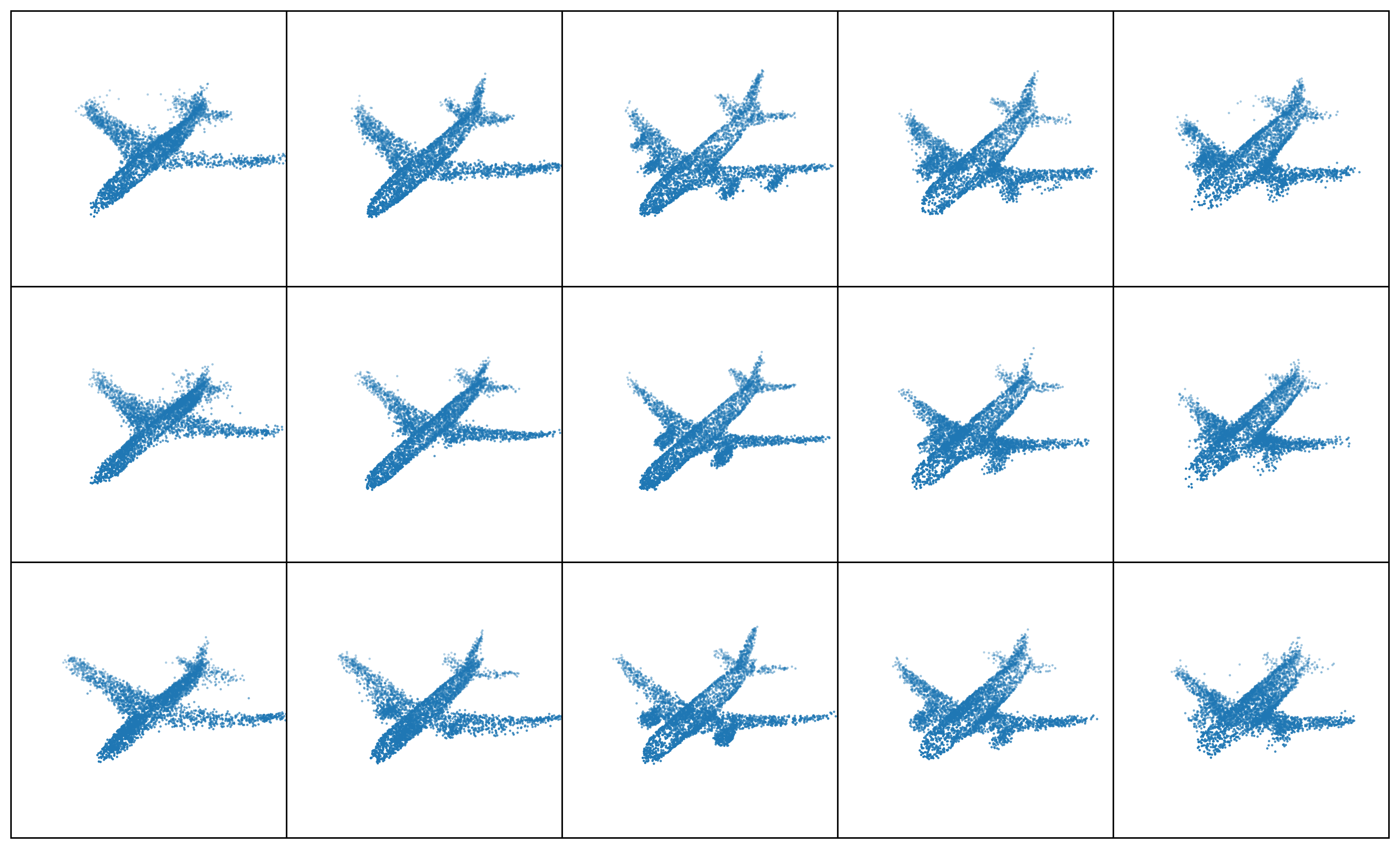}
    \label{app:fig:airplane_data_3_dir_14_engine_removal}
    }
    \end{subfigure}
\hfill
    \begin{subfigure}[Steerable factor: engine (direction 1).]
    {\includegraphics[width=0.47\linewidth]{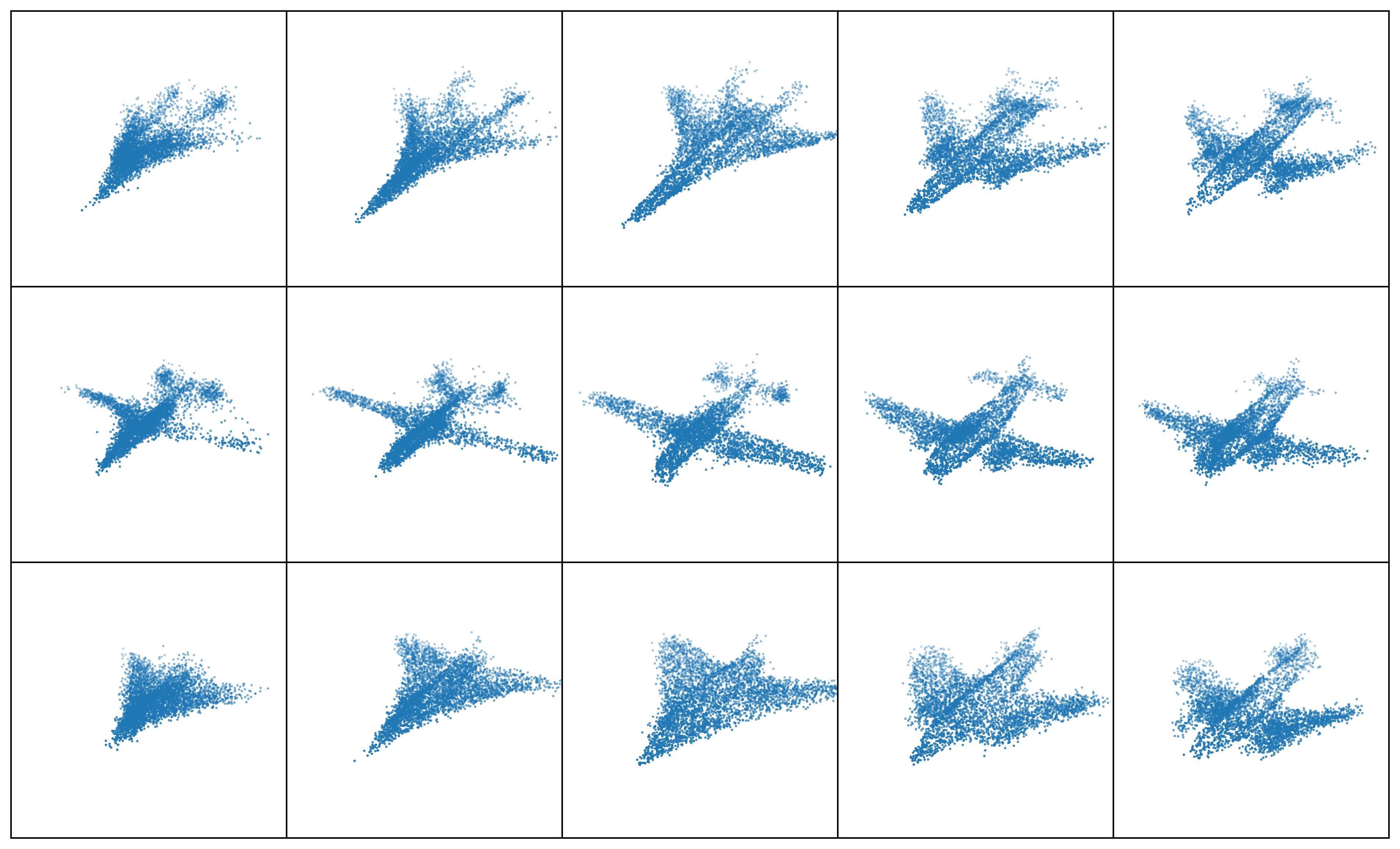}
    \label{app:fig:airplane_data_3_dir_14_engine_add}
    }
    \end{subfigure}
\\
\vspace{-1ex}
    \begin{subfigure}[Steerable factor: fuselage length (direction 2).]
    {\includegraphics[width=0.47\linewidth]{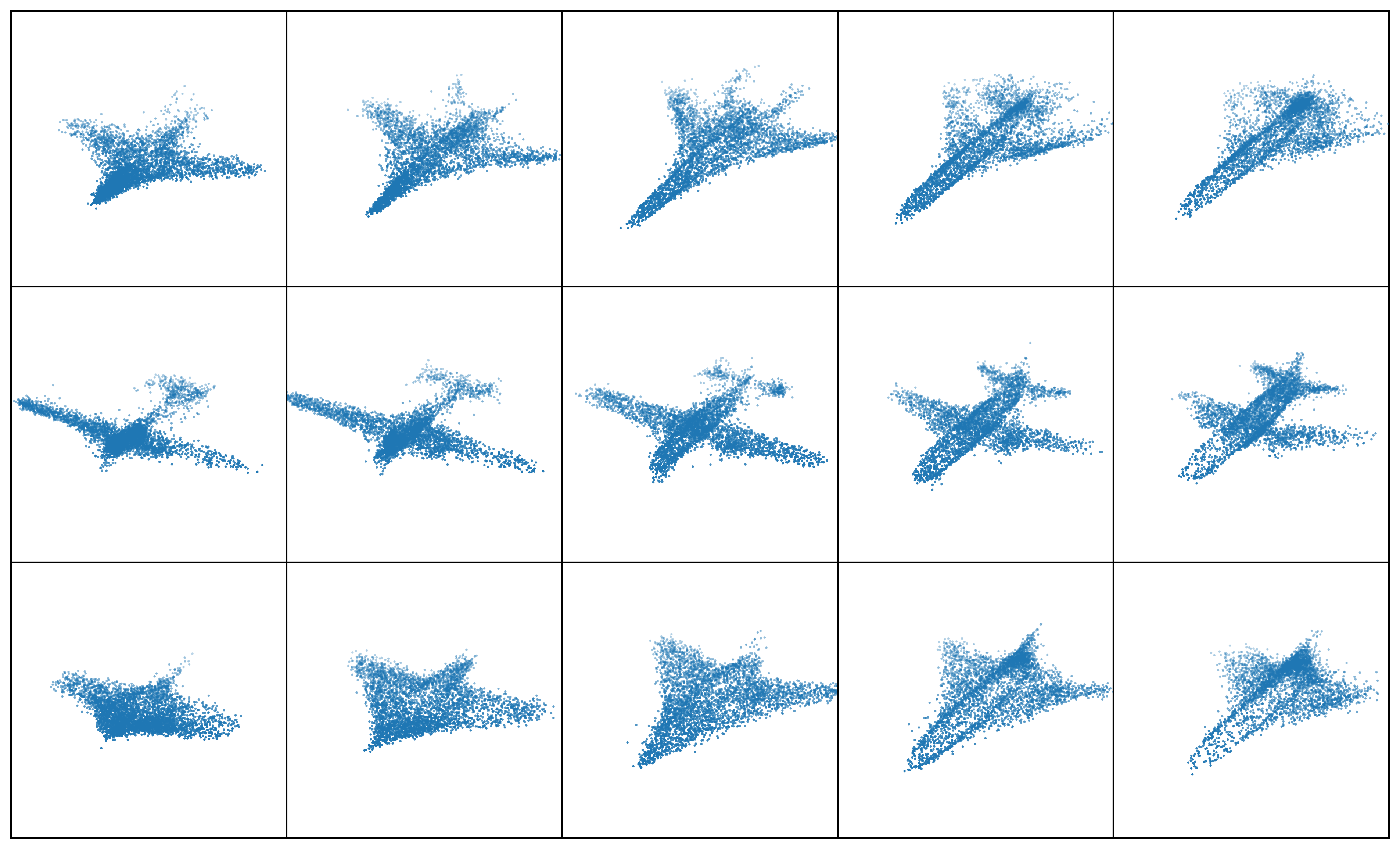}
    \label{app:fig:airplane_data_3_dir_0}
    }
    \end{subfigure}
\hfill
    \begin{subfigure}[Steerable factor: wing size (direction 3).]
    {\includegraphics[width=0.47\linewidth]{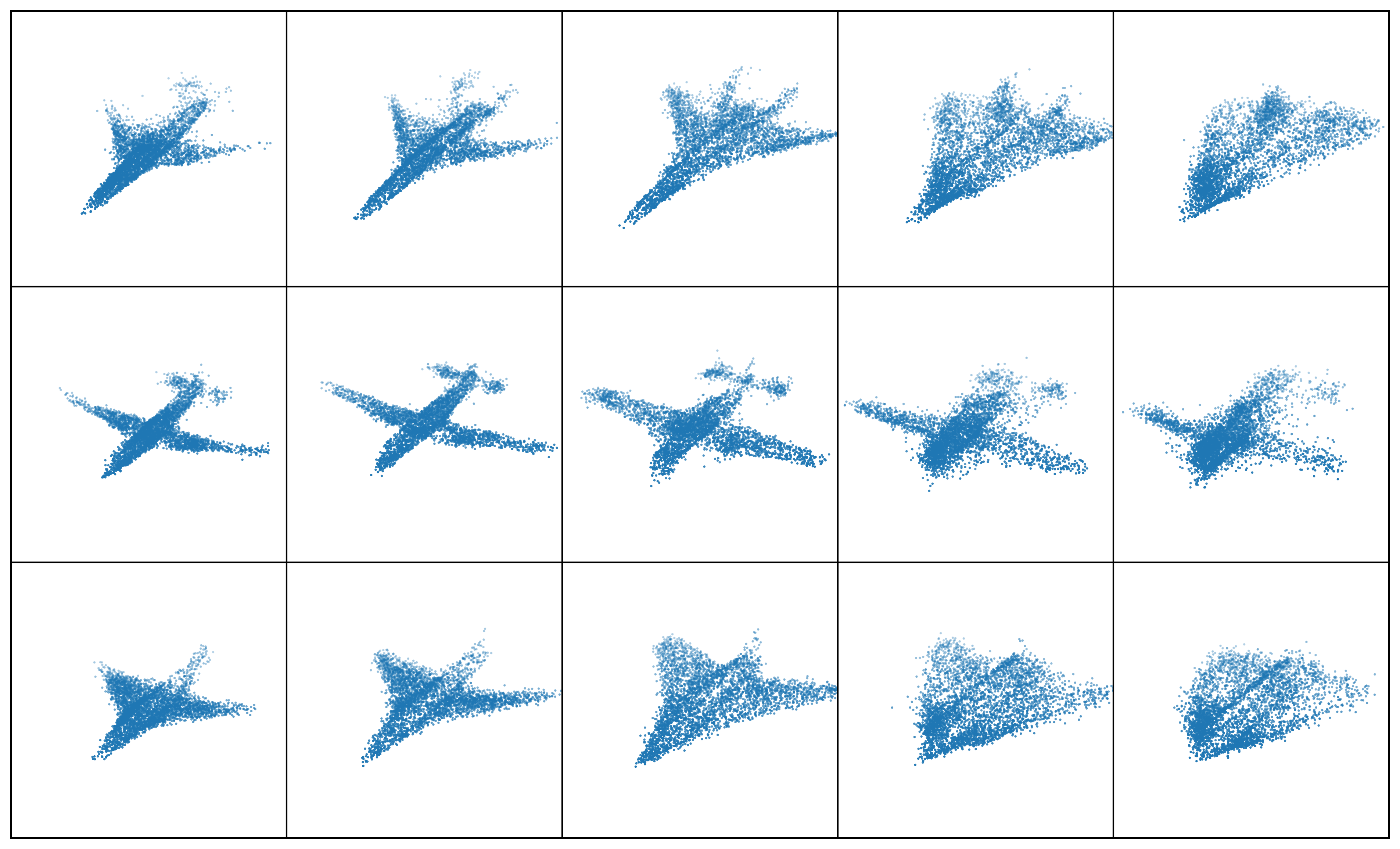}
    \label{app:fig:airplane_data_3_dir_7}
    }
    \end{subfigure}
\\
\vspace{-1ex}
    \begin{subfigure}[Steerable factor: wing shape(direction 4).]
    {\includegraphics[width=0.47\linewidth]{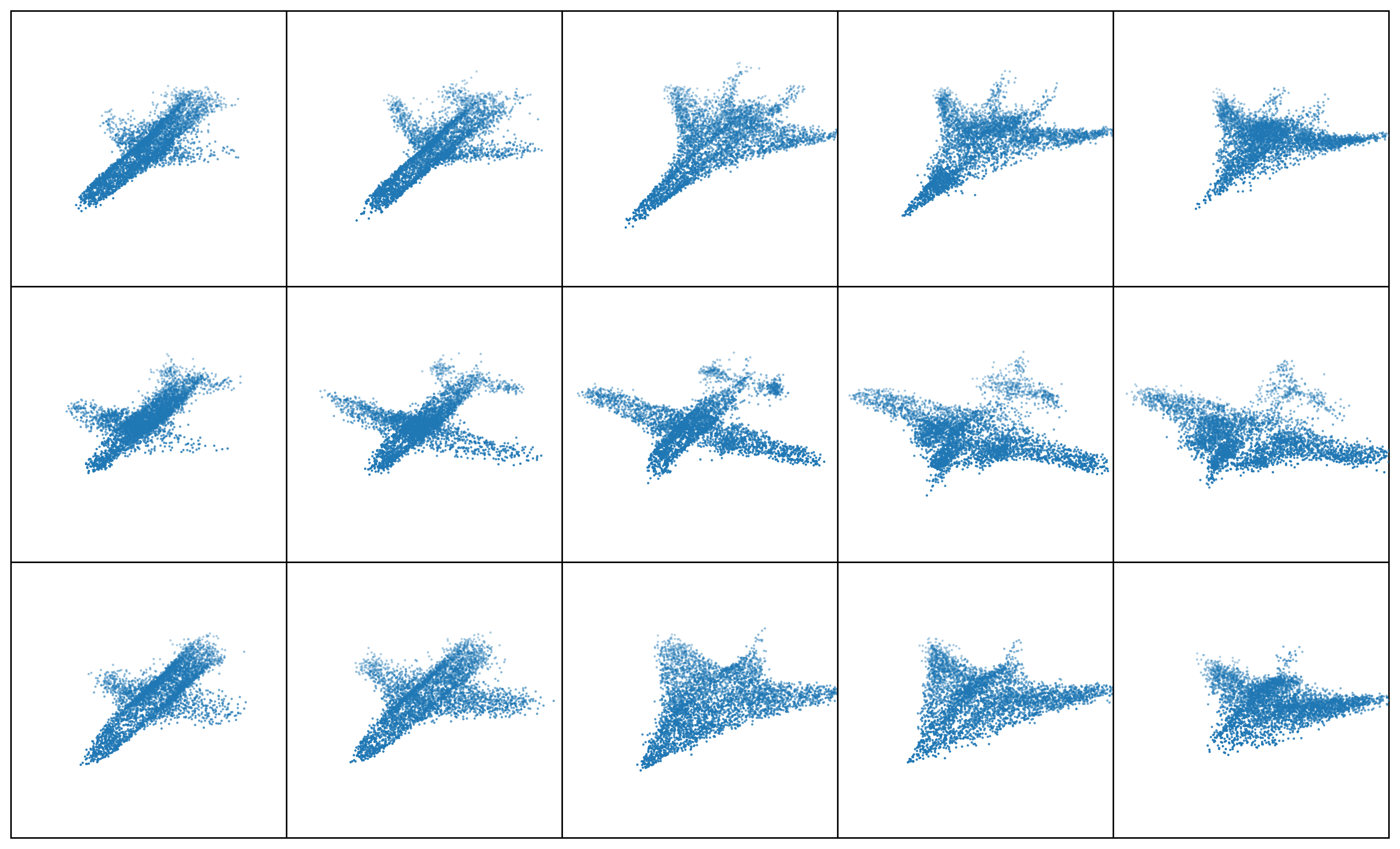}
    \label{app:fig:airplane_data_3_dir_9}
    }
    \end{subfigure}
\hfill
    \begin{subfigure}[Steerable factor: wing thickness (direction 5).]
    {\includegraphics[width=0.47\linewidth]{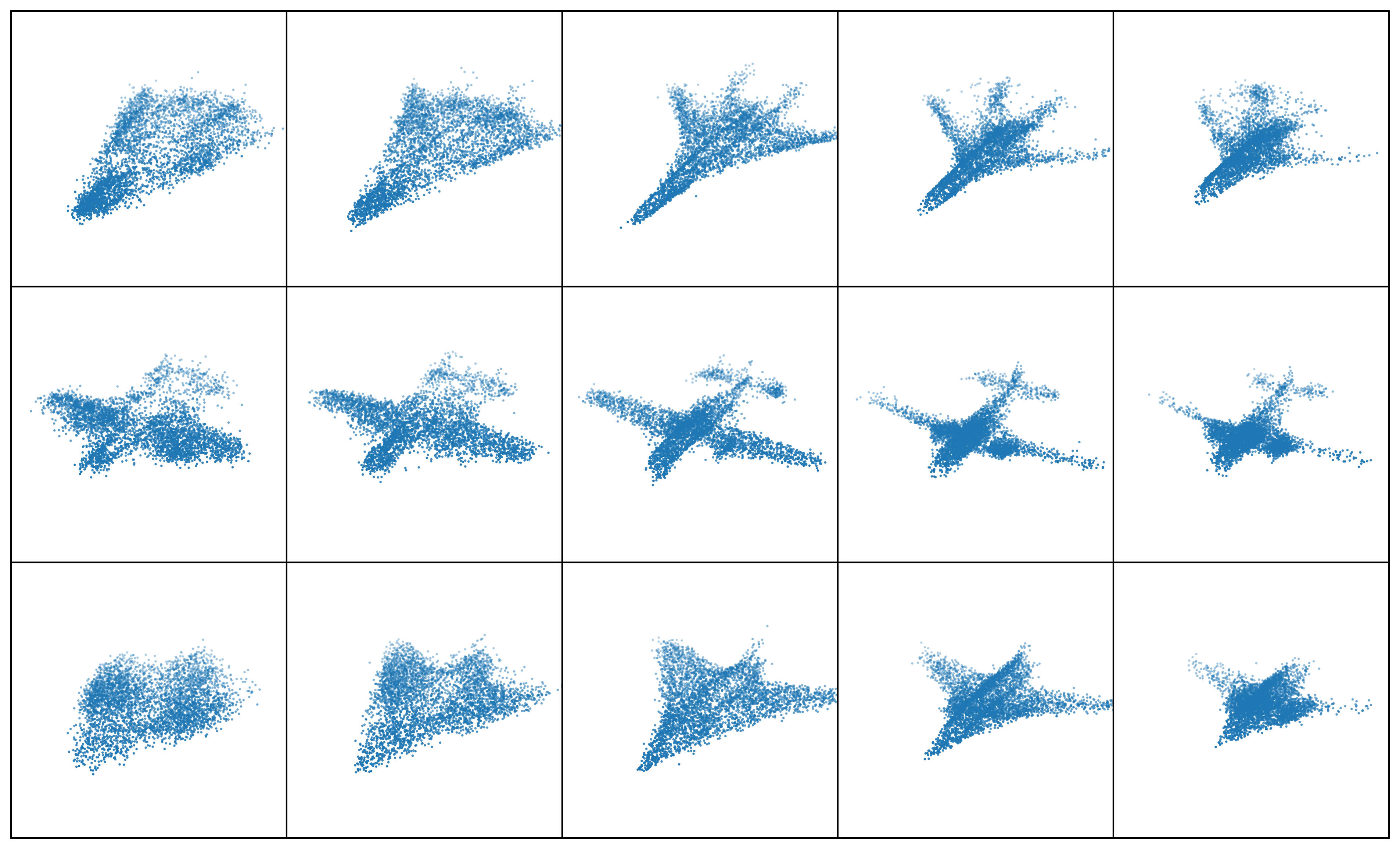}
    \label{app:fig:airplane_data_3_dir_19}
    }
    \end{subfigure}
\\
\vspace{-2ex}
\caption{
\model{} for point clouds (Airplane) editing. It can successfully reflect these steerable factors: engine, fuselage length, wing size, wing shape, and wing thickness.
}
\end{figure}
\clearpage

\subsection{Non-linear Editing Function} \label{app:sec:point_clouds_nonlinear_results}

\begin{figure}[H]
\centering
    \begin{subfigure}[Steerable factor: wing shape (direction 1).]
    {\includegraphics[width=0.47\linewidth]{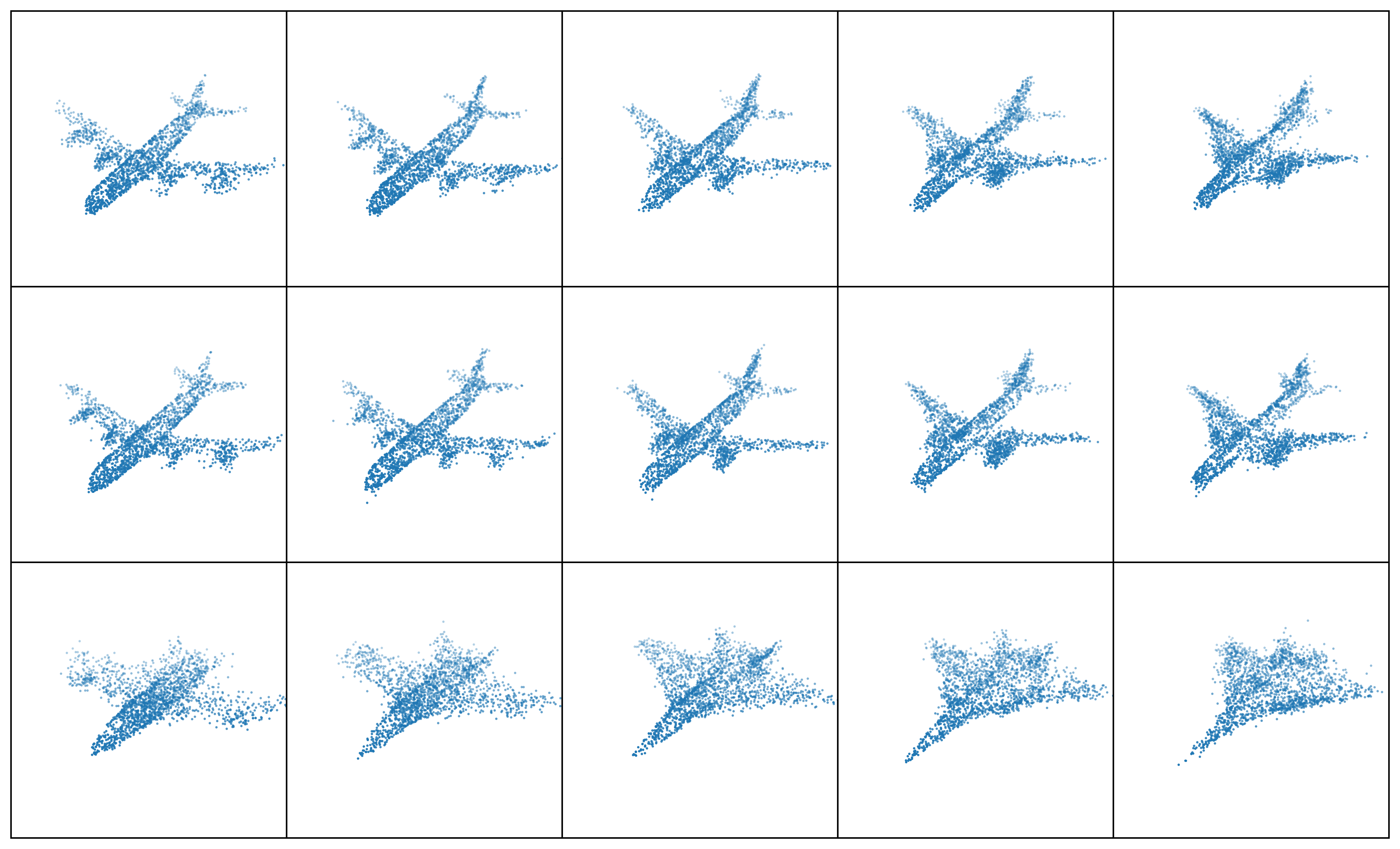}
    }
    \end{subfigure}
\hfill
    \begin{subfigure}[Steerable factor: wing length (direction 2).]
    {\includegraphics[width=0.47\linewidth]{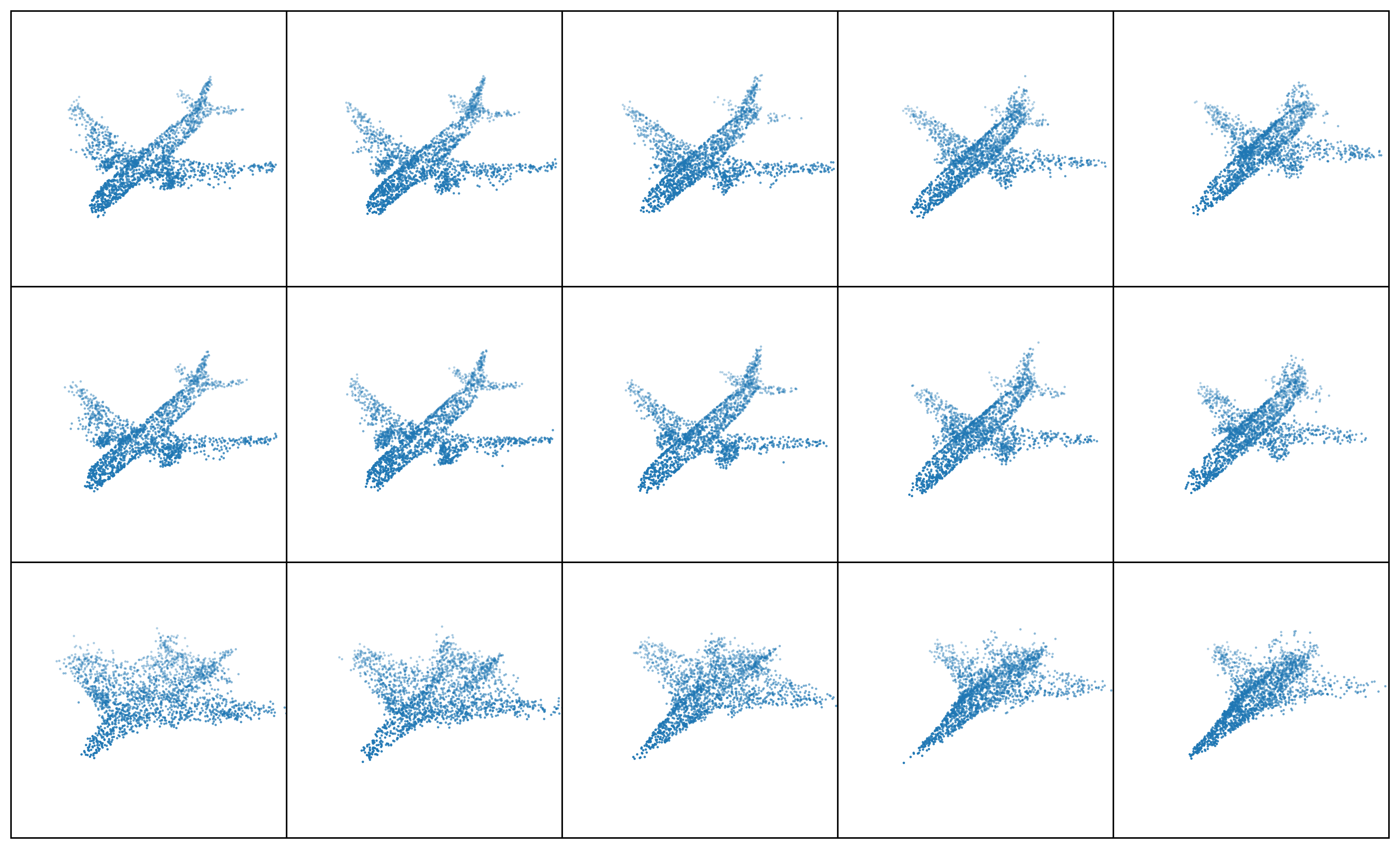}
    }
    \end{subfigure}
\\
\vspace{-2ex}
\caption{
\model{} for point clouds (Airplane) editing. It can successfully reflect these steerable factors: wing shape and wing length.
}
\end{figure}

\begin{figure}[H]
\centering
    \begin{subfigure}[Steerable factor: size (direction 1).]
    {\includegraphics[width=0.47\linewidth]{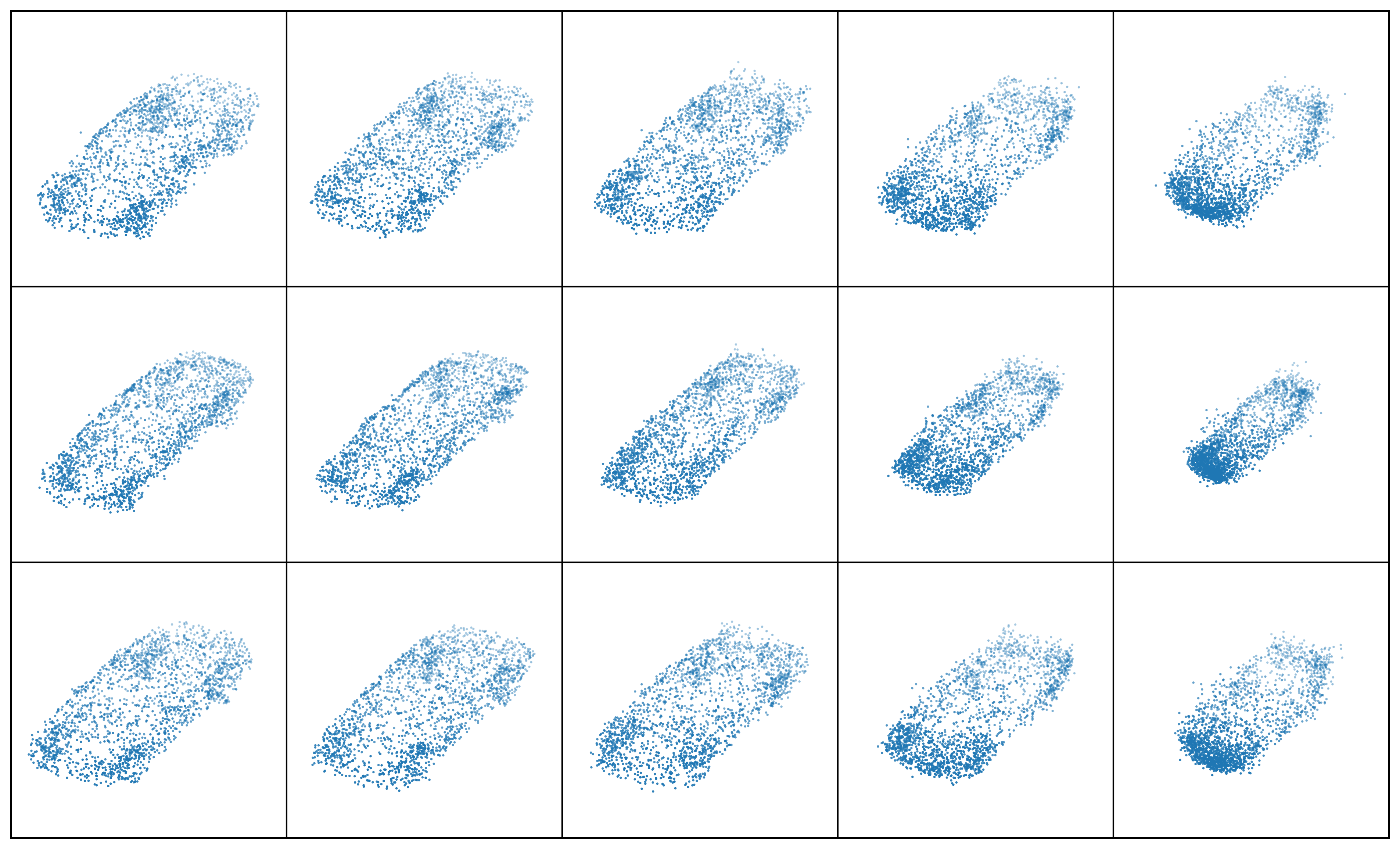}
    }
    \end{subfigure}
\hfill
    \begin{subfigure}[No obvious steerable factor (direction 2).]
    {\includegraphics[width=0.47\linewidth]{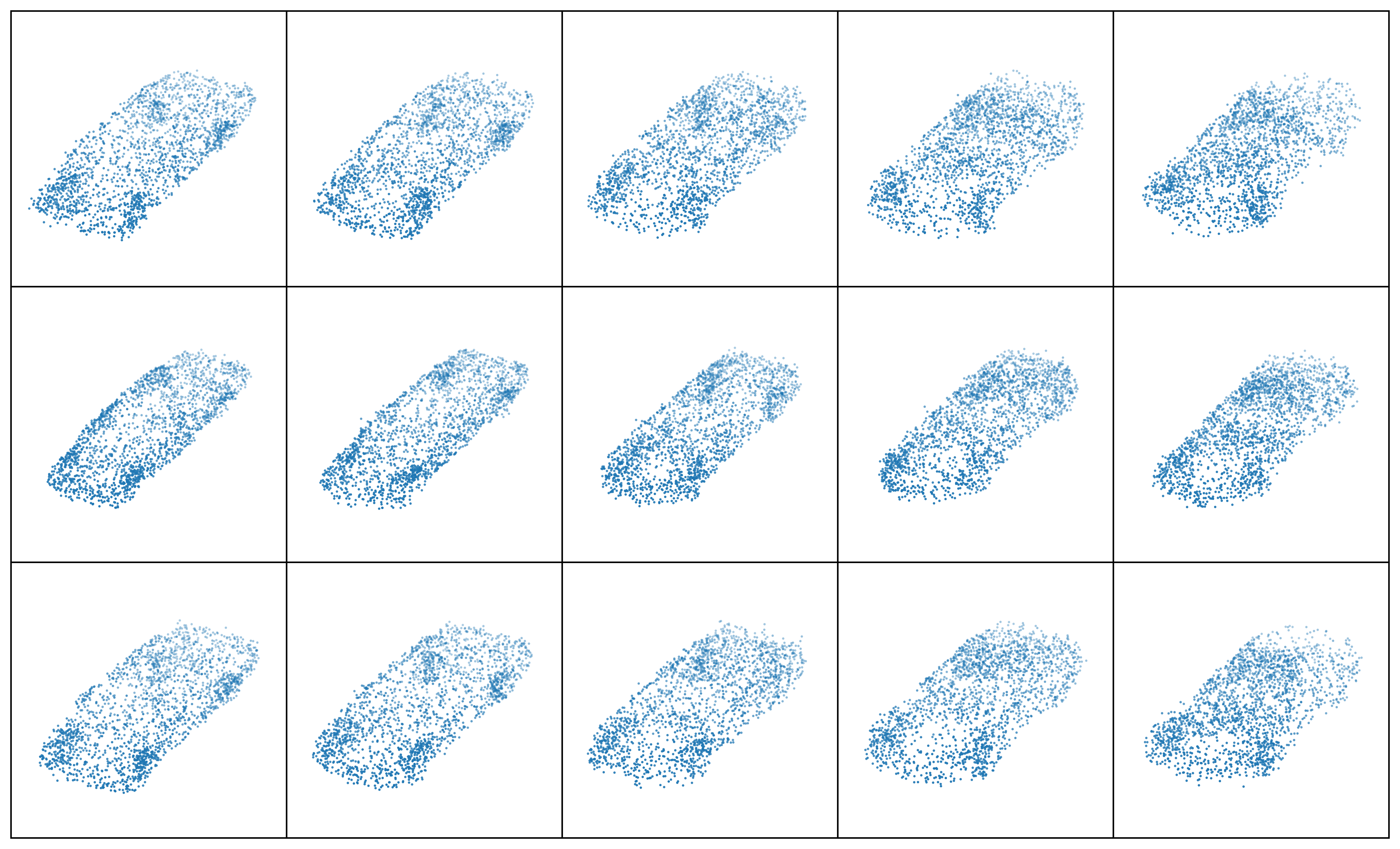}
    }
    \end{subfigure}
\\
\vspace{-2ex}
\caption{
\model{} for point clouds (Car) editing. The steerable factors on this dataset are not obvious, and here we only plot the car size editable with one directions.
}
\end{figure}

\begin{figure}[H]
\centering
    \begin{subfigure}[Steerable factor: leg height (direction 1).]
    {\includegraphics[width=0.47\linewidth]{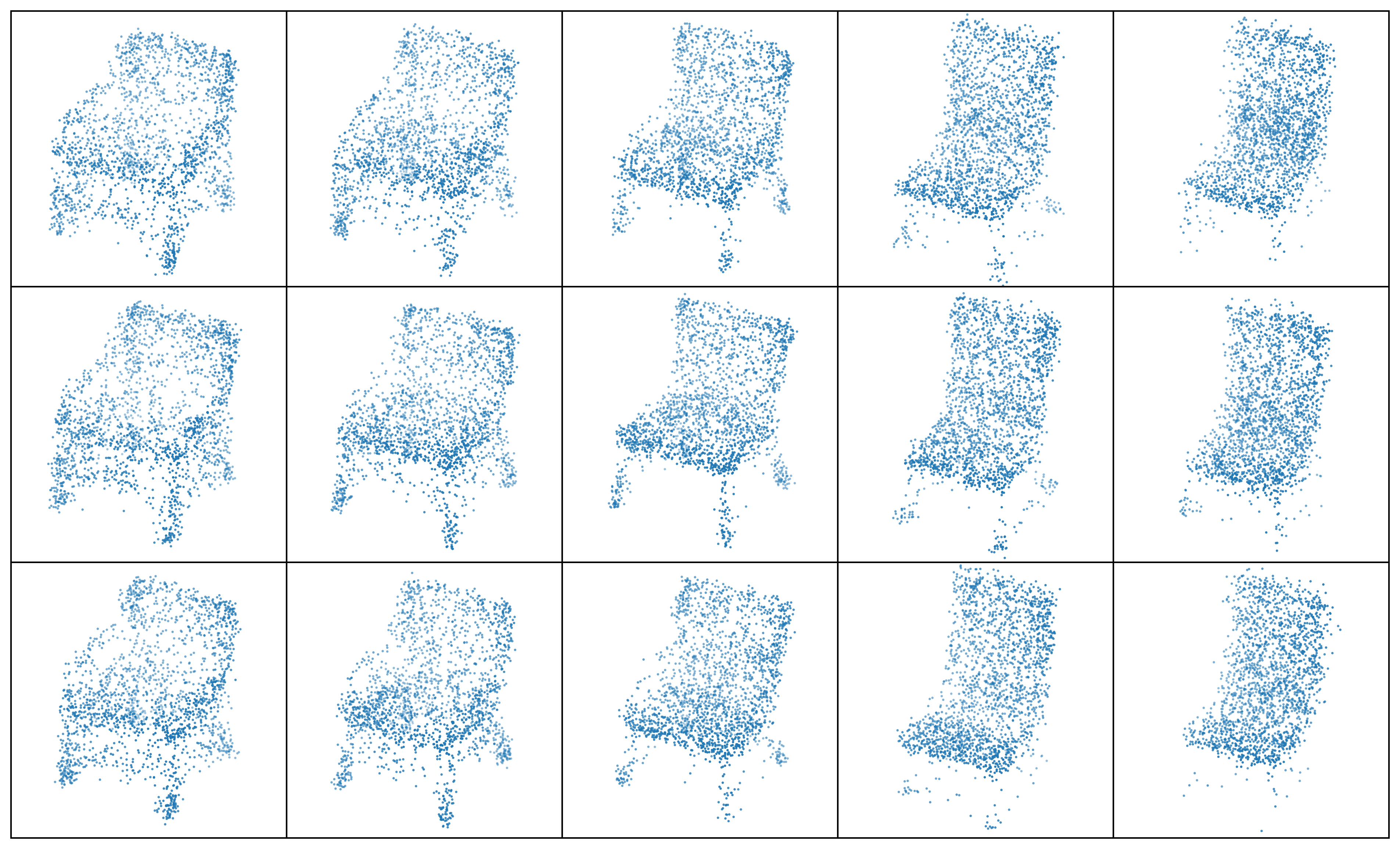}
    }
    \end{subfigure}
\hfill
    \begin{subfigure}[Steerable factor: seat size (direction 2).]
    {\includegraphics[width=0.47\linewidth]{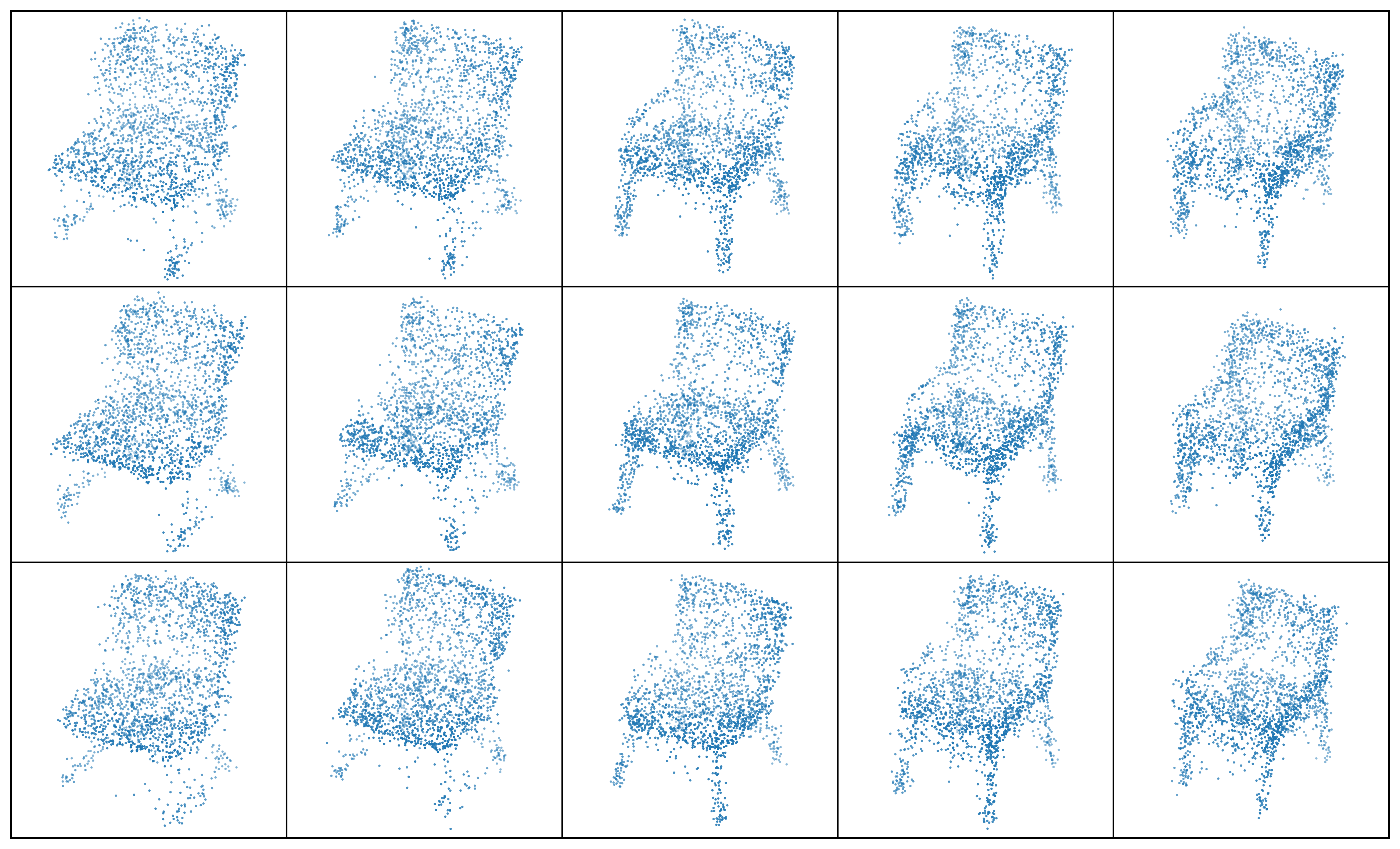}
    }
    \end{subfigure}
\\
\vspace{-2ex}
\caption{
\model{} for point clouds (Chair) editing. It can successfully reflect these steerable factors: leg height, and seat size.
}
\end{figure}
\clearpage

\end{document}